%% file: Thesis.tex
\documentclass[a4paper, 12pt, oneside ,table,xcdraw]{LJMU_Astro_Thesis}  

\graphicspath{{Figures/}}  

\usepackage[square, numbers, comma, sort&compress]{natbib}  
\setcitestyle{authoryear,open={(},close={)}} 

\usepackage{algorithm}
\usepackage{multirow}
\usepackage{eqnarray,amsmath}
\usepackage{algpseudocode}
\usepackage{verbatim}  
\usepackage{vector}  
\usepackage{pdfpages}
\hypersetup{urlcolor=black, colorlinks=true}  

\begin{document}
\frontmatter      

\title  {Point Cloud Registration of non-rigid objects in sparse 3D Scans with applications in Mixed Reality}
\authors  {\texorpdfstring
            {\href{https://www.manoramajha.com/}{Manorama Jha}}
            {Manorama Jha}
            }
\addresses  {\groupname\\\deptname\\\univname}  
\date       {\today}
\subject    {}

\maketitle

\setstretch{1.5}  

\fancyhead{}  
\rhead{\thepage}  
\lhead{}  

\pagestyle{fancy}  

\pagestyle{plain}
\pagenumbering{roman}
\setcounter{page}{2}
\include{preface}
\newpage
\input{declaration}
\newpage
\input{abstract}
\newpage
\input{acknowledgements}
\newpage
\Declaration{

\addtocontents{toc}{\vspace{1em}}  

I, \textbf{Manorama Jha}, declare that this thesis titled, "\textbf{Point Cloud Registration of non-rigid objects in sparse 3D Scans with
applications in Mixed Reality}" and the work presented in it are my own. I confirm that:

\begin{itemize} 
\item[\tiny{$\blacksquare$}] This work was done wholly or mainly while in candidature for a research degree at this University.
 
\item[\tiny{$\blacksquare$}] Where any part of this thesis has previously been submitted for a degree or any other qualification at this University or any other institution, this has been clearly stated.
 
\item[\tiny{$\blacksquare$}] Where I have consulted the published work of others, this is always clearly attributed.
 
\item[\tiny{$\blacksquare$}] Where I have quoted from the work of others, the source is always given. With the exception of such quotations, this thesis is entirely my own work.
 
\item[\tiny{$\blacksquare$}] I have acknowledged all main sources of help.
 
\item[\tiny{$\blacksquare$}] Where the thesis is based on work done by myself jointly with others, I have made clear exactly what was done by others and what I have contributed myself.
\\
\end{itemize}

Signed: Manorama Jha\\
\rule[1em]{25em}{0.5pt}  
 
Date: 07.12.2022\\
\rule[1em]{25em}{0.5pt}  
}
\clearpage  

\pagestyle{empty}  

\null\vfill
\begin{center}
\textit{`` DREAM, DREAM, DREAM - Till your dreams come true :) ''}
\end{center}


\vfill\vfill\vfill\vfill\vfill\vfill\null
\clearpage  

\addtotoc{Abstract}  
\abstract{
\addtocontents{toc}{\vspace{1em}}  

Point Cloud Registration is the problem of aligning the corresponding points of two 3D point clouds referring to the same object. The challenges include dealing with noise and partial match of real-world 3D scans. For non-rigid objects, there is an additional challenge of accounting for deformations in the object shape that happen to the object in between the two 3D scans. In this project, we study the problem of non-rigid point cloud registration for use cases in the Augmented/Mixed Reality domain. We focus our attention on a special class of non-rigid deformations that happen in rigid objects with parts that move relative to one another in joints, for example, robots with hands and machines with hinges. We propose an efficient and robust point-cloud registration workflow for such objects and evaluate it on real-world data collected using Microsoft Hololens 2, a leading Mixed Reality Platform.

}

\clearpage  

\setstretch{1.3}  



\clearpage  

\pagestyle{fancy}  

\lhead{\emph{Contents}}  
\tableofcontents  

\lhead{\emph{List of Figures}}  
\listoffigures  

\lhead{\emph{List of Tables}}  
\listoftables  

\setstretch{1.5}  
\clearpage  
\lhead{\emph{Abbreviations}}  
\listofsymbols{ll}  
{
\textbf{PCR} & \textbf{P}oint \textbf{C}loud \textbf{R}egistration\\
\textbf{MR} & \textbf{M}ixed \textbf{R}eality\\
\textbf{HMD} & \textbf{H}ead \textbf{M}ounted \textbf{D}isplay\\
\textbf{RGB} & \textbf{R}ed \textbf{G}reen \textbf{B}lue\\
\textbf{IMU} & \textbf{I}nertial \textbf{M}easurement \textbf{U}nit\\
\textbf{3D} & \textbf{3}-\textbf{D}imensional\\
\textbf{GPU} & \textbf{G}raphics \textbf{P}rocessing \textbf{U}nit\\
\textbf{FPGA} & \textbf{F}ield \textbf{P}rogrammable \textbf{G}ate \textbf{A}rray\\
\textbf{UX} & \textbf{U}ser \textbf{E}xperience\\
\textbf{XR} & \textbf{E}xtended \textbf{R}eality\\
\textbf{MRI} & \textbf{M}agnetic \textbf{R}esonance \textbf{I}maging\\
\textbf{CT} & \textbf{S}can \textbf{C}omputed \textbf{T}omography \textbf{S}can\\
\textbf{2D-CNN} & \textbf{C}onvolutional \textbf{N}eural \textbf{N}etworks\\
\textbf{3D-CNN} & \textbf{C}onvolutional \textbf{N}eural \textbf{N}etworks\\
\textbf{MLP} & \textbf{M}ulti \textbf{L}ayer \textbf{P}erceptron\\
\textbf{KPConv} & \textbf{K}ernel \textbf{P}oint \textbf{C}onvolution\\
\textbf{2D} & \textbf{2}-\textbf{D}imensional\\
\textbf{RANSAC} & \textbf{R}ANdom \textbf{S}Ample \textbf{C}onsensus\\
\textbf{ICP} & \textbf{I}terative \textbf{C}losest \textbf{P}oint\\
\textbf{MP} & Mega Pixel\\
\textbf{CAD} & \textbf{C}omputer \textbf{A}ided \textbf{D}esign\\
\textbf{SLAM} & \textbf{S}imultaneous \textbf{L}ocalization \textbf{A}nd \textbf{M}apping () \\
\textbf{IR} & \textbf{I}nlier \textbf{R}atio\\
\textbf{NFMR} & \textbf{N}on-rigid \textbf{F}eature \textbf{M}atching \textbf{R}ecall\\
\textbf{KPFCN} & \textbf{F}ully \textbf{C}onvolutional \textbf{N}eural \textbf{N}etwork \\
\textbf{KNN} & \textbf{K}-\textbf{N}earest \textbf{N}eighbor\\
\textbf{N-ICP} & \textbf{N}on-\textbf{R}igid \textbf{I}terative \textbf{C}losest \textbf{P}oint \\
\textbf{NASA} &  \textbf{N}ational \textbf{A}eronautics and \textbf{S}pace \textbf{A}dministration\\

}
\textbf{}\textbf{}



\setstretch{1.3}  

\pagestyle{empty}  
\dedicatory{
\ldots}

\addtocontents{toc}{\vspace{2em}}  

\mainmatter	  
\pagestyle{fancy}  


\input{Chapters/Chapter1} 

\input{Chapters/Chapter2} 

\input{Chapters/Chapter3} 

\input{Chapters/Chapter4} 

\input{Chapters/Chapter5} 

\input{Chapters/Chapter6} 

\input{Chapters/Chapter7} 

\nocite{*}
\label{Bibliography}
\lhead{\emph{Bibliography}}  
\bibliographystyle{mn2e} 
\bibliography{Bibliography}  

\addtocontents{toc}{\vspace{2em}} 

\appendix 
\input{Appendices/Appendix-A}	

\addtocontents{toc}{\vspace{2em}}  
\backmatter


\end{document}

%% file: declaration.tex
\chapter*{Declaration}
\addcontentsline{toc}{chapter}{Declaration}
The work presented in this thesis was carried out at the School of Computer Science and Mathematics,
Liverpool John Moores University. Unless otherwise stated, it is the original work
of the author.\\
While registered as a candidate for the degree of Master of Science in Artificial Intelligence and Machine Learning, for which submission 
is now made, the author has not been registered as a candidate for any other award. This thesis
has not been submitted in whole, or in part, 
for any other degree.\\

\vfill
MANORAMA JHA\\
School of Computer Science and Mathematics\\
Liverpool John Moores University\\
James Parsons Building\\
3 Byrom Street\\
Liverpool\\
Merseyside\\
L3 3AF\\
UK\\
\vfill
{\sc \hfill\today}


%% file: abstract.tex
\chapter*{Abstract}
\addcontentsline{toc}{chapter}{Abstract}
Point Cloud Registration is the problem of aligning the corresponding points of two 3D point clouds referring to the same object. The challenges include dealing with noise and partial match of real-world 3D scans. For non-rigid objects, there is an additional challenge of accounting for deformations in the object shape that happen to the object in between the two 3D scans. In this project, we study the problem of non-rigid point cloud registration for use cases in the Augmented/Mixed Reality domain. We focus on a special class of non-rigid deformations that happen in rigid objects with parts that move relative to one another in joints, for example, robots with hands and machines with hinges. We propose an efficient and robust point-cloud registration workflow for such objects and evaluate it on real-world data collected using the Microsoft Hololens 2, a leading Mixed Reality device.
\vfill
{\sc MANORAMA JHA \hfill\today}


%% file: acknowledgements.tex
\chapter*{Acknowledgements}
\addcontentsline{toc}{chapter}{Acknowledgements}
I would like to take this opportunity to thank all the people and organizations that have played a critical role in making this thesis a reality. Right at the beginning, I would like to thank GridRaster Inc. for providing me with a rare opportunity to carry out research at this wonderful intersection of Mixed Reality and Artificial Intelligence. I would like to especially thank my wonderful mentors from GridRaster -- Rishi, Pradeep, Bhaskar, Yiyong, Dijam and Sita -- for constantly supporting me in my job and encouraging me to pursue academic research.\\
I would like to thank my parents - Nityananda Jha and Ranjana Jha for going against all odds to support my education and well-being. I would like to thank my little brother Harimohan Jha for being my support system and always cheering me up when the time is low.\\
I would like to thank Prof. Pabitra Mitra and Prof. Jhareshwar Maiti at Indian Institute of Technology Kharagpur for introducing me to the world of Extended Reality. Without their kind support, it would not have been possible for me to build a career in Mixed Reality. I would also like to thank my mentor and friend Anirban Santara for helping me build a research portfolio in the early days of my career.\\
My life would be meaningless without my dear friends Anjali Sharma, Junmoni Boroghain and Khushi Gangireddy. I would like to thank them for being there for me and giving me hope when I could not find a reason to push on by myself.\\

\vfill
{\sc Manorama Jha \hfill\today}


%% file: Chapters/Chapter1.tex
\chapter{Introduction}
\label{ch:introduction}
\addcontentsline{toc}{chapter}{Introduction}
\lhead{\emph{Introduction}}
In this thesis, we study the problem of alignment in Mixed Reality (MR) with a special focus on deformable or non-rigid objects. The problem of alignment in MR is synonymous with the problem of Point Cloud Registration (PCR) in Computer Vision. PCR entails finding a spatial transformation function that maps a source point cloud to a target point cloud (usually of the same object or scene as the source, but displaced or deformed) such that the corresponding points superimpose. PCR has numerous practical applications in a wide range of fields including design, manufacturing, and medicine. For example, for repairing a precision-engineered component in a damaged aircraft, being able to overlay a reference model of the component on the real instance would be of immense help to the technicians. PCR is performed in two steps - a) Correspondence Matching: determining matching pairs of points in the two-point clouds, and b) Transformation Estimation: estimating the transformation function that would register the corresponding points in both point clouds. Although the problem of point-cloud registration has been studied extensively in the case of rigid objects, it still remains a topic of extensive research for non-rigid objects. In real-world applications point cloud scans are seldom clean. For such conditions, a common challenge is dealing with noisy point clouds and only partial matches between the source and target point clouds. In this thesis, we will study the problem of point cloud registration for non-rigid objects in the presence of noise and partial match.

\section{Background}
In this section, we will briefly introduce the topic of Mixed Reality that forms the foundation of this thesis.

\subsection{Mixed Reality}
In the digital age of today, we humans and almost all things we interact with on a daily basis have a digital self in addition to physical existence. Mixed Reality (MR) is the idea of creating an immersive and seamless experience by blending digital entities in a real-world environment. Mixed reality applications can render accurate to-scale digital models of objects — both static and dynamic — and allow the user to interact with them using intuitive gestures that resemble the way they would interact with the real object. High-precision physics simulation allows accurate reproduction of real-world outcomes of these interactions — thus providing a highly capable tool for education, design, and planning. Mixed reality is also capable of virtual teleportation of human users by mapping their actions and expressions on virtual avatars, thus taking remote collaboration to a whole new level. Mixed reality tools are set to revolutionize engineering, design, productivity, education, and entertainment. This makes mixed reality one of the most actively researched areas of computer science at the moment. 

\subsection{MR Headsets}

\begin{figure}[!h]
    \centering
    \includegraphics[width=\textwidth]{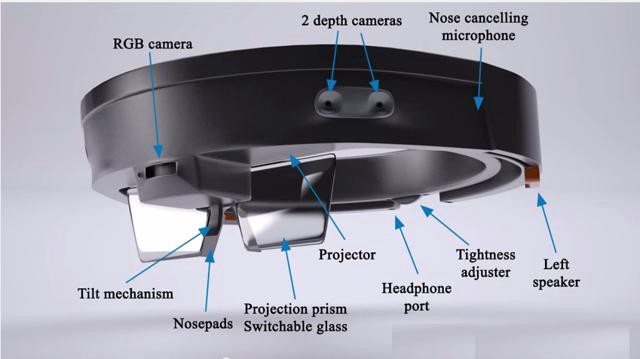}
    \caption{Microsoft HoloLens 2, the Mixed Reality platform used in our study.}
    \label{fig:hololens-headset}
\end{figure}

The most important hardware component of a Mixed Reality system is the Head Mounted Display (HMD) or Headset. Some popular examples are Microsoft HoloLens \citep{ungureanu2020hololens} and Magic Leap. Mixed Reality headsets have a transparent visor through which the user can see the world and different kinds of optics and display elements that project the virtual objects in the field of view of the user. An elaborate set of RGB cameras and laser scanners is used for spatial mapping, localization, and tracking. MR headsets also have an Inertial Measurement Unit (IMU) consisting of an accelerometer, a magnetometer, and a gyroscope that tracks the movement of the user’s head through space in six degrees of freedom. Figure \ref{fig:hololens-headset} shows the components of Microsoft HoloLens 2, the MR headset that we use in this thesis. MR Headsets provide multi-modal user interfaces that combine head gestures, voice commands, and eye gaze.

\subsection{Engineering Challenges faced by MR}
MR headsets must be power efficient, lightweight and ergonomic. Spatial mapping using multiple modalities, scene reconstruction, gesture recognition, voice recognition, and high-polygon rendering are some primary tasks that are at the core of most MR experiences. These 3D Computer Vision and Computer Graphics tasks are compute-intensive and the limited on-device resources are not adequate for supporting an immersive MR experience. To mitigate this issue, most MR platforms employ a hybrid architecture where the compute-intensive tasks like those involving deep neural networks and rendering are off-loaded to high-performance and specialized hardware (e.g. GPU, FPGA) in the cloud, and the on-board hardware is used for encoding-decoding of data, communication with the cloud and interactions with the user (UX), etc.

\section{Problem Statement}
In this project, we aim to study the problem of non-rigid partial point cloud registration for MR applications. MR is transforming the industry by providing unique opportunities in engineering design, remote operations, and collaboration. We specifically intend to implement a state-of-the-art point cloud registration algorithm - Lepard \citep{li2022lepard} - on Microsoft HoloLens 2 Mixed Reality Platform \citep{ungureanu2020hololens} and investigate potential causes of drift and latency bottlenecks in real-world MR use cases. The anticipated outcomes of this research project are a detailed account of failure modes of Lepard and related algorithms and a robust and fast inference pipeline that is robust to outliers and noise in real-life 3D scans and capable of supporting real-time MR applications.

\section{Aims and Objectives}

In this thesis, we aim to study the problem of 3D Point Cloud Registration for non-rigid objects. Apart from wide-spread relevance in computer vision, robotics, and medical science, this problem statement is central in many Mixed Reality applications. Although significant advancements have been made in 3D representation learning and computer vision in recent years, several engineering challenges stand in the way of deploying these algorithms at scale. The objectives of this thesis are as follows:

\begin{itemize}
    \item Prepare a thorough literature review of prominent approaches to 3D Point Cloud Registration with emphasis on non-rigid objects and relevance in Mixed Reality.
    \item Implement one of the current state-of-the-art algorithms - Lepard \citep{li2022lepard} - on Microsoft HoloLens 2 \citep{ungureanu2020hololens}.
    \item Evaluate real-time 3D point cloud registration for deformable and non-deformable objects in uncontrolled real-world settings in the presence and absence of distraction and with varying levels of noise in the 3D scans.
    \item Identify and mitigate any causes of drift or misalignment and potential bottlenecks slowing down the inference pipeline.
\end{itemize}

\section{Research Questions}
In this thesis we will address the following research question:

\begin{center}
\begin{minipage}{0.7\textwidth}
“Can we make state-of-the-art non-rigid point cloud registration algorithms robust and low-latency for Mixed Reality use cases in real-world environments?”
\end{minipage}
\end{center}

\section{Scope of the Study}
In this thesis, we will focus on the problem of non-rigid point cloud registration only. Although we use models trained on public benchmark PCR datasets, the focus of this thesis is to achieve high performance on scans collected using the mixed reality platform - Microsoft Hololens 2. 
We want to focus on objects that are composed of rigid parts that can move relative to one another (about a hinge, for example). Such objects are very common in enterprise mixed reality applications. Machine parts, doors, drawers and antennas are a few examples. We will only consider a recent state-of-the-art method Lepard \citep{li2022lepard} as our baseline. 

\section{Significance of the Study}
Mixed Reality (MR) is a form of eXtended Reality (XR) that allows the user to render and interact with virtual 3D holograms in a real-world environment [\citep{speicher2019mixed}, \citep{rokhsaritalemi2020review}]. In an interior design use case, MR would allow the user to render the hologram of a couch in an empty room and move it around the room with hand gestures [\citep{huang2018augmented}] to find out where in the room it would look best [\citep{kent2021mixed}, \citep{jain2019current}]. One of the primary use cases of this technology is to overlay or register a virtual reference model of an object on top of a real instance of the same object for comparison. For example, in manufacturing, a virtual reference model of a precision engineering part can be overlaid on a damaged instance for inspection, repair [\citep{abate2013mixed}, \citep{borsci2015empirical}] and training [\citep{mueller2003marvel}, \citep{kirkley2005creating}, \citep{wang2004mixed}]. In remote assistance, a circuit diagram can be overlaid on a real switchboard and remote instructions can be issued for debugging the connections \citep{ladwig2018literature}. In surgery, pre-operative scans (like MRI, CT Scan, etc.) can be overlaid on the patient’s organ in real-time for precise guidance to the point of surgery \citep{chen2017recent}. 

\section{Organization of the Thesis}

The thesis is organized as follows:

\begin{itemize}
    \item In order to carry out research on non-rigid point cloud registration it is crucial to build a strong understanding of the foundational concepts and algorithms. In Chapter \ref{ch:literature_review}, we describe different point cloud registration workflows and present a detailed survey of the literature in this field. It is also important to understand the challenges involved in this real world applications of these methods and we also discuss these in the same chapter.
    \item Point cloud registration is a vast area of computer vision and it is important to scope the problems studied in this Masters' thesis. In Chapter \ref{ch:research_methodology}, we chart out the questions that we aim to address in this thesis. We also describe the datasets, tools, and compute resources that we use in our study.
    \item As a primer to the proposed methods, we present a detailed account of the prior work that our proposed algorithm is based off in Chapter  \ref{ch:non_rigid_point_cloud_registration}. We identify the drawbacks of these methods when it comes to the application of our interest and set the stage for the presentation of our original contributions.
    \item Finally, in Chapter \ref{ch:proposed_method}, we present our algorithm for non-rigid registration of rigid objects with moving parts.
    \item In Chapter \ref{ch:results_and_discussion}, we present a detailed experimental study of real world evaluation of our proposed approach.
    \item We conclude the thesis in Chapter \ref{ch:conclusion} with a summary of our contributions and recommended directions for future work.
\end{itemize}





\section{Conclusion}
In this section, we introduced the problem statement studied in this thesis. We articulated our aims and objectives and research questions to be addressed. We also described the scope, significance and structure of this study.

%% file: Chapters/Chapter2.tex
\chapter{Literature Review}
\label{ch:literature_review}
\addcontentsline{toc}{chapter}{Literature Review}
\lhead{\emph{Literature Review}}

\section{Introduction}
In the previous chapter, we introduced the problem of point cloud registration and explained the motivation of this thesis - point cloud registration.
This chapter presents a review of point cloud registration algorithms. We also discuss the challenges of this task and ways in which people have approached a solution.

\section{Point Cloud Registration} 
Point cloud registration, also called alignment, is the task of finding a spatial transformation (such as rotation, transformation and scaling) which when applied to a source point cloud results in point-wise superposition with a destination point cloud of the same object/scene. These algorithms fall in two categories - rigid and non-rigid - depending on whether deformations in the object/scene between the two scans can be addressed.

The general approach to point-cloud registration involves two steps: a) feature extraction, and b) feature matching and registration.

\subsection{Feature extraction}
Each point in a point cloud must be assigned a feature vector representing its 3D position and context including local geometry, color and texture \cite{weinmann2017geometric}. The first step is to define a neighborhood for each point. Some common examples are spherical \cite{lee2002perceptual}, \cite{linsen2001local} or cylindrical \cite{filin2005neighborhood}, \cite{niemeyer2014contextual} neighborhoods parameterized by radius \cite{lee2002perceptual}, \cite{filin2005neighborhood} or the number of nearest neighbours by Euclidean distance \cite{linsen2001local}, \cite{niemeyer2014contextual}. These parameters give us the means to control the scale at which local 3D structures must be encoded. The value of the scale parameter is usually chosen using prior knowledge \cite{weinmann2015contextual} or learned from data \cite{weinmann2015semantic}, \cite{mitra2003estimating}, \cite{lalonde2005scale} \cite{demantke2011dimensionality} and multi-scale approaches are also popular \cite{niemeyer2014contextual}, \cite{brodu20123d}, \cite{schmidt2014contextual}. Once a neighbourhood of a point is defined, feature extraction encodes the local 3D geometry to attach semantics or context to the point. Certain shape primitives can be obtained by computing the eigenvalues of a 3D structure tensor constructed using the spatial coordinates of neighbouring points \cite{jutzi2009nearest}, \cite{west2004context}, \cite{pauly2003multi}. Other features that are extracted are angular characteristics \cite{munoz2009contextual}, height \cite{mallet2011relevance}, moments \cite{hackel2016fast}, surface properties, slope, vertical profiles and 2D projections \cite{guo2015classification}, shape distributions \cite{osada2002shape}, \cite{blomley2016classification}, and point-feature histograms \cite{rusu2009fast}. 

Modern approaches use Deep Neural Network \citep{goodfellow2016deep} based representation learning. Projection Networks project the 3D point cloud onto 2D image planes from multiple viewpoints and use 2D Convolutional Neural Networks (2D-CNN) \citep{goodfellow2016deep} to process them \citep{su2015multi}, \citep{boulch2017unstructured}, \citep{lawin2017deep}. Voxel-based methods project the point cloud onto a 3D grid \citep{maturana2015voxnet}, \citep{roynard2018classification}, \citep{ben20183dmfv}. Sparse data structures like octree and hash-maps are used for better efficiency and larger context sizes \citep{riegler2017octnet,graham20183d}. These grids are further processed using 3D Convolutional Neural Networks (3D-CNN) \citep{goodfellow2016deep}. The main drawbacks of these approaches arise from the loss of details in the original point cloud structure during projection onto grids \citep{thomas2019kpconv}. Graph Convolutional Networks address this problem by retaining the original position of each point and combining features on local surface patches \citep{verma2018feastnet,wang2019dynamic}. However, their representation is invariant to the deformations of those patches in Euclidean space which is not helpful for estimating non-rigid transformations between point clouds \citep{thomas2019kpconv}. Pointwise Networks like PointNet \citep{qi2017pointnet} apply a shared neural network to each point followed by global max-pooling. This approach set new benchmarks in point-cloud classification and different variants using Multi-Layer Perceptron (MLP) \citep{qi2017pointnet++}, \citep{liu2019point2sequence,li2018so} and CNN \citep{hua2018pointwise,xu2018spidercnn,groh2018flex,atzmon2018point}.

\begin{figure}[!h]
    \centering
    \includegraphics[width=\textwidth]{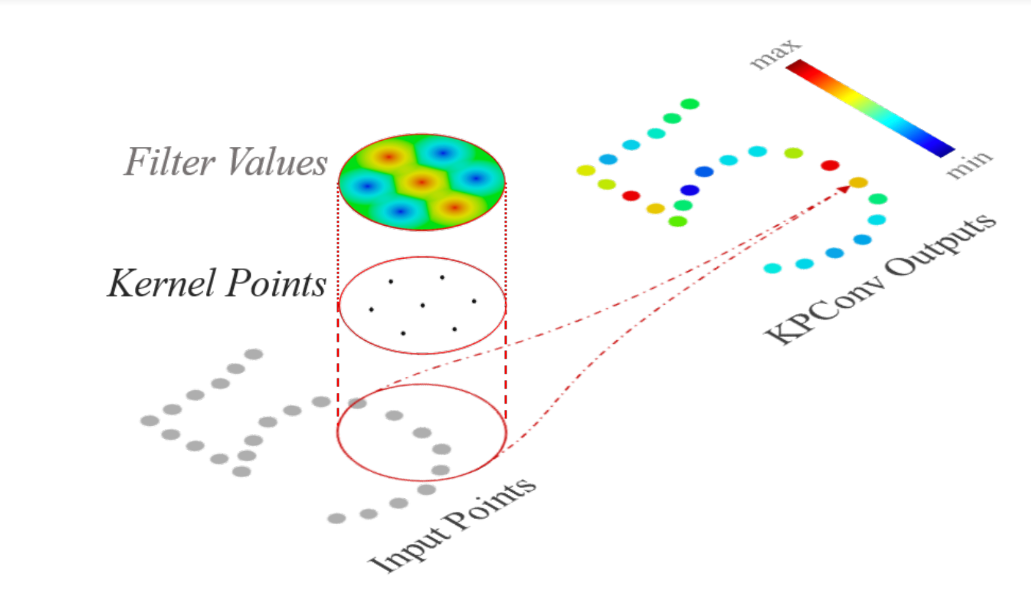}
    \caption{Kernel Point Convolution \citep{thomas2019kpconv} (KPConv) demonstrated on 2D points.}
    \label{fig:KPConv}
\end{figure}

Kernel Point Convolution (KPConv) \citep{thomas2019kpconv} is one of the recent breakthroughs in point cloud representation learning and is particularly suitable for deformable point clouds. Figure \ref{fig:KPConv} illustrates the procedure. It learns a kernel function to compute pointwise filters and increases representation power using deformable kernels. Instead of a grid-shaped kernel (as is the case with regular 2D and 3D CNNs), the kernel points in KPConv are spread freely in space. Each kernel point accumulates the features of the point-cloud points within a spherical neighborhood around itself with weights that decay as the points get farther. In deformable KPConv, each kernel point also has a learnable offset that allows it to learn to adapt the shape of a kernel to different inputs. We will use KPConv to learn representations of our point clouds in the experiments for this thesis.

\subsection{ Feature matching and registration}
After the points have been represented using feature vectors, the next task is to find corresponding points between the given pair of point clouds. 

\subsubsection{Rigid Registration} 

Traditional methods use different variants of the RANdom SAmple Consensus (RANSAC) algorithm for finding matching points \citep{holz2015registration,schnabel2007efficient}. RANSAC is an iterative algorithm and works by classifying all possible correspondences into inliers and outliers. RANSAC is simple yet powerful and a bulk of the literature on point cloud registration uses RANSAC with both handcrafted and learned features \citep{ao2021spinnet}, \citep{bai2020d3feat}, \citep{choy2019fully}, \citep{deng2018ppf}, \citep{deng2018ppfnet}, \citep{el2021unsupervisedr}, \citep{yu2021cofinet}, \citep{zeng20173dmatch}. For rigid point clouds, another popular algorithm is Iterative Closest Point (ICP) \citep{arun1987least}, \citep{besl1992method}, \citep{zhang1994iterative}. ICP assumes that the two point clouds are spatially close. In each iteration, for each point in the source point cloud, the closest point in the destination cloud is assigned as the matching point. Then the rigid transformation (rotation and translation) is computed by minimizing mean-square loss. These steps are repeated till convergence. Direct Registration approaches combine feature extraction, feature matching, and registration within a single architecture through end-to-end pose optimization \citep{besl1992method}, \citep{zhou2016fast}, \citep{aoki2019pointnetlk}, \citep{choy2020deep}. 

\subsubsection{Non-rigid registration}

Non-rigid correspondence has the added challenge of accounting for deformations. Different approaches have been proposed including projective correspondence \citep{newcombe2015dynamicfusion}, Siamese Network \citep{schmidt2016self}, and Scene Flow estimation \citep{li2021neural}, \citep{liu2019flownet3d}, \citep{puy2020flot}. Non-rigid correspondence is also studied in Geometry Processing where the input data is in the form of manifold surfaces. Methods like isometric deformation \citep{huang2008non}, latent code optimization \citep{groueix20183d}, and functional maps \citep{ovsjanikov2012functional} have been studied in this context.

\subsection{Challenges}

Real-world applications of point cloud registration are faced with several challenges. We discuss these in this section.

\subsubsection{Lack of ordering or structure}
A point cloud is a collection of points in 3D space with no ordering or structure (like a grid). The absence of a grid structure makes it difficult to apply traditional deep learning methods like CNNs because they leverage the grid structure of their inputs (images/videos) for learning representations. As described in the previous section, there have been attempts \citep{lawin2017deep,maturana2015voxnet,roynard2018classification,ben20183dmfv}, to coerce/quantize/project point clouds to grids in order to harness the power of traditional CNNs while trading off the loss of the inherent structural details of the point cloud. New forms of convolution like KPConv \citep{thomas2019kpconv} also attempt to address this challenge with flexible non-grid kernels.

\subsubsection{Sparsity}

\begin{figure}[!h]
    \centering
    \includegraphics[width=0.7\textwidth]{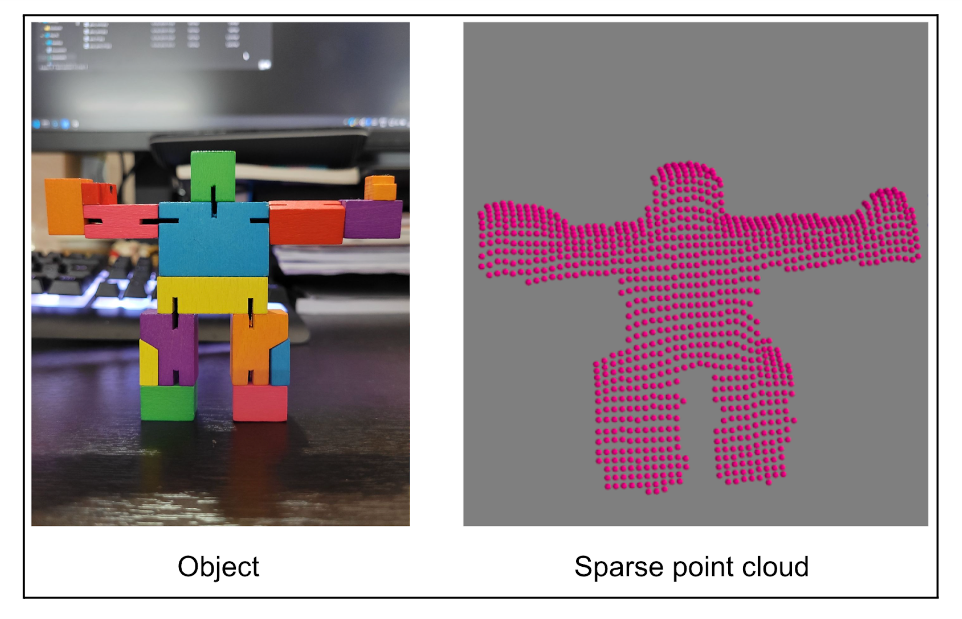}
    \caption{Example of sparse point cloud scan}
    \label{fig:sparse-pc}
\end{figure}

Most real-world 3D scans tend to be sparse \ref{fig:sparse-pc}. As a result of this, a large amount of computing gets wasted processing zero entries. For efficient processing of sparse point clouds using Deep Neural Networks, the recently proposed Minkowski Engine \citep{choy20194d} has proved to be a game changer. It represents point clouds as a position-indexed array and performs computation only for invalid regions. It can also leverage specialist hardware like GPU for greater throughput.

\subsubsection{Noise and Outliers}

\begin{figure}[!h]
    \centering
    \includegraphics[width=0.5\textwidth]{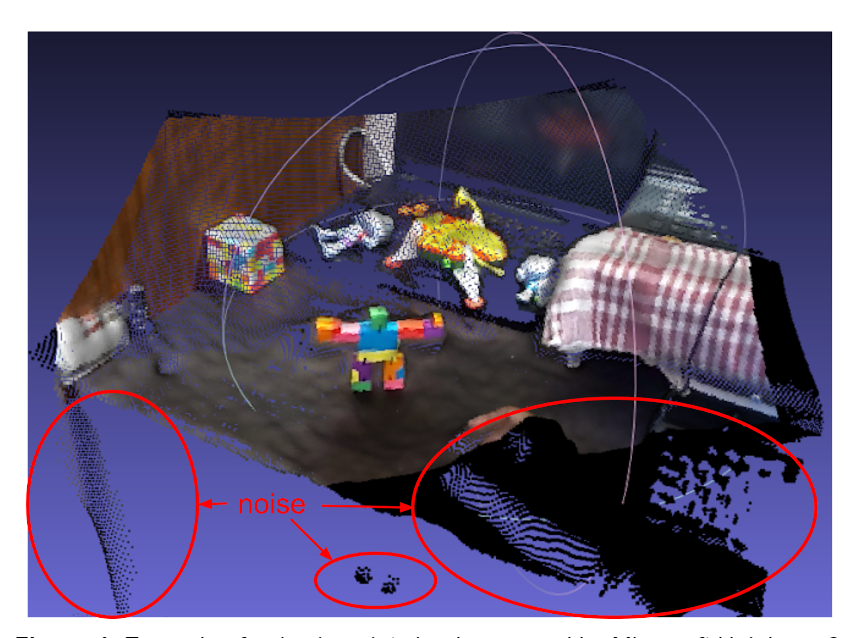}
    \caption{Example of noise in point clouds scanned by Microsoft HoloLens 2.}
    \label{fig:noisy-pc}
\end{figure}

Real-world 3D scans contain a large amount of noise and outliers (e.g. Figure \ref{fig:noisy-pc}). These are points that do not belong to the intended surface. One of the sources of outliers is a specular reflection from shiny surfaces like metal. In rigid registration, the RANSAC algorithm \citep{holz2015registration,schnabel2007efficient} attempts to remove the outliers that do not conform with the model with maximum consensus. Learning approaches to outlier detection and removal include \citep{bai2021pointdsc,pais20203dregnet}.

\subsubsection{Partial Overlap}
\begin{figure}[!h]
    \centering
    \includegraphics[width=\textwidth]{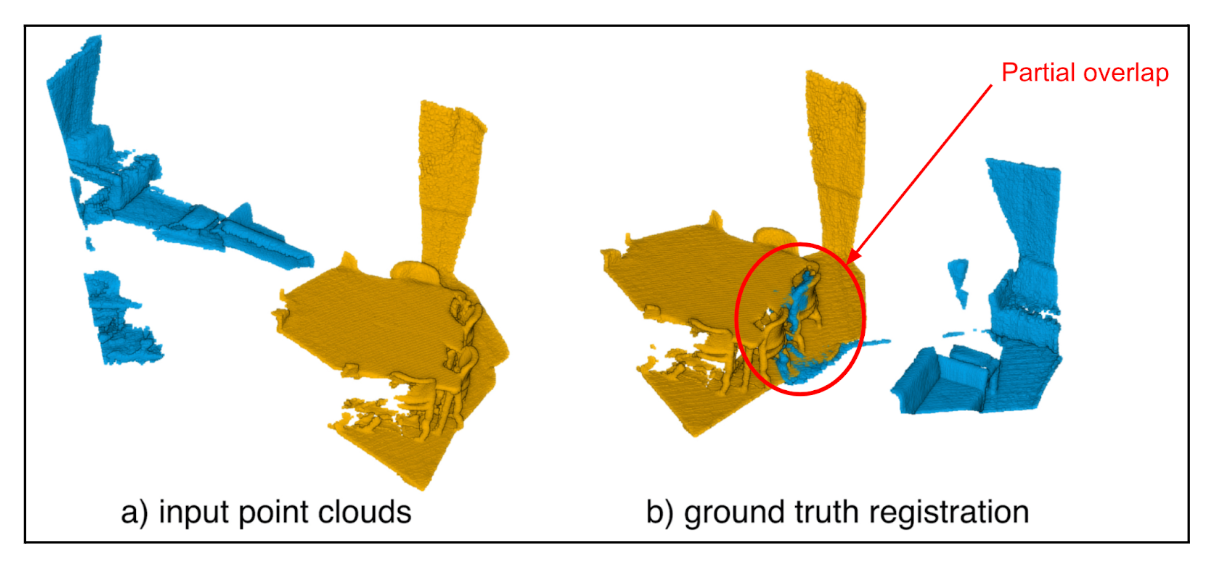}
    \caption{Example of partial overlap between the source and target point clouds}
    \label{fig:partial-overlap}
\end{figure}

In a real-world use case of point cloud registration, the source and target point clouds may only have a partial match \ref{fig:partial-overlap} albeit the fact that they are of the same object. This might be due to occlusion, object motion, or viewpoint change. A variety of methods have been proposed to perform the registration of point clouds that have a partial match \citep{thomas2019kpconv,xu2021omnet, attaiki2021dpfm, rodola2017partial}.

\section{Conclusion}
In this chapter, we presented a review of point cloud registration algorithms. We presented the challenges of point cloud registrations that arise from real world use cases. This section serves to provide a background for the studies that we present in the following chapters of the thesis.

%% file: Chapters/Chapter3.tex
\chapter{Research Methodology}
\label{ch:research_methodology}
\addcontentsline{toc}{chapter}{Research Methodology}
\lhead{\emph{Research Methodology}}

\section{Introduction}
In this chapter, we present our research methodology. We describe the datasets used for training the neural network models used in our experiments. We also present the details of our real world evaluation setup. We use different point cloud registration metrics to evaluate the proposed algorithms. These are also described in this chapter.

\section{Research Tools}
First, we discuss our research tools. Our research tools include specialised hardware and software. 

\subsection{Hardware}
The hardware required in this project can be categorized into three categories:

\begin{itemize}
    \item Data capture devices,
    \item Data processing devices, and
    \item Data rendering devices.
\end{itemize}

\subsubsection{Data capture devices}
For the capture of 3D data, we use Microsoft HoloLens 2 \citep{ungureanu2020hololens}. As shown in Figure \ref{fig:hololens-headset}, Hololens 2 comes with a $1$-MP time-of-flight based depth sensor for scanning point clouds and an $8$-MP RGB camera for capturing textures. The depth camera has short and long throw modes for capturing points at different distances from the user. In addition to this, it has $4$ visible light cameras for head tracking. This produces posed point clouds. 

\subsubsection{Data processing devices}
For computation, we use two high-performance laptop computers with one GPU each. The following are the details:
\begin{itemize}
  \item 2020 Lenovo Legion with RTX-2080 GPU. This GPU comes with $8$-GB GDDR6 memory.
  \item 2016 Asus Republic Of Gamers with GTX-1080 GPU. This GPU comes with $8$-GB GDDR5X memory.
\end{itemize}
Large point clouds can be tricky to fit along with large neural network models within the limited GPU memory. As explained in Section \ref{sec:data-preparation}, we may need to downsample point clouds before feeding them into the registration pipeline. This creates a trade-off between the use of high-performance neural networks that are heavy on GPU memory and the level of detail in the input point clouds.

\subsubsection{Data rendering devices}
For rendering the results of registration, we use the same Microsoft Hololens 2. The source CAD model is assumed to be located at the origin of the Hololens frame coordinates. However, during the computation of transformations for registration, we use world coordinates for both the source and target point clouds. The mapping from Hololens frame coordinates and world coordinates is obtained through Simultaneous Localization And Mapping (SLAM) \citep{durrant2006simultaneous}. After the transformations for registration have been obtained in the world coordinate system, they are converted into the Hololens' current frame coordinates. These transformations are then applied to the source model and the transformed model is rendered.  

\subsection{Software}
Our registration codebase is a fork of our open-source baseline, Lepard [\citep{li2022lepard}. The code is written mostly in Python. We use HoloLens 2 Research Mode \citep{ungureanu2020hololens} for processing the raw scans from the MR platform and extracting and processing point clouds. For visualisation, we use Unity \citep{jerald2014developing}, CloudCompare \citep{girardeau2016cloudcompare} and Meshlab \citep{cignoni2011meshlab}.

\section{Datasets for our Study}
\label{sec:datasets}
We will consider training and fine-tuning our models on the following datasets.
\begin{itemize}
  \item 3DMatch \citep{zeng20173dmatch} and 3DLoMatch \citep{li20214dcomplete}: These are datasets for rigid-body partial point cloud registration consisting of indoor scans. “Lo” in the name 3DLoMatch denotes the low level of overlap between the source and target point clouds. 3D Match is a collection of 62 scenes which are mostly scans of indoor environments. The official 3DMatch dataset only contains those examples that have more than 30 percent overlap between the source and target point clouds. 3DLoMatch is its counterpart that contains the scan pairs with 10-30 percent overlap. 
\begin{figure} [h!]
    \centering
    \includegraphics[width=0.7\textwidth]{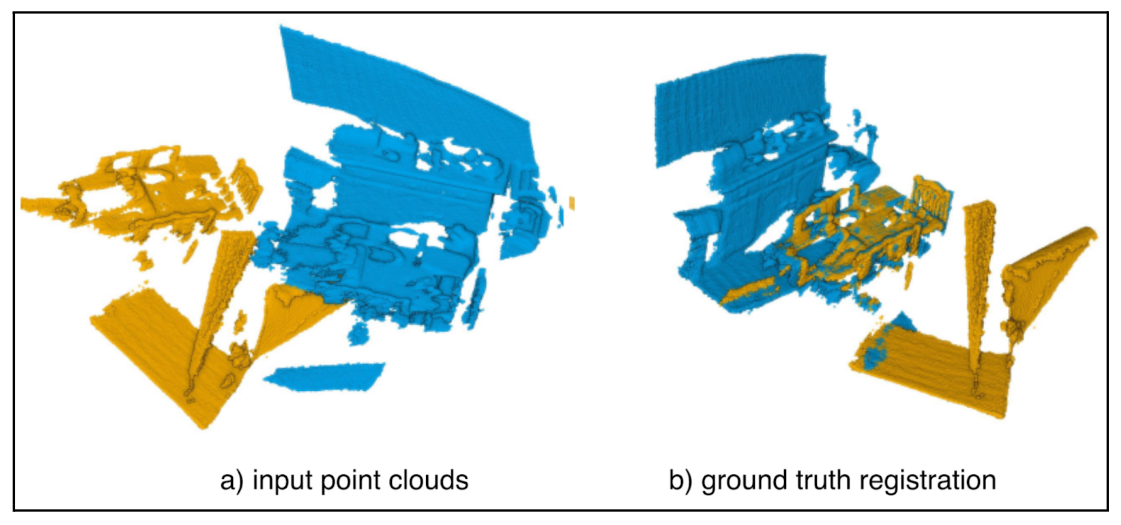}
    \caption{Kernel Sample data point from 3DMatch dataset.}
    \label{fig:3D-match}
\end{figure}

\begin{figure}[h!]
    \centering
    \includegraphics[width=0.7\textwidth]{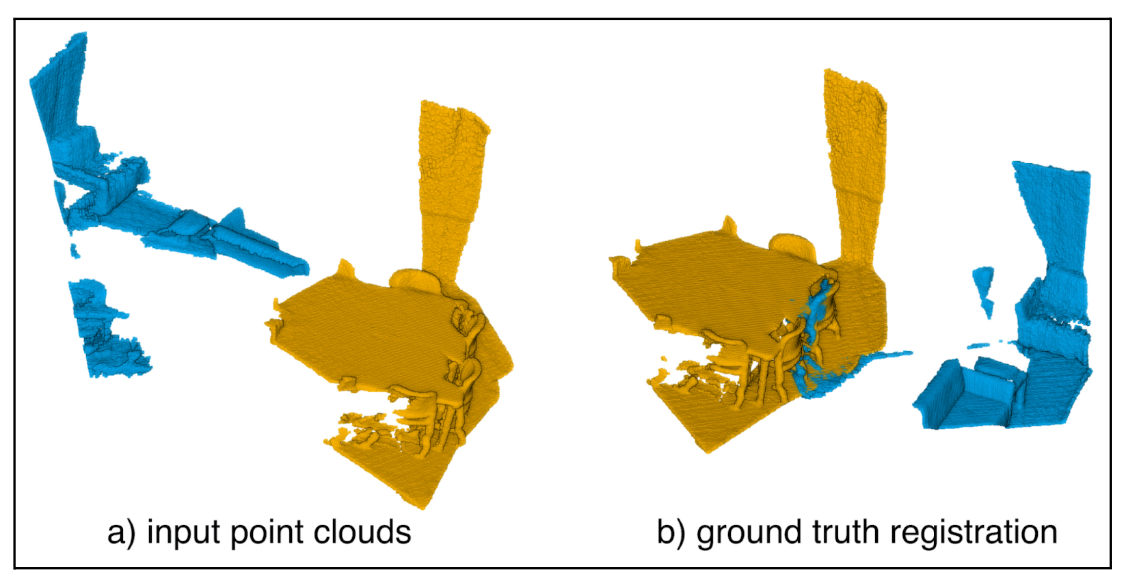}
    \caption{Sample data point from 3DLoMatch dataset.}
    \label{fig:3D-LOMatch}
\end{figure}

\item 4DMatch and 4DLoMatch \citep{li2022lepard}: These are datasets for non-rigid partial point cloud registration derived from DeformingObjects4d \citep{li20214dcomplete}, a dataset of animated characters. “Lo” in the name 4DLoMatch denotes the low level of overlap between the source and target point clouds in the test set. The DeformingObjects4d dataset has 1972 animation sequences with dense ground-truth correspondences between points. The 4DMatch dataset consists of a randomly selected subset of 1761 animation sequences from DeformingObjects4d. The training, validation and test splits of this dataset are of sizes 1232, 176 and 353 respectively. The 4DLoMatch dataset has the same training and validation splits as the 4DMatch dataset. The 353 test samples are distributed between 4DMatch and 4DLoMatch as follows: 4DLoMatch has all the samples that have an overlap ratio between the source and target point clouds of less than 45 percent. If the overlap ratio is 45 percent or higher, the sample is assigned to 4DMatch. Figure 8 shows a few examples from these datasets.
\begin{figure}[!h]
    \centering
    \includegraphics[width=\textwidth]{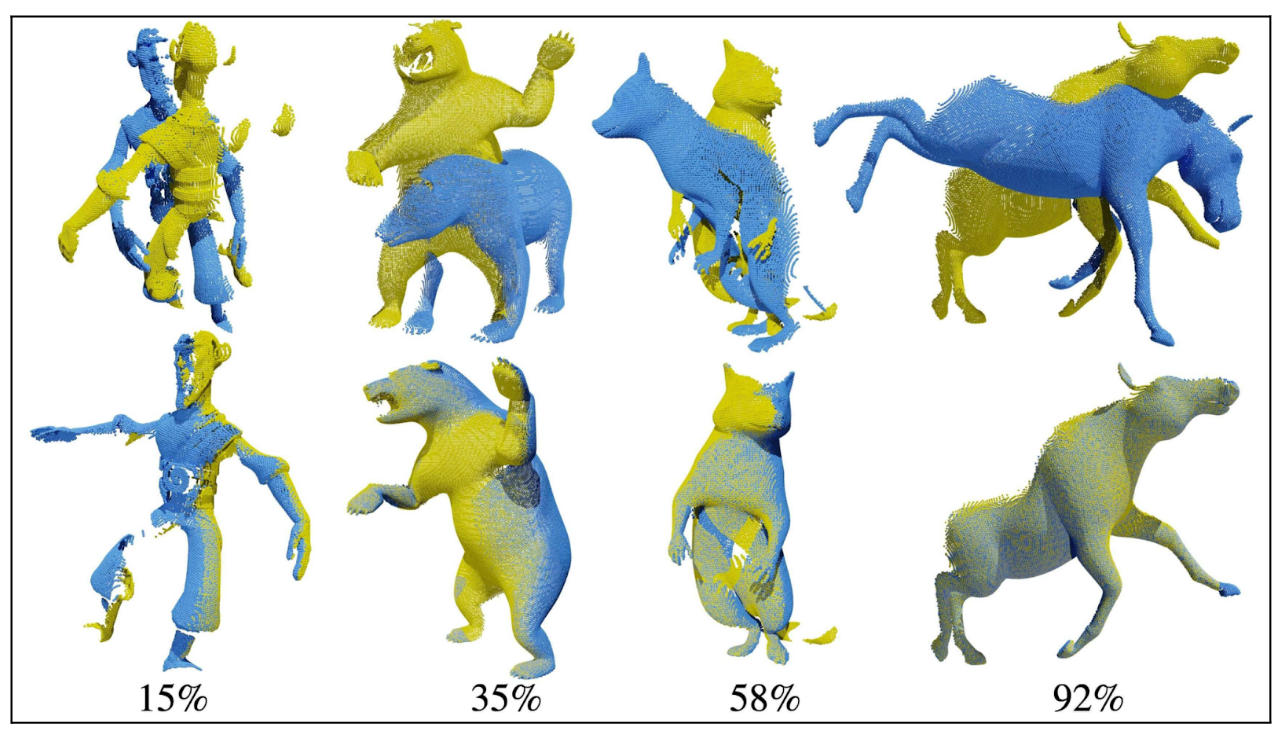}
    \caption{Samples from the 4DMatch/4DLoMatch dataset. The top row shows the input source and target point clouds and the bottom row shows the ground truth registration outcomes. The percentages at the bottom of the figure denote the percentage of overlap between the source and target point clouds in each of these examples.}
    \label{fig:4D-match / 4D- LoMatch}
\end{figure}

4DMatch and 4DLoMatch \citep{li2022lepard} are composed of animated objects. However, for real-world use cases in Mixed Reality, Robotics, etc. we must evaluate the performance of non-rigid registration on noisy scans of real deformable objects. We collect such data with our Microsoft HoloLens 2 MR device and Azure Kinect RGB-D camera. Figure 9 shows some test objects that we plan to use for evaluation in this project. Figures 10 and 11 show our test environment that is cluttered and represents a typical real-world usage environment of Mixed Reality devices.

\begin{figure}[!h]
    \centering
    \includegraphics[width=0.8\textwidth]{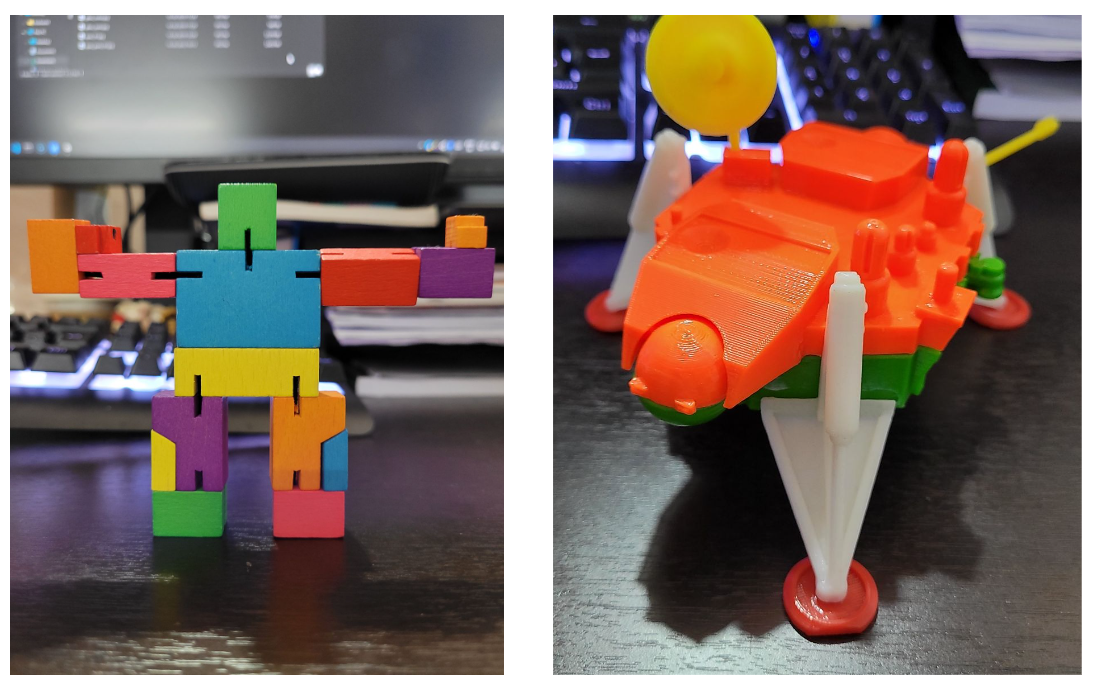}
    \caption{Real-world objects used for our study. (Left) Cubebot robot model (non-rigid). (Right) NASA Lander model (rigid)}
    \label{fig:Lander-ROBO}
\end{figure}

\begin{figure}[!h]
    \centering
    \includegraphics[width=0.6\textwidth]{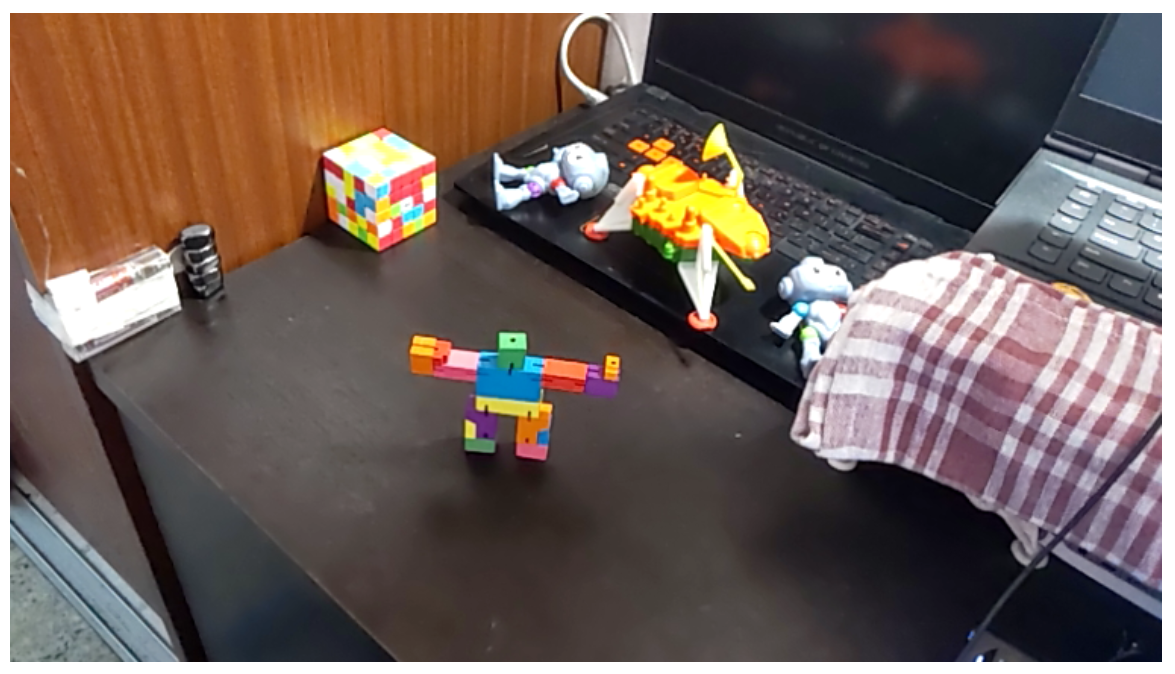}
    \caption{Real-world environment (cluttered) used in our tests.}
    \label{fig:real-world-cluttered}
\end{figure}

\begin{figure}[!h]
    \centering
    \includegraphics[width=0.6\textwidth]{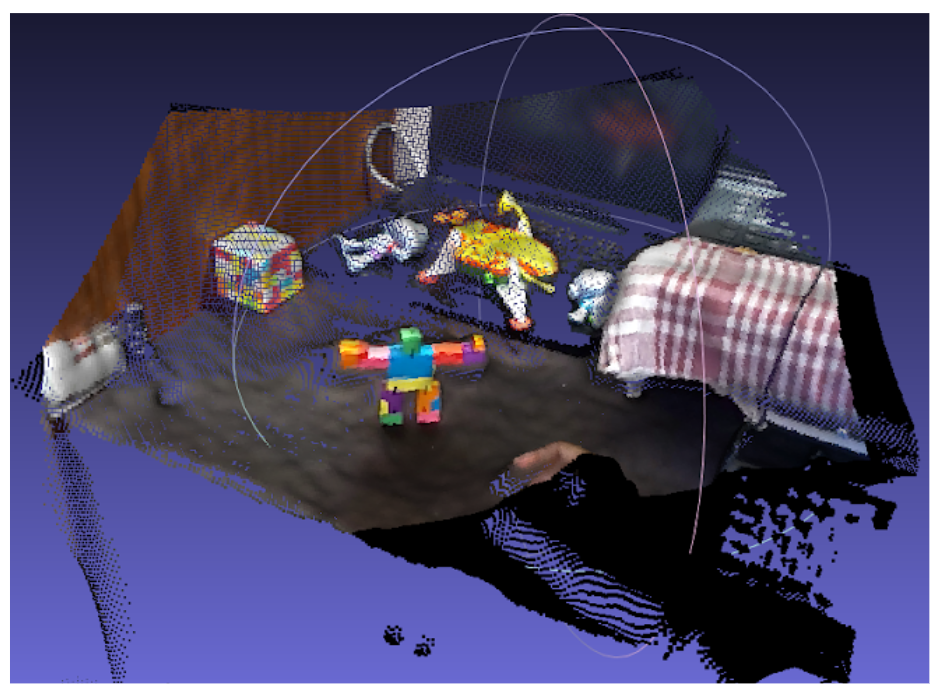}
    \caption{Point cloud scan of the real-world cluttered test environment. Scan generated using Microsoft HoloLens 2.}
    \label{fig:pc-cluttered}
\end{figure}

\end{itemize}

\section{Data Visualisation}
Visualization is extremely necessary for debugging and analyzing point cloud registration results. We use several tools for visualization. For debugging purposes, we use Matplotlib \citep{hunter2007matplotlib}, Open3D \cite{zhou2018open3d}, and Mayavi \citep{ramachandran2011mayavi} Python libraries within our workflow. For a detailed analysis of registration results, we use Cloud Compare \citep{girardeau2016cloudcompare} and MeshLab \citep{cignoni2011meshlab}.

\section{Training Objectives}
The following are the loss functions used to train Lepard \citep{li2022lepard}:

\begin{itemize}
  \item \textbf{Matching Loss: }The Focal Loss over the confidence matrix returned by the matching layer is minimised.
  \item \textbf{Warping Loss:} This is the L1 Loss between the target point cloud and the source point cloud warped by the Rotation and Translation matrices returned by the Procrustes layer.
\end{itemize}

\section{Evaluation Metrics}
The following are the metrics used for benchmarking point cloud registration algorithms:

\begin{itemize}
  \item \textbf{Inlier ratio (IR).}The inlier Ratio is the fraction of correct matches in the predicted correspondences set. 
  \begin{figure}[!h]
    \centering
    \includegraphics[width=0.9\textwidth]{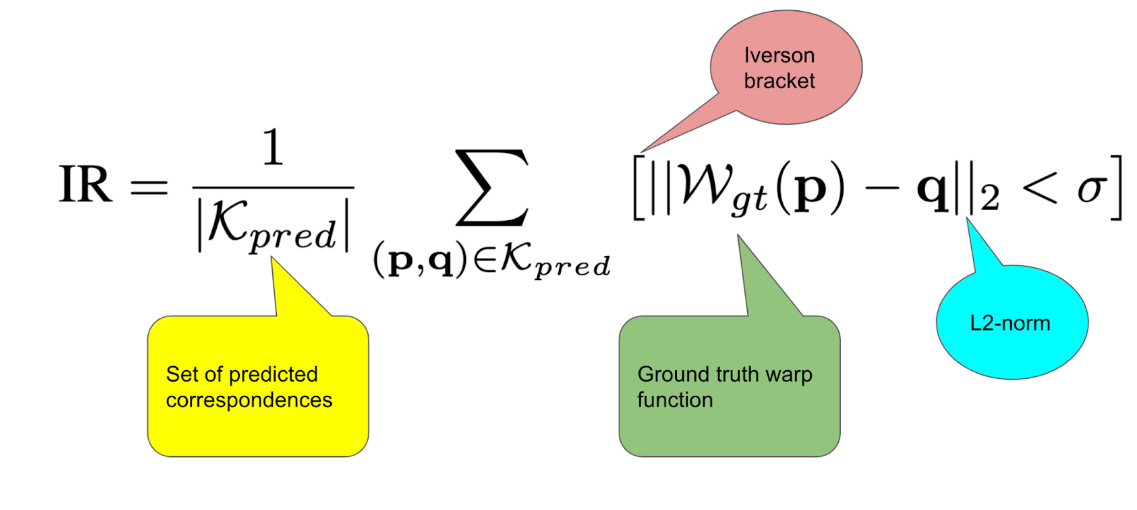}
    \caption{Illustration of the Inlier Ratio (IR) metric.}
    \label{fig:Inlier-ratio}
\end{figure}
  \item \textbf{Non-rigid Feature Matching Recall (NFMR).} NFMR measures the fraction of ground-truth matches that were successfully recovered by the correspondence prediction algorithm. 
    \begin{figure}[!h]
    \centering
    \includegraphics[width=\textwidth]{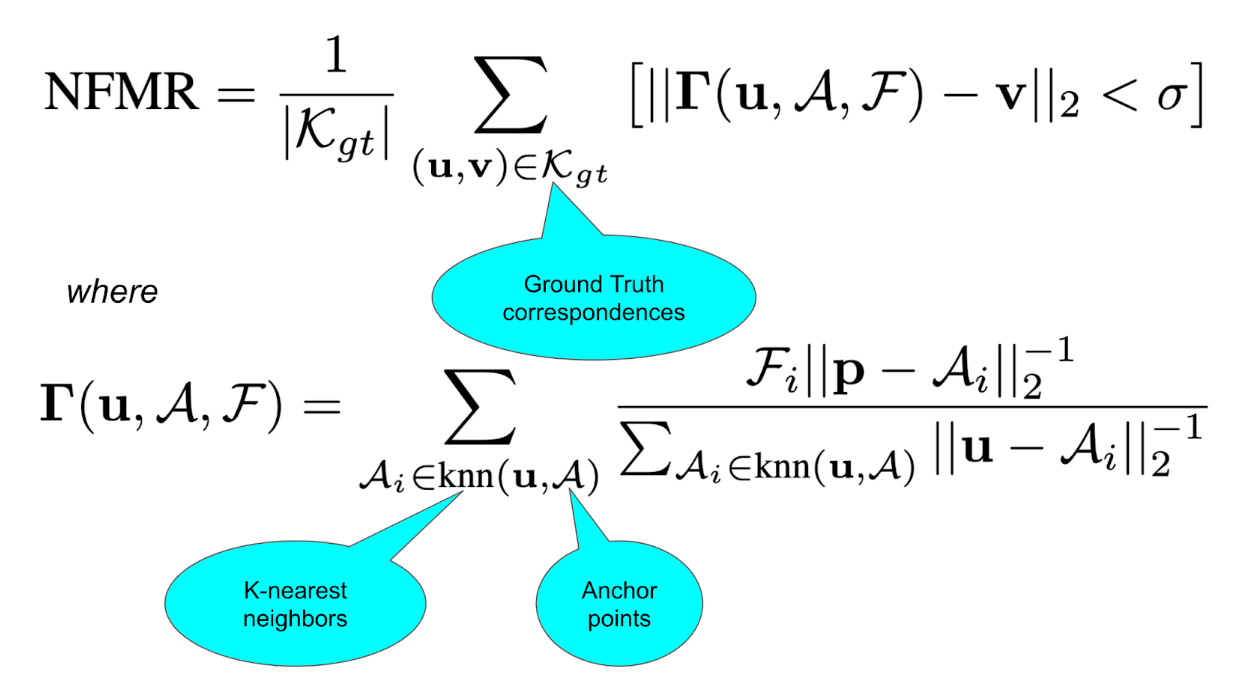}
    \caption{Illustration of the Non-rigid Feature Matching Recall (NFMR) metric}
    \label{fig:NFMR}
\end{figure}
\end{itemize}

\section{Baseline Selection}
We choose Lepard \citep{li2022lepard} as our baseline algorithm as it is one of the current state-of-the-art methods for partial non-rigid point cloud registration. The algorithm uses powerful KPConv [\citep{thomas2019kpconv}] for learning deformable point cloud representations. In order to add positional information to the translation invariant KPConv representation, the authors use Rotary Positional Encoding [93]. For point cloud matching it uses the RANSAC algorithm [\citep{holz2015registration}] for rigid objects and deformable ICP for non-rigid objects. A transformer is used for position-aware feature matching and Rigid Fitting with Soft Procrustes [\citep{besl1992method}] is used to obtain the final Rotation and Translation values for the transformation/warp function.
\begin{figure}[!h]
    \centering
    \includegraphics[width=\textwidth]{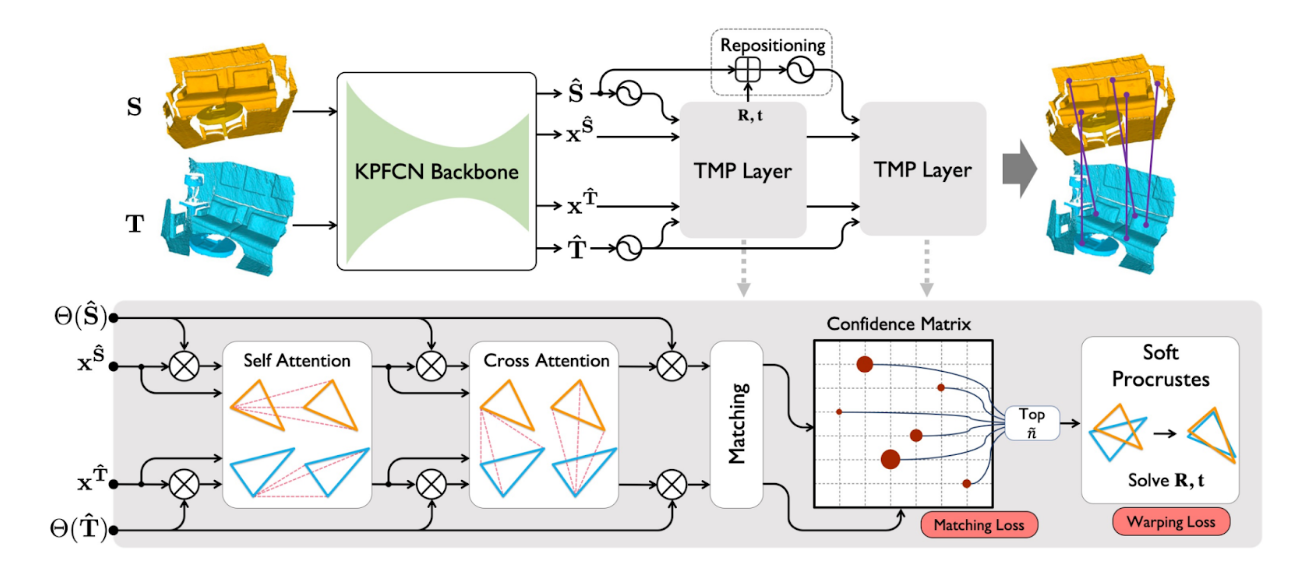}
    \caption{Flowchart depicting the architecture of Lepard. Please refer to the paper \citep{li2022lepard} for details.}
    \label{fig:arch-lepard}
\end{figure}

\section{Conclusion}
In this chapter, we described our research methodology. We provided key details about the datasets used for training and real world evaluation. Mixed reality is a compute intensive field. So we also disclosed the hardware available for our study. We described our baselines and explained the evaluation metrics.

%% file: Chapters/Chapter4.tex
\chapter{Non-Rigid Point Cloud Registration}
\label{ch:non_rigid_point_cloud_registration}
\addcontentsline{toc}{chapter}{Non-Rigid Point Cloud Registration}
\lhead{\emph{Non-Rigid Point Cloud Registration}}

In this chapter, we present a detailed account of the non-rigid point cloud registration workflow adopted in this research project. Our workflow is based on Lepard \citep{li2022lepard} and Iterative Closest Point (ICP) algorithms \citep{arun1987least,besl1992method,zhang1994iterative} that we discuss in details in this chapter. This will serve as a prelude to the novel methods that we present in the next chapter of the thesis.

\section{Introduction}
\label{s:non-rigid-intro}
We adopt a modular workflow similar to Lepard \citep{li2022lepard}. Modular workflows are more interpretable and easily upgradable than end-to-end workflows \citep{glasmachers2017limits}. There are three main steps in this workflow:
\begin{enumerate}
    \item Finding correspondences between ``feature points'' in the source and target point clouds.
    \item Whole body ``Rigid fitting'' using the correspondences discovered.
    \item Non-rigid registration to warp the surface of the source point cloud onto the target.
\end{enumerate}

The most challenging task in the entire workflow is finding the correspondences in the presence of arbitrary deformations, occlusions, and holes. Lepard \citep{li2022lepard} is a state-of-the-art algorithm for finding non-rigid correspondences. We use the same in our pipeline. Lepard provides two options to accomplish the second step - RANSAC \citep{holz2015registration,schnabel2007efficient} and soft-prucrustes \citep{arun1987least}. It is important to note that non-rigid deformations are applied to the source point cloud only at the final step. The authors of Lepard leveraged the Non-rigid Iterative Closest Point algorithm \citep{li2008global,zollhofer2014real} for finding a rigid transformation for each point on a mesh representation of the surface of the source point cloud. We discuss these components in detail in this section.

\section{Lepard: Learning partial point cloud matching in rigid and deformable scenes}

Lepard is a point cloud matching algorithm that has been shown to be effective even when the source and target point clouds are noisy, and incomplete and the degree of overlap is low. It addresses the challenge of disambiguation between repetitive geometry patterns - for example, the two hands of a human, and multiple instances of identical chairs in a room. Prior methods used the geometry features extracted by 3D Convolutional Neural Networks (3D-CNNs) \citep{singh20193d} that are designed to be translation invariant like other Convolutional Neural Networks. These representations also tend to be rotation invariant due to the use of max-pooling layers in the network and data augmentation using random rotations of the same input during training. The authors observe that rotation and transformation invariant features can cause repetitive geometry patterns to be assigned identical representations. So left and right hands of the same person would look the same after embedding. All the chairs of the same kind in the scene would also be difficult to distinguish in the embedding space. Motivated by the tendency of humans to associate things not just their appearance but also by their relative positions, the authors propose to enhance the 3D geometry representations with 3D positional information to help with disambiguation. 

\subsection{Notation}

We use the same notation as the authors of Lepard in our paper. Let $\mathbf{S} \in \mathbb{R}^{n\times 3}$ be the source point cloud (with $n$ points) and $\mathbf{T} \in \mathbb{R}^{m\times 3}$ be the target point cloud (with $m$ points). Our goal is to calculate a warp function $\mathcal{W}: \mathbb{R}\rightarrow \mathbb{R}$ that maps $\mathbf{S}$ to $\mathbf{T}$. For an arbitrary non-rigid deformation, $\mathcal{W}$ generalizes to a dense per-point warp field. For the rigid case, $\mathcal{W}$ can be parameterized as a $SE(3)$ transformation which is a single rotation and a single transformation applied to the whole body of $\mathbf{S}$. 

\subsection{Local geometry feature extraction}

Lepard uses Kernel Point Convolution \citep{thomas2019kpconv} based on a Fully Convolutional Neural Network (KPFCN) for representation learning. The KPFCN backbone uniformly downsamples both $\mathbf{S}$ and $\mathbf{T}$ for efficient computation. This is particularly crucial for the transformer and matching modules that come next. Both of them have $\mathcal{O}(n^2)$ time complexity. Again, following the notation used by the authors of Lepard, we denote the downsampled source and target point clouds by $\hat{\mathbf{S}}$ and $\hat{\mathbf{T}}$ respectively. These go as input to the KPFCN network for feature extraction. We discuss KPFCN next.

\subsubsection{Kernel Point Convolution}

The remarkable progress of image recognition algorithms in the past decade has been fuelled in a large part by Convolutional Neural Networks (CNNs) \citep{rawat2017deep}. These networks leverage the grid structure of 2D images for high comput efficiency on modern hardware like GPUs. Inspired by this success, many 3D convolutional network based algorithms were proposed for processing point cloud scans. However, point clouds do not come in a 3D grid structure. Hence 3D-CNN methods had to synthetically put the points in a regular grid structure resulting in the loss of critical structural information. The was particularly amplified in the case of sparse clouds and varying density. The authors of Kernel Point Convolution (KPConv) \citep{thomas2019kpconv} were motivated by the need to work around the grid assumption and proposed a flexible, deformable convolution operator that retains the original structure of the point cloud (spatial localization). 

We formally introduce KPConv using notation from the paper \citep{thomas2019kpconv}. Let us consider a point cloud $\mathbf{P} \in \mathbb{R}^{n\times 3}$ with point features $\mathcal{F} \in \mathbb{R}^{n\times D}$. Let $f_i \in \mathcal{F}$ denote the feature vector representing point $x_i \in \mathbf{P}$. KPConv consists of a set of local 3D filters. For a kernel $g$, the normal convolution operation works as shown in Equation \ref{eq:normal_conv}.

\begin{equation}
    (\mathcal{F}* g)(x) = \sum_{x_i \in \mathcal{N}(x)} g(x_i - x)f_i
\label{eq:normal_conv}
\end{equation}

Where $\mathcal{N}(x)$ denotes a neighborhood of the point $x$, a region where the kernel $g$ is applied. The authors note that a spherical neighborhood is more effective against varying density than a K-nearest neighbor (KNN) neighborhood. So, $\mathcal{N}(x) = \{x_i \in \mathbf{P} \mid ||x_i - x|| \le r\}$ where $r\in \mathbb{R}$ is the radius of the neighborhood. The authors call this neighborhood $\mathcal{N}(x)$ centered at $x$, the domain of definition of kernel $g$. All points in $\mathbf{P}$ that lie within this domain of definition will be influenced by $g$. However, we would like to have different amounts of influence on different points depending on their distance from $x$. Writing relative distance from $x$ as $y_i = x_i - x$, we have the domain of definition of $g$ as $\mathcal{B}_r^3 = \{y\in \mathbb{R}^3 \; \mid\; ||y|| \le r\}$. Each kernel $g$ has $K$ kernel points $\{\Tilde{x}_k\}_{k=0}^{K-1} \subset \mathcal{B}_r^3$. Each kernel point $\Tilde{x}_k$ is assigned a weight $W_k \in \mathbb{R}^{D\times D_{out}}$. For any point $y\in \mathcal{B}_r^3$ the kernel function $g$ is defined as in equation \ref{eq:kpconv_def}.

\begin{equation}
    g(y) = \sum_{k=0}^{K-1} h(y, \Tilde{x}_k) W_k 
\label{eq:kpconv_def}
\end{equation}

Here $h$ is the correlation between the features of the point $y$ and the kernel point $\Tilde{x}_k$. Points closer to $\Tilde{x}_k$ will likely have a higher correlation with $\Tilde{x}_k$ and hence will be influenced more by the corresponding weight $W_k$. The authors compute the correlation function using the linear formulation in equation \ref{eq:correlation_def}.

\begin{equation}
    h(y, \Tilde{x}_k) = \max \left( 0, 1 - \frac{||y-\Tilde{x}_k||}{\sigma} \right)
\label{eq:correlation_def}
\end{equation}

KPConv comes in two variants - rigid and deformable. In rigid KPConv, the kernel points are fixed. In deformable KPConv, the kernel points have learnable offset parameters $\Delta_k(x)$. Deformable KPConv is able to adjust the shape of the kernel and adapt the kernel points to the local geometry.

\subsection{Relative 3D Positional Encoding}

Representations learned by the KPFCN backbone are translation and rotation invariant due to reasons discussed before. In order to disambiguate between repetitive geometry features, the authors of Lepard propose to enhance the KPFCN features using 3D positional encoding. 

The most popular kind of positional encoding method is sinusoidal positional encoding \citep{vaswani2017attention}. A set of $M$ discrete sinusoids and co-sinusoids at different frequencies are defined over the size of the input space and these are sampled at each point $x$ to obtain a $2M$ dimensional encoding that reflects the relative position of the point. The authors of Lepard chose to use the Rotary Positional Encoding scheme of the Roformer paper \citep{su2021roformer}. Given a 3D point $S_i\in \mathbb{R}^3$ with feature vector $x_i^S \in \mathbb{R}^D$, the positional encoding function is defined as in equation \ref{eq:roformer-eqn}.

\begin{equation}
    PE(S_i, x_i^S) = \Theta(S_i)x_i^S
\label{eq:roformer-eqn}
\end{equation}

Where $\Theta(S_i)$ is a block diagonal matrix.

The authors argue that it is a better choice than sinusoidal in the case of point clouds because:
\begin{enumerate}
    \item The embedding only changes the feature's direction, not the feature's length. This can potentially improve the stability of training.
    \item The dot product of two spherical embedding feature vectors explicitly reveal the Euclidean distance between the corresponding points as demonstrated in equation \ref{eq:positional_encoding_dot_product}.
    
    \begin{equation}
        PE(S_i, x_i^S)^T PE(S_j, x_j^S) = [\Theta(S_i)x_i^S]^T\Theta(S_j)x_j^S = (x_i^S)^T\Theta(S_j-S_i)x_j^S
    \label{eq:positional_encoding_dot_product}
    \end{equation}
    
    In the last step $\Theta(S_i)^T\Theta(S_j) = \Theta(S_j-S_i)$ is a result of the construction of the matrix $\Theta$.
\end{enumerate}

\subsection{Transformer}

Let us represent the outcome of local geometry feature extraction as $x^{\Tilde{S}}$ and $x^{\Tilde{T}}$ for the source and target point clouds respectively. Lepard uses a transformer network to compute the self attention within the points of the same cloud and cross attention between the points in the source and target clouds. While self-attention collects global context, cross attention helps in the exchange of information between the source and target clouds.

Attention mechanism works by computing query ($\mathbf{q}$), key ($\mathbf{k}$) and value ($\mathbf{v}$) vectors using learnable projection matrices $W_q$, $W_k$ and $W_v$ for each point in the point cloud as follows: 

\begin{eqnarray}
    q_i =& \Theta(\hat{S}_i)W_q x_i^{\hat{S}} \\ \notag
    k_i =& \Theta(\hat{S}_i)W_k x_i^{\hat{S}} \\ \notag
    v_i =& W_v x_i^{\hat{S}} \\ \notag
\end{eqnarray}

\subsubsection{Self Attention Layer}

In the self attention layer, the features of each point are updated as in equation

\begin{equation}
    x_i^{\hat{S}} = x_i^{\hat{S}} + MLP(\text{cat}[q_i, \sum_j a_{ij}v_j])
\end{equation}

where $a_{ij} = softmax(q_i k_j^T / \sqrt D)$ is the attention weight, $MLP$ is a $3$-layer multi-layer perceptron and $\text{cat}$ is the concatenation operator.

\subsubsection{Cross Attention Layer}

The mathematical construct of the cross attention layer is identical to the self attention layer. The only difference is that the query $q$ and the key-value pair $(k, v)$ come from different point clouds. Lepard computes cross attention both ways: $\hat{S} \rightarrow \hat{T}$ and $\hat{T} \rightarrow \hat{S}$.

\subsection{Position-aware Feature Matching}

The transformer layer is followed by feature matching. The scoring matrix $\mathcal{S}$ between the source and target point clouds is computed as in equation \ref{eq:feature-matching}.

\begin{equation}
    S(i,j) = \frac{1}{\sqrt D} \langle \Theta(\hat{S}_i)W_{\hat{S}}x_i^{\hat{S}}, \; \Theta(\hat{T}_j)W_{\hat{T}}x_j^{\hat{T}}\rangle
\label{eq:feature-matching}
\end{equation}

Here, $\langle , \rangle$ denotes inner product. $W_{\hat{S}}$, $W_{\hat{T}} \in \mathbb{R}^{D\times D}$ are learnable projection matrices. The use of point representations with positional encoding makes sure that the matching algorithm takes spatial distance into account. The scoring matrix is converted into a confidence matrix $\mathcal{C}$ using a dual-softmax operation as shown in equation \ref{eq:dual-softmax}.

\begin{equation}
  \mathcal{C}(i, j) = \text{Softmax}(\mathcal{S}(i, \cdot)) \cdot \text{Softmax}(\mathcal{S}(\cdot, j))
\label{eq:dual-softmax}
\end{equation}

The confidence scores are thresholded at a given minimum value $\theta_c$ to determine matches. $\theta_c$ is a hyperparameter.

\subsection{Rigid Fitting with Soft Prucrustes}
\label{sec:soft-procrustus}

Once the correspondence matches between the points in $\hat{S}$ and $\hat{T}$ have been found, a rigid transformation (consisting of a rotation matrix $R\in SO3$ and a translation vector $t\in \mathbb{R}^3$) is derived to fit this correspondence has closely as possible. This process is known as rigid fitting with Soft Prucrustes \citep{arun1987least}. From the correspondence matrix $\mathcal{C}$, the top $\hat{n}$ scoring matches are chosen. Let us call this set $\mathcal{K}_{soft}$. The next step is to construct the matrix $\mathcal{H}$ as in equation \ref{eq:matching-H}.

\begin{equation}
    H = \sum_{(i,j)\in \mathcal{K}_{soft}} \Tilde{C}(i,j) \hat{S}_i \hat{T}_j^T
\label{eq:matching-H}
\end{equation}

$\Tilde{C}(i,j)$ is the normalized confidence score. Next, rotation $R$ is computed as in equation \ref{eq:matching-R}.

\begin{equation}
    R = U \text{diag} (1, 1, det(UV^T))V
\label{eq:matching-R}
\end{equation}

$U$ and $V$ are obtained from an SVD decomposition of $H = U\Sigma V^T$. The translation $t$ is obtained as in equation \ref{eq:matching-t}.

\begin{equation}
    t = \frac{1}{|\mathcal{K}_{soft}|} \left( \sum_{(i,\cdot)\in \mathcal{K}_{soft}} \hat{S}_i - R \sum_{(\cdot, j) \in \mathcal{K}_{soft}} \hat{T}_j\right) 
\label{eq:matching-t}
\end{equation}



\section{Non-rigid Registration}

After rigid fitting with the $R$ and $t$ matrices derived in the previous section, non-rigid registration is performed for warping the surface of the source point cloud on to the target. Non-rigid Iterative Closest Point (N-ICP) algorithm \citep{li2008global,li20214dcomplete} is one of the common approaches to this. There are two categories of registration algorithms:

\begin{itemize}
    \item \textbf{Dense registration:} Dense registration methods find a mapping from each point in the template on to the target.
    \item \textbf{Sparse registration:} Sparse registration methods find correspondences only for selected feature points.
\end{itemize}
N-ICP is a dense registration algorithm.

\subsection{Non-rigid Iterative Closest Point (N-ICP)}

Registering two surfaces means finding a mapping between a template surface and a target surface that describes the position of semantically corresponding points. This is also known as warping the template onto the target. The task of a registration algorithm is to choose the correct deformation from all possible warps by imposing constraints on the deformation. This is known as “regularisation of the deformation field”. 

N-ICP\citep{li2008global,li20214dcomplete} applies a locally affine regularisation that assigns an affine transformation to each vertex and minimises the difference in the transformation of neighbouring vertices. With this regularisation, the optimal deformation for fixed correspondences and fixed stiffness can be determined exactly and efficiently. The algorithm loops over a series of decreasing stiffness weights that results in incremental deformation of the template surface towards the target. In order to determine the optimal deformation for a given stiffness:

\begin{enumerate}
    \item Preliminary correspondences are estimated by a nearest point search.
    \item The optimal deformation of the template for these fixed correspondences and the active stiffness is then calculated.
    \item Go to Step 1 to find new correspondences by searching from the displaced template vertices.
\end{enumerate}

In the ideal case, for noiseless and complete data, the correct registration should have a one-to-one correspondence. But in practice, the surfaces contain holes and artefacts resulting from the scanning process.
A useful registration method needs to be robust against outliers and must fill in missing data in a sensible way. Missing data needs to be filled in using knowledge from the template mesh. N-ICP achieves this by smoothly deforming the template mesh.

\section{Conclusion}
In this chapter, we described key non-rigid registration algorithms that are used in our pipeline. We presented Lepard \citep{li2022lepard} for non-rigid correspondence matching and N-ICP for non-rigid registration. We also pointed out the challenges involved in applying these methods in a real world mixed reality application and motivated the methods proposed in the following chapter for the use case of our focus.

%% file: Chapters/Chapter5.tex
\chapter{Proposed Method}
\label{ch:proposed_method}
\addcontentsline{toc}{chapter}{Proposed Method}
\lhead{\emph{Proposed Method}}

\section{Introduction}
In the previous chapter, we discussed the non-rigid point cloud registration workflow. We discussed how Lepard \citep{li2022lepard} provides an effective method of finding correspondences between point clouds undergoing non-rigid deformations. We also discussed how correspondence matching step is followed by a non-rigid registration step where methods like N-ICP \citep{holz2015registration,schnabel2007efficient} are used to warp a mesh representation of the source point cloud on to the target point cloud taking into account noise, holes and missing parts. Our contributions in the thesis are restricted to the non-rigid registration step and they are tailor-made for an important task in Mixed Reality applications which we discuss in this chapter.

\section{Motivation}

In N-PCR algorithms like Lepard, after rigid fitting, dense registration methods like N-ICP \citep{li2008global,li20214dcomplete} are used to optimally morph the source mesh on the surface of the target point cloud. This involves finding a transformation for each point on the source mesh and the computational requirement can blow up for large meshes.

In this work we want to focus on the problem of registration of CAD models to 3D scans of rigid objects with rigid movable parts. Some examples are machine parts with hinges and appliances with movable antennas. As we manipulate such objects, the movable parts may move relative to the object resulting in a non-rigid deformation. Most applications of non-rigid registration in enterprise mixed reality involve such objects. 

For the objects of interest, we can safely assume that the parts do not deform within themselves. Under this assumption, the non-rigid adjustment step after rigid fitting boils down to finding a rigid transformation for each part that can move relative to one another.

\section{Challenges}
There are several challenges being addressed by the proposed research. 
\begin{enumerate}
  \item The first is to make sure that the object model does not disintegrate after applying part-wise registration. 
  \item The second challenge is to search for a given part in the right part of the target point cloud. 
  \item The third challenge is to take into consideration the presence of outliers. For small parts, outliers can have a large impact on the predicted correspondences.
\end{enumerate}

\section{Our approach}
\begin{figure}[!h]
    \centering
    \includegraphics[width=\textwidth]{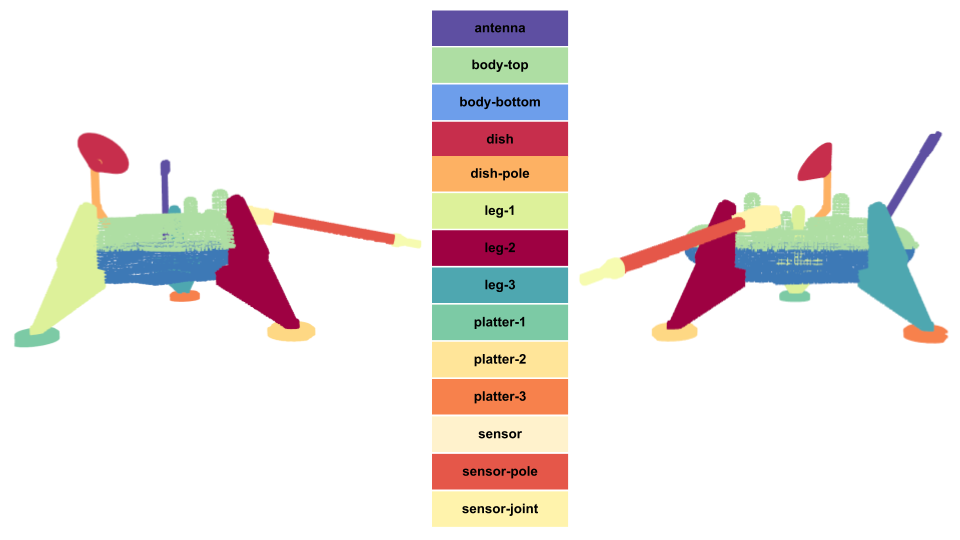}
    \caption{NASA Viking Lunar Lander with parts labelled. We show two views to reveal the 3D shape of the model.}
    \label{fig:Viking-lander-labelled}
\end{figure}

\begin{figure}[!h]
    \centering
    \includegraphics[width=\textwidth]{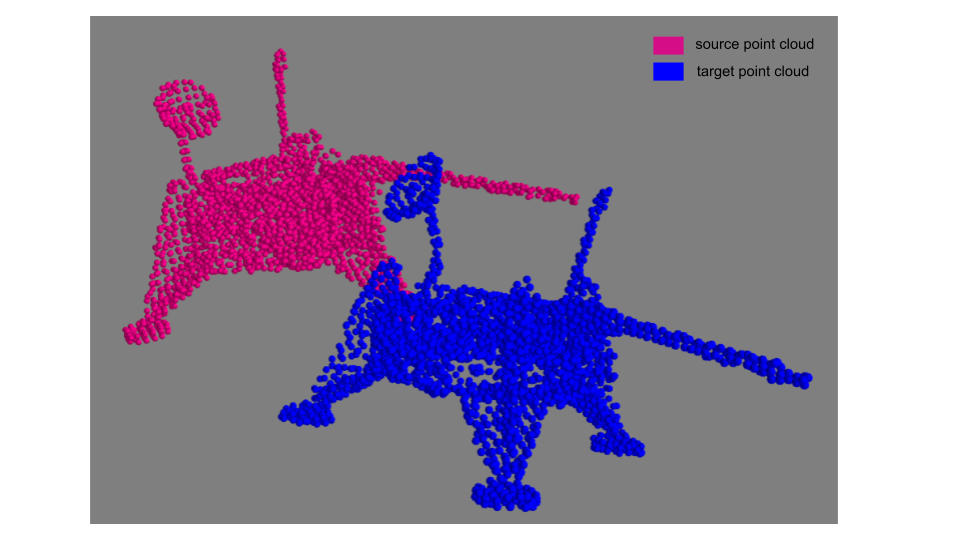}
    \caption{Source and target point clouds in the example that we use to describe our workflow in this section.}
    \label{fig:source-and-target-before-rigid-fitting}
\end{figure}

In this section, we discuss our proposed algorithm for non-rigid registration of rigid objects with rigid moving parts. We present our psuedocode of our proposed method in Algorithm \ref{alg:pseudocode}. We explain our algorithm with an example registration task on a model of the NASA Lunar Lander\footnote{Open source model of NASA Viking Lander obtained from \url{https://nasa3d.arc.nasa.gov/detail/viking-lander}} shown in Figure \ref{fig:Viking-lander-labelled}. 
The source and target point clouds are shown in figure \ref{fig:source-and-target-before-rigid-fitting}.
 In addition to a whole body rigid translation and rotation, the deformable parts \textit{dish}, \textit{dish-pole}, \textit{sensor}, \textit{sensor-pole}, \textit{sensor-joint} and \textit{antenna} have moved relative to the static components of the model in the target point cloud. The algorithm proceeds as explained below.

\begin{algorithm}
\caption{Pseudocode of part-whole registration.}
\label{alg:pseudocode}

\hspace*{\algorithmicindent} \textbf{Input}: Source point cloud $\mathbf{S}$ with its parts $\{p_i^S\}_{i=1}^{n_{parts}^S}$ labelled; Target point cloud $\mathbf{T}$; Retention fraction $f_{retention}^{subsample}$ for point-cloud subsampling; Maximum correspondence distance $d_{max}^{corr}$ for ICP; Minimum number of correspondences $n_{min}^{corr}$ for RANSAC.\\

\hspace*{\algorithmicindent} \textbf{Output}: A rigid transformation $(R_i, t_i)$ for each part $p_i^S$ in the source point cloud.\\

\begin{algorithmic}[1]

\State Ingest $S$ and $T$ and subsample them, retaining $f_{retention}^{subsample}$ fraction of the samples in order to fit in the GPU memory. Let the subsampled point clouds be $\hat{S}$ and $\hat{T}$.

\State Construct the source part-whole graph $G_{S}$. Estimate the bounding boxes for each part.

\State Run Lepard to discover the set of correspondences $\mathcal{K}$ between $\hat{S}$ and $\hat{T}$.

\State Derive whole body rigid transformation $(R_{whole}, t_{whole})$ from $\hat{S}$ to $\hat{T}$ using $\mathcal{K}$. (Section \ref{sec:soft-procrustus}). Apply $(R_{whole}, t_{whole})$ to $S$ to obtain $S'$ and to $\hat{S}$ to obtain $\hat{S}'$.

\State Sort the parts of $S$ in $G_{S}$ in the decreasing order of volume. Let the sorted list be $\{p_j^S\}_{j=1}^{n_{parts}^S}$.

\For{$j = 0, 1, \dots n_{parts}$}

\State Segment out the part of $\hat{S}'$, that corresponds to $p_j^{S'}$ by extraction just the points that fall within the bounding box of $p_j^{S'}$. We need this step to filter out the potential feature points for $p_j^{S'}$. Let us call this set of feature points $p_j^{\hat{S}'}$ .

\State Estimate the region of interest in $T$ -- the set of points in $T$ that may correspond to $p_j^{S'}$. This set is determined by a bounding box that is the smallest one to enclose all the points in $T$ that correspond to feature points in $p_j^{\hat{S}'}$ and the whole of $p_j^{\hat{S}'}$. Let the set of correspondences for this part be $\mathcal{K}_j$.

\While{User not satisfied}
\State Run RANSAC using the correspondences $\mathcal{K}_j$ to obtain the first-level registration results for $p_j$ -- $(R_j, t_j)$.
\EndWhile

\While{User not satisfied}
\State Run ICP with $\mathcal{K}_j$ and the initial transformation set to $(R_j, t_j)$ from the previous step for second-level registration for part $p_j$. Record the output transformation -- $p_j$ -- $(R_j, t_j)$ for part $p_j$.
\EndWhile

\State Apply $(R_j, t_j)$ to part $p_j^{S'}$.

\EndFor

\State Return $\{(R_i, t_i)\}_{i=1}^{n_{parts}}$.

\end{algorithmic}
\end{algorithm}

\subsubsection{Graph representation of source}
\begin{figure}[!h]
    \centering
    \includegraphics[width=0.6\textwidth]{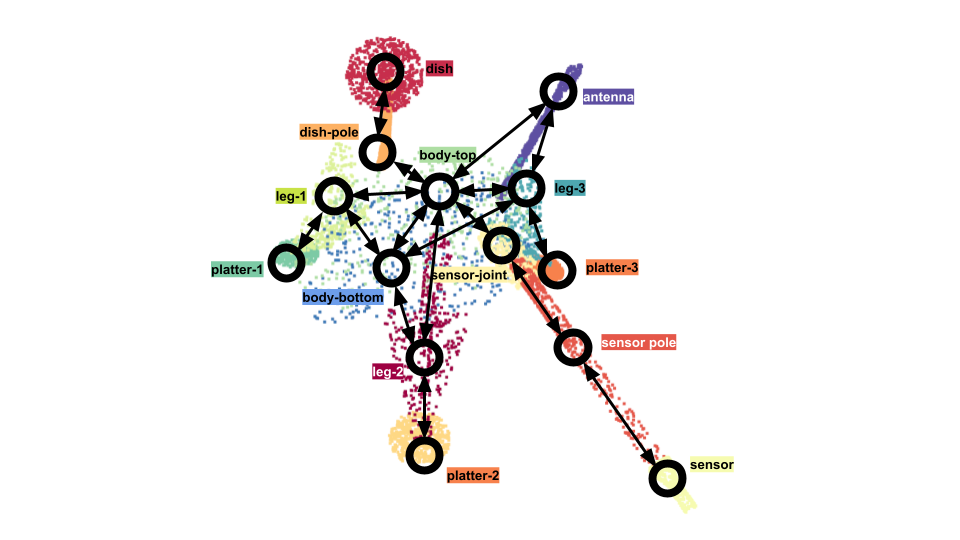}
    \caption{Graph Construction of NASA Viking Lunar Lander labelled}
    \label{fig:Graph-construction-Viking-lander-labelled}
\end{figure}

Our method begins by constructing a bidirectional graph of the source point cloud (which is available as a CAD model). Each node of this graph represents one of the rigid parts. Each pair of adjacent parts is connected with an edge. We illustrate this in figure \ref{fig:Graph-construction-Viking-lander-labelled}.

\subsubsection{Correspondence matching using Lepard}

\begin{figure}[!h]
    \centering
    \includegraphics[width=1.0\textwidth]{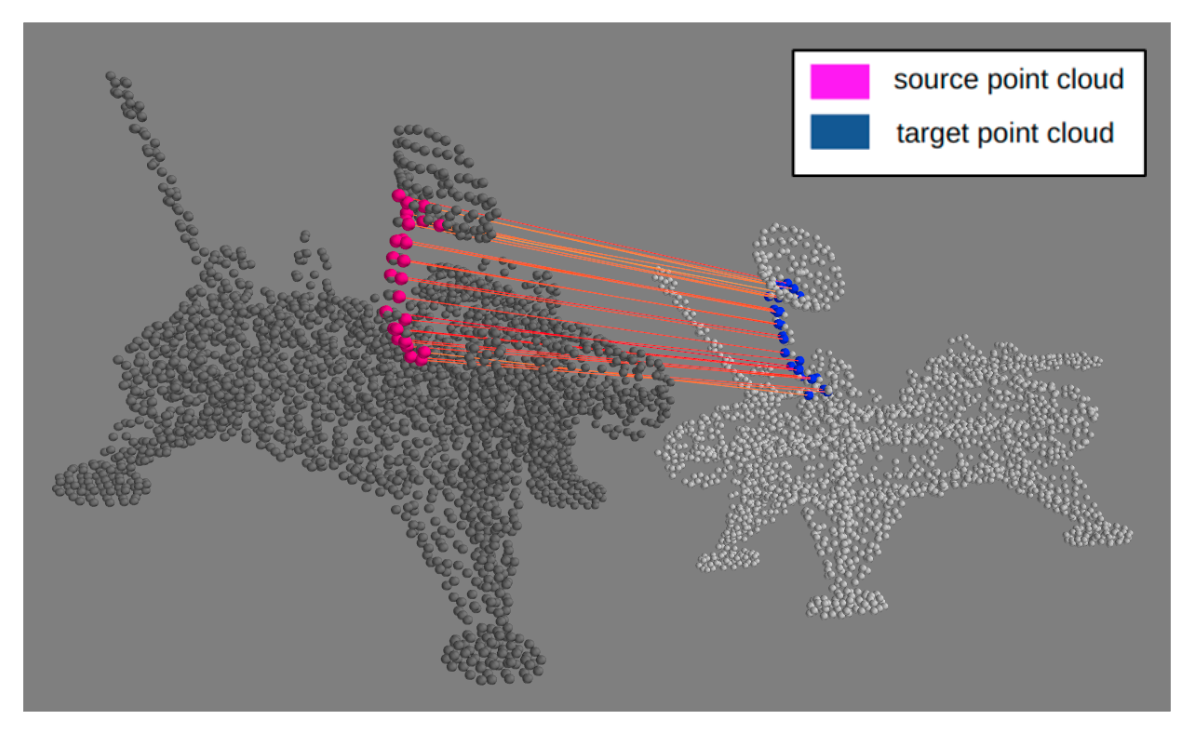}
    \caption{Correspondences of Dish-pole of NASA Viking Lunar Lander }
    \label{fig:Correspondences_Dish-pole}
\end{figure}

\begin{figure}[!h]
    \centering
    \includegraphics[width=0.9\textwidth]{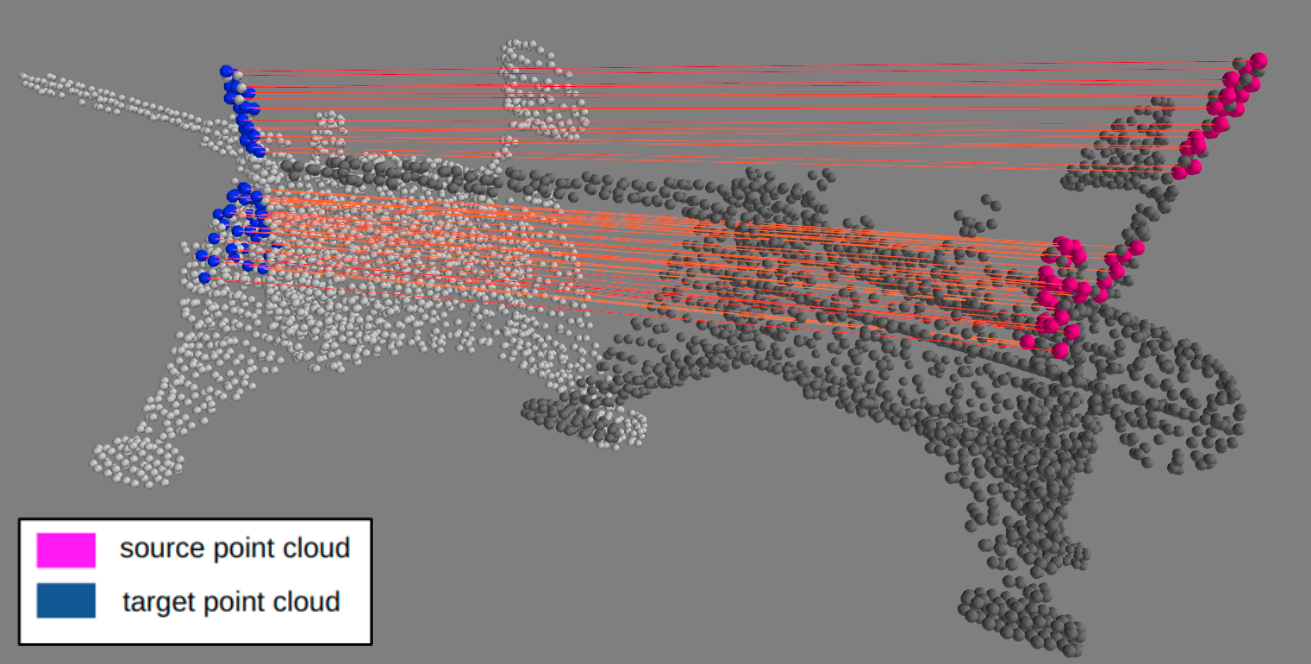}
    \caption{Correspondences of partial antenna and leg3 of NASA Viking Lunar Lander }
    \label{fig:Correspondences-antenna}
\end{figure}

\begin{figure}[!h]
    \centering
    \includegraphics[width=1.0\textwidth]{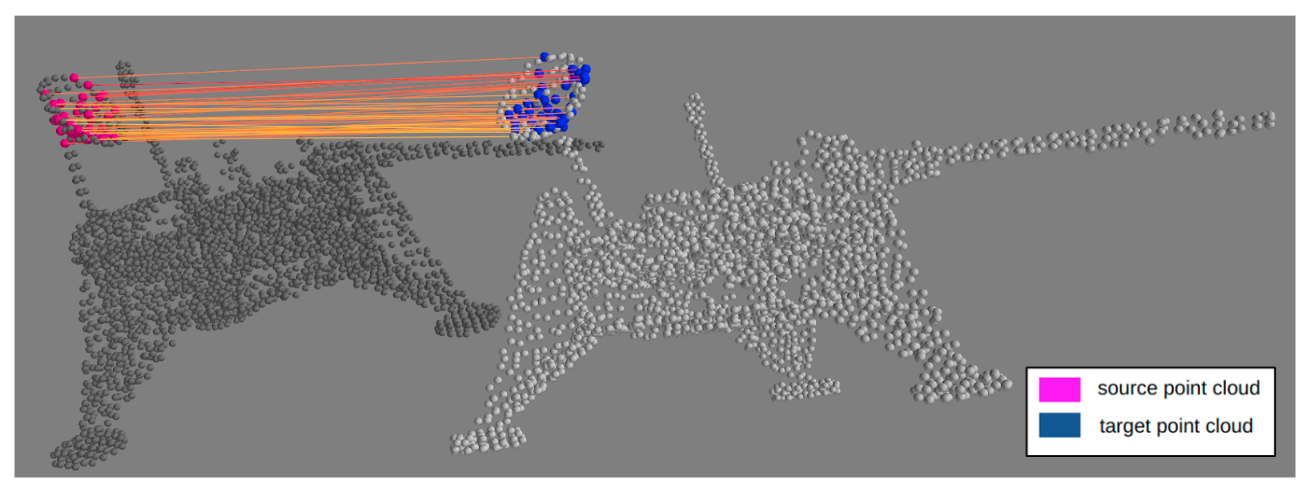}
    \caption{Correspondences of the dish of NASA Viking Lunar Lander }
    \label{fig:Correspondences-dish}
\end{figure}

\begin{figure}[!h]
    \centering
    \includegraphics[width=1.0\textwidth]{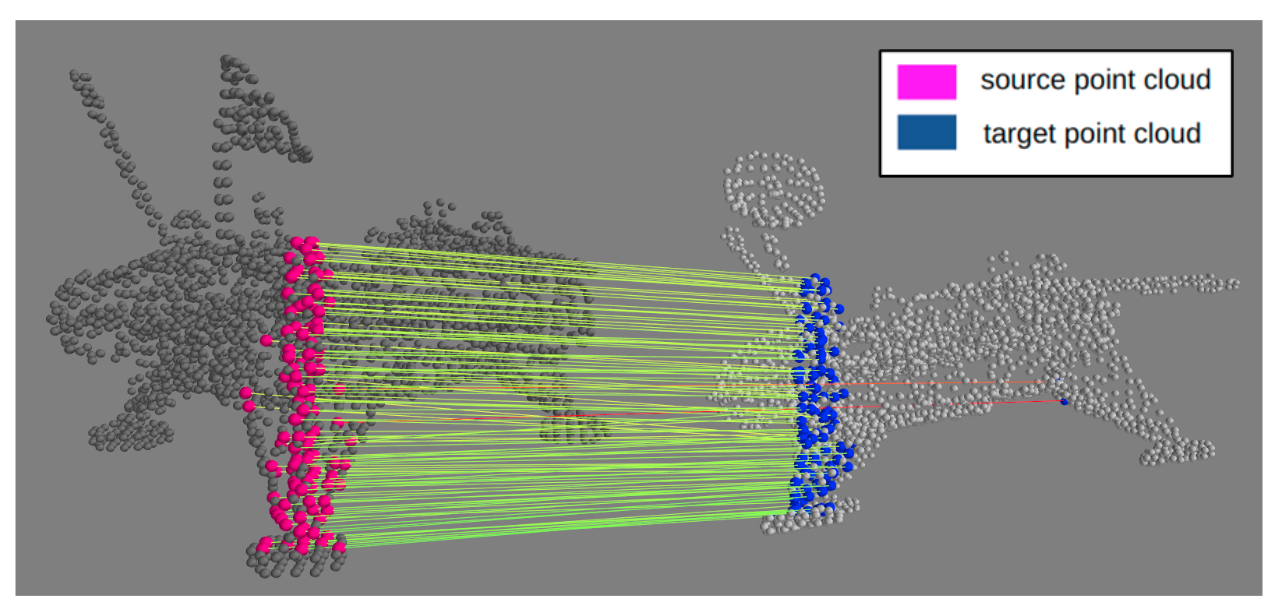}
    \caption{Correspondences of leg 1 of NASA Viking Lunar Lander }
    \label{fig:Correspondences-leg-1}
\end{figure}

In the next step, we run Lepard to obtain the corresponding points in the source and target clouds. Figures \ref{fig:Correspondences-antenna}, \ref{fig:Correspondences_Dish-pole}, \ref{fig:Correspondences-dish}, and \ref{fig:Correspondences-leg-1} show the correspondence matches obtained for some of the lander-parts.

\subsubsection{Whole body rigid fitting}

For rigid fitting, we derive the rigid body transformation for the whole source point cloud. 
\begin{figure}[!h]
    \centering
    \includegraphics[width=\textwidth]{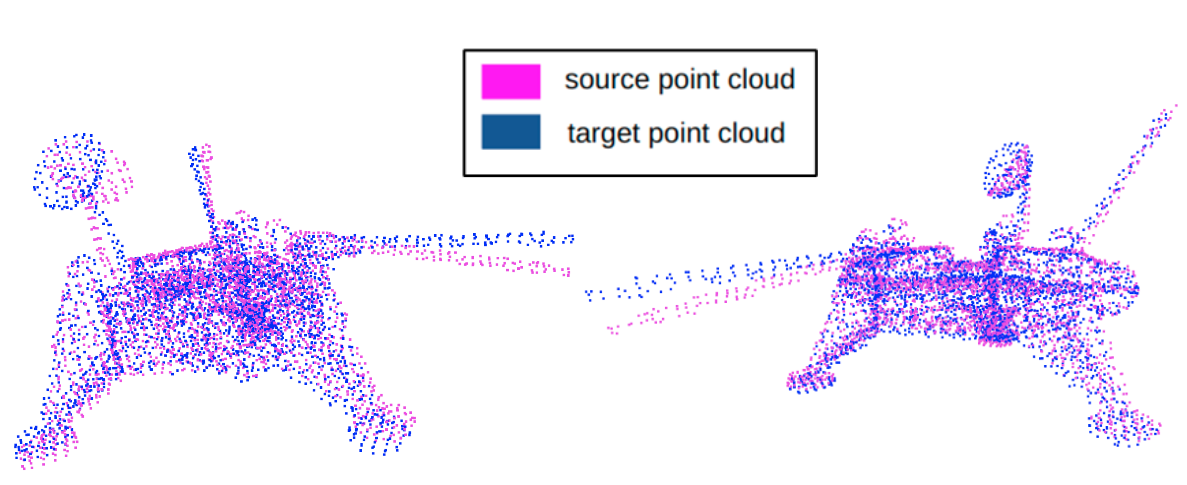}
    \caption{Point Cloud of NASA Viking Lunar Lander Source and target after Rigid fitting}
    \label{fig:source-and-target-after-rigid-fitting}
\end{figure}
Note that the parts that moved relative to the body resulting in a non-rigid deformation could not be aligned in this step.

\subsubsection{Part-wise tuning}

The next step involves part-wise tuning. We tune the parts in the reverse order of sizes. Each part is tuned in two steps - first using RANSAC \citep{} and the correspondences for the feature points in that part. The Lepard pipeline breaks the graph structure of the source point cloud during feature point extraction in the previous step. We recover the feature points for each part using the (axis-aligned) bounding boxes of the part. The output of the RANSAC algorithm is a rotation and translation for the part.
\pagebreak

\pagebreak

The second step in the tuning process uses ICP to tightly fit the part in its location in the target point cloud. The area of interest in the target point cloud is determined as an axis-aligned bounding box whose extremities cover the axis-aligned bounding box of the source part and all the corresponding points in the target point cloud for the feature points in the source part. Since ICP assumes that the source and target point clouds are spatially close, we initialize the ICP algorithm with the transformation obtained from RANSAC in the previous step. The output of this step is also a rotation and translation for the part in question.

\begin{figure}[!h]
    \centering
    \includegraphics[width=1.0\textwidth]{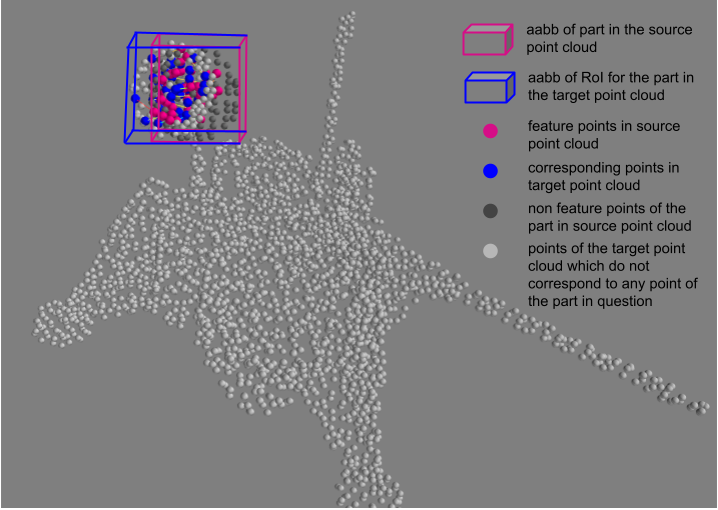}
    \caption{Diagram showing how we compute the region of interest }
    \label{fig:Diagram showing how we compute the region of interest}
\end{figure}
\pagebreak

We observe that failure to determine the correct region of interest in the target point cloud can adversely affect the performance of the ICP algorithm.
\begin{figure}[!h]
    \centering
    \includegraphics[width=0.8\textwidth]{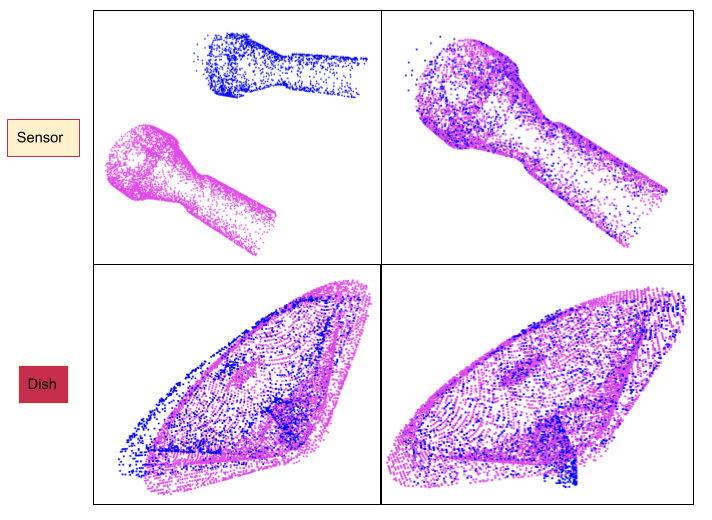}
    \caption{Results: Left - After RANSAC, Right - After ICP }
    \label{fig:After-ransac-and-after-ICP}
\end{figure}

\subsection{Final Result}
\begin{figure}[!h]
    \centering
    \includegraphics[width=0.8\textwidth]{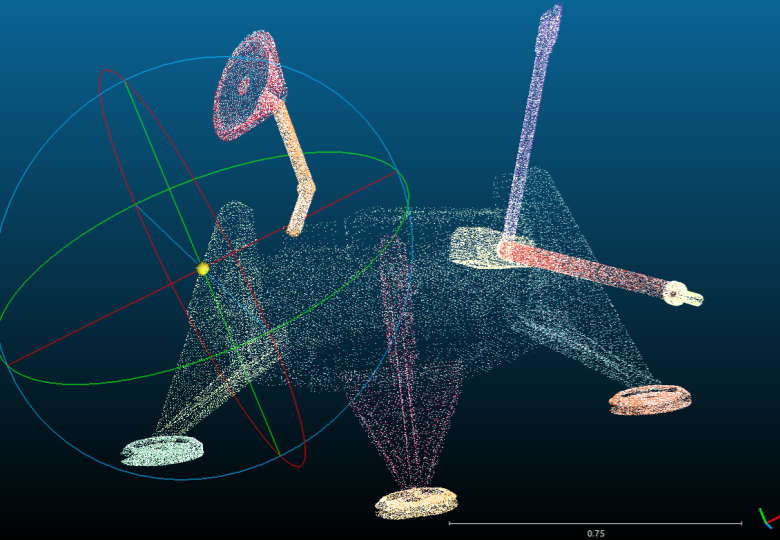}
    \caption{Final Result after ICP }
    \label{fig:final-result-ICP}
\end{figure}

\section{Addressing the challenges}

\begin{enumerate}
   \item To prevent the object model from disintegrating upon part-wise transformation we take the following steps:
   \begin{enumerate}
     \item We adjust the parts in decreasing order of sizes.
     \item In RANSAC, we add extra correspondences in the form of points that are at the junction of the current part and its larger neighbor that must remain at the same position as they are before the adjustment.
     \item If a transformation is causing a joint to be broken, we skip the adjustment
     \item If the number of feature points is too low for a given part, we skip the RANSAC step.
    \end{enumerate}
   \item To make sure that a part is searched for in the right part of the target point cloud, we take the following steps
   \begin{enumerate}
     \item We perform whole body rigid fitting to roughly align the two point clouds.
     \item Since the output of RANSAC is subject is randomness, we have an interactive interface which keeps repeating the RANSAC step until the user is satisfied with the alignment.
     \item We calculate a region of interest in the target point cloud that encompasses the part after rigid registration and all the corresponding points in the target point cloud for the feature points of the part.
    \end{enumerate}
    \item To take care of outliers:
    \begin{enumerate}
     \item KPConv applies a uniform downsampling for preventing the compute load (quadratic in the number of samples) from blowing up. If the density of a point cloud is low, this leads to serious loss in detail and significant degradation in performance. We make sure there are a minimum of 50 points in each part of the input point clouds. 
     \item We perform fine-tuning with the whole point cloud of the part rather than only the feature points.
     \item We perform ICP after RANSAC to correct for the effect of spurious correspondences.
    \end{enumerate}
 \end{enumerate}

\section{Conclusion}
In this section, we described our proposed algorithm for non-rigid registration for rigid objects with moving parts. We use the example of a NASA Viking Lander model that represents a challenging use case. We explain how the challenges identified in Chapter \ref{ch:non_rigid_point_cloud_registration} are addressed in our algorithm. In the following chapter, we present the results of our empirical evaluation.

%% file: Chapters/Chapter6.tex
\chapter{Results and Discussion}
\label{ch:results_and_discussion}
\addcontentsline{toc}{chapter}{Results and Discussion}
\lhead{\emph{Results and Discussion}}

\section{Introduction}
In Chapters \ref{ch:non_rigid_point_cloud_registration} and \ref{ch:proposed_method}, we presented our proposed method of non-rigid point cloud registration for rigid objects that are composed of movable parts. In this section, we present the results of the empirical evaluation of our approach. We also discuss the key takeaways from the experiments and present some failure modes.

\section{Experimental Setup}
We begin by presenting our experimental setup which includes key open source repositories that we depend upon, data collection and data preparation.

\subsection{Code base}
As explained in Chapter \ref{ch:proposed_method}, our algorithm builds on top of Lepard \citep{li2022lepard}. We use the official implementation of Lepard from the GitHub repository\footnote{https://github.com/rabbityl/lepard} released by the authors. We used Open3D and Mayavi libraries for visualization. The graph operations were performed using the Python NetworkX library.

\subsection{Data collection}
\begin{figure}
    \centering
    \includegraphics[width=0.5\textwidth]{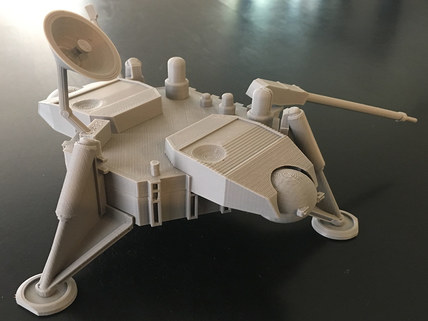}
    \caption{CAD Model of NASA Viking Lander used as source point cloud in our experiments.}
    \label{fig:lander-source-cad-model}
\end{figure}

\begin{figure}[!h]
    \centering
    \includegraphics[width=0.4\textwidth]{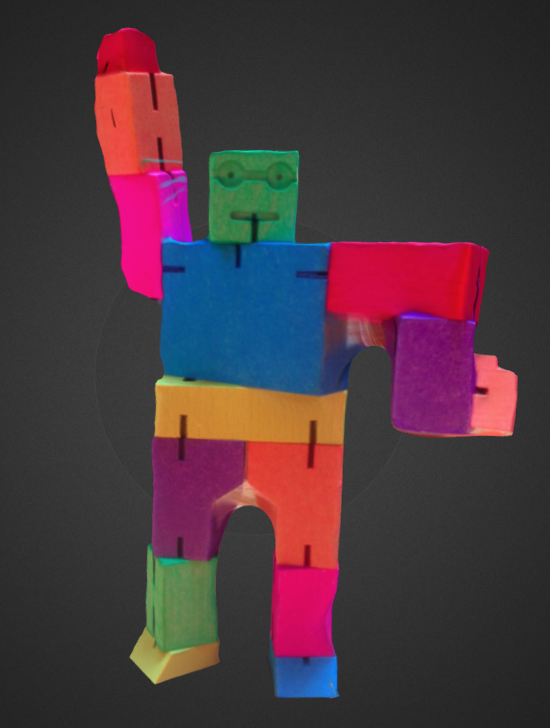}
    \caption{CAD Model of dancing Cubebot used as source point cloud in our experiments.}
    \label{fig:cubebot-source-cad-model}
\end{figure}

\begin{figure}
    \centering
    \includegraphics[width=\textwidth]{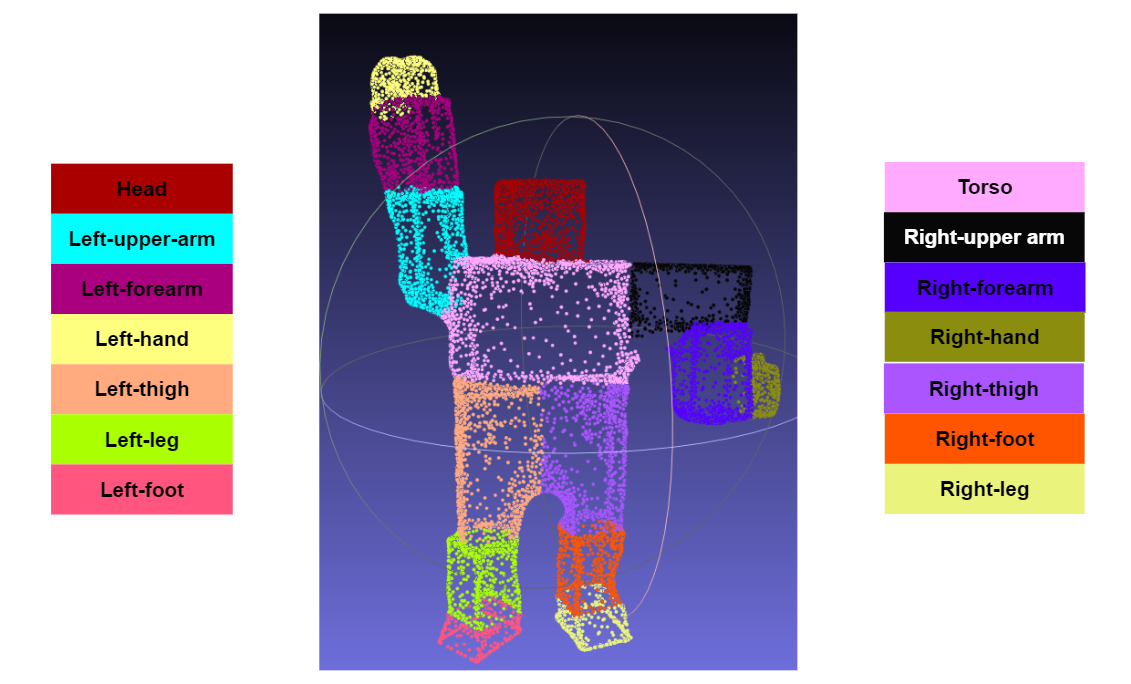}
    \caption{Cubebot with parts labelled.}
    \label{fig:cubebot-labelled}
\end{figure}

\begin{figure}[!h]
    \centering
    \includegraphics[width=\textwidth]{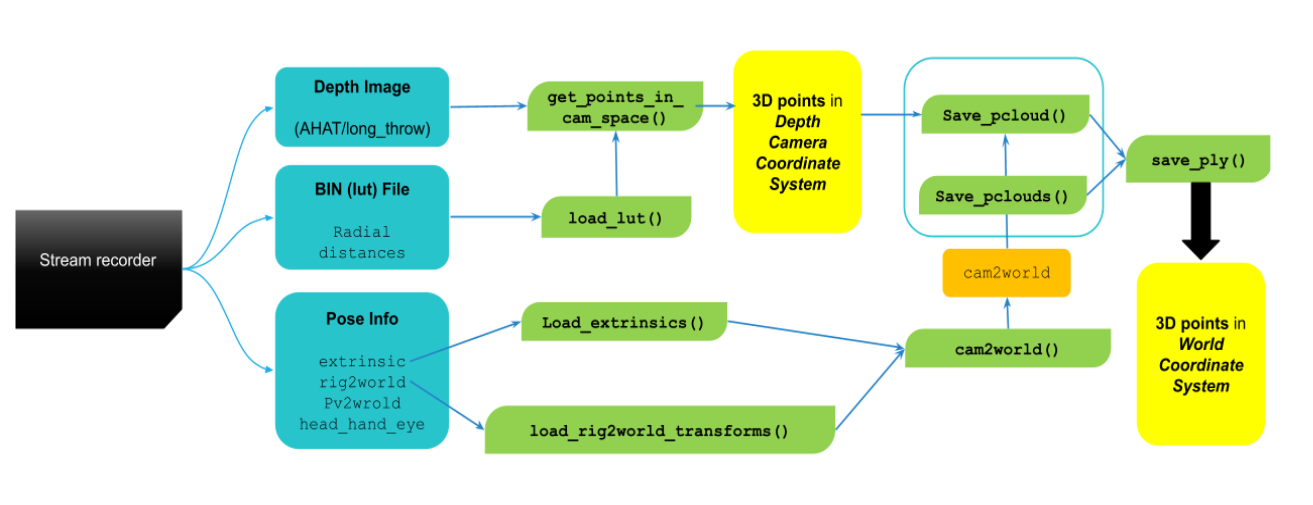}
    \caption{Pipeline for converting data recorded with Microsoft HoloLens 2 into the input format of Lepard.}
    \label{fig:point-cloud-generation}
\end{figure}

We use the trained models open-sourced by the authors of Lepard \citep{li2022lepard} for demonstrating the performance of our proposed method. There are two models - one trained on the ``3DMatch'' dataset and one on the ``4DMatch'' dataset. These datasets are comprised of a wide range of examples of rigid (3DMatch) and non-rigid (4DMatch) registration with varying degrees of noise and percentage of overlap. We discussed these datasets in detail in Section \ref{sec:datasets}.
As mentioned in Section \ref{ch:non_rigid_point_cloud_registration}, our test subjects in this thesis are rigid objects with parts that can move about joints. We use 3D printed models of the NASA Viking Lander\footnote{Open source model of NASA Viking Lander obtained from \url{https://nasa3d.arc.nasa.gov/detail/viking-lander.}} and the Cubebot\footnote{Open source model of Cubebot in a dancing pose obtained from \url{https://sketchfab.com/3d-models/dancing-cubebot-fe41ed9dce7e41b5a46fed8705a6821e.}} as shown in Figures \ref{fig:lander-source-cad-model} and \ref{fig:cubebot-source-cad-model} as representatives in our experiments. The parts of the NASA Viking Lander and the Cubebot are labelled in Figures \ref{fig:Viking-lander-labelled} and \ref{fig:cubebot-labelled} respectively. The NASA Viking Lander poses the challenge of complex geometry. On the other hand, the Cubebot has all its parts made of cuboids and has a considerably simpler geometry. However, it poses a different set of challenges due to ambiguity between the parts of similar geometry and can test the efficacy of the Lepard \citep{li2022lepard} pipeline as claimed by the authors. The source point clouds are extracted directly from the CAD models. For scanning the target point cloud, we use Microsoft Hololens 2 \citep{ungureanu2020hololens}. We set up the workflow shown in Figure \ref{fig:point-cloud-generation} to ingest Hololens 2 data into the Lepard pipeline.


\subsection{Data preparation}
\label{sec:data-preparation}
The uniform subsampling implemented by KPConv within the Lepard pipeline is sensitive to the scale of the point cloud. Small scans are often lost after the subsampling operation. We scaled the point cloud up by $10\times$ along all three axes to prevent critical data loss. In order to segment out the target object from background objects and walls in the scanned point cloud from Hololens 2, we use distance thresholding and DBSCAN clustering \citep{ester1996density}. In the experiments with a low amount of noise, we fuse multiple scans from different angles to obtain a near-complete target point cloud. 

\section{Results}
In this section, we present the results of our experiments. We design experiments with progressive levels of difficulty to demonstrate the strengths and weaknesses of our algorithm. Difficulty arises from the number of deformations and degrees of noise in the target point cloud.

\subsection{Experiment 1: No deformations, low noise}
\label{ex:experiment_1}

The first experiment is designed to demonstrate the best case scenario of non-rigid registration which is the case of zero deformations and a clean scan of the target point cloud. The source and the target point clouds can be aligned through a whole body rigid transformation. 

\subsubsection{Discussion}

\begin{figure}
    \centering
    \includegraphics[width=0.8\textwidth]{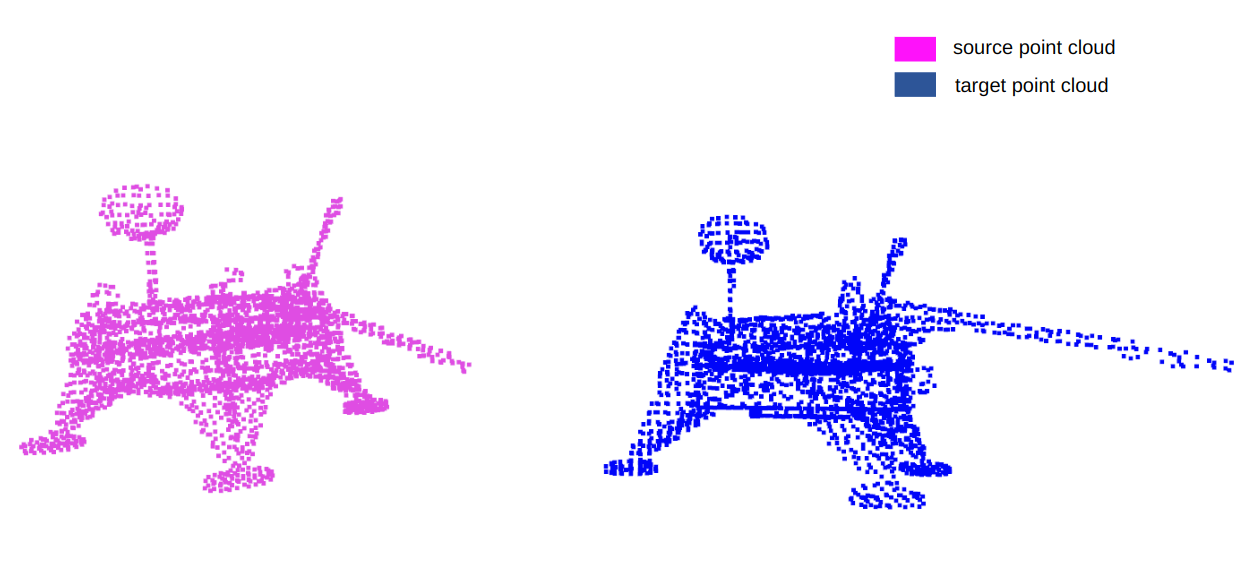}
    \caption{Source and target point clouds for Experiment 1 in Section \ref{ex:experiment_1} with NASA Viking Lander.}
    \label{fig:lander-source-and-target-expt1}
\end{figure}

\begin{figure}
    \centering
    \includegraphics[width=0.8\textwidth]{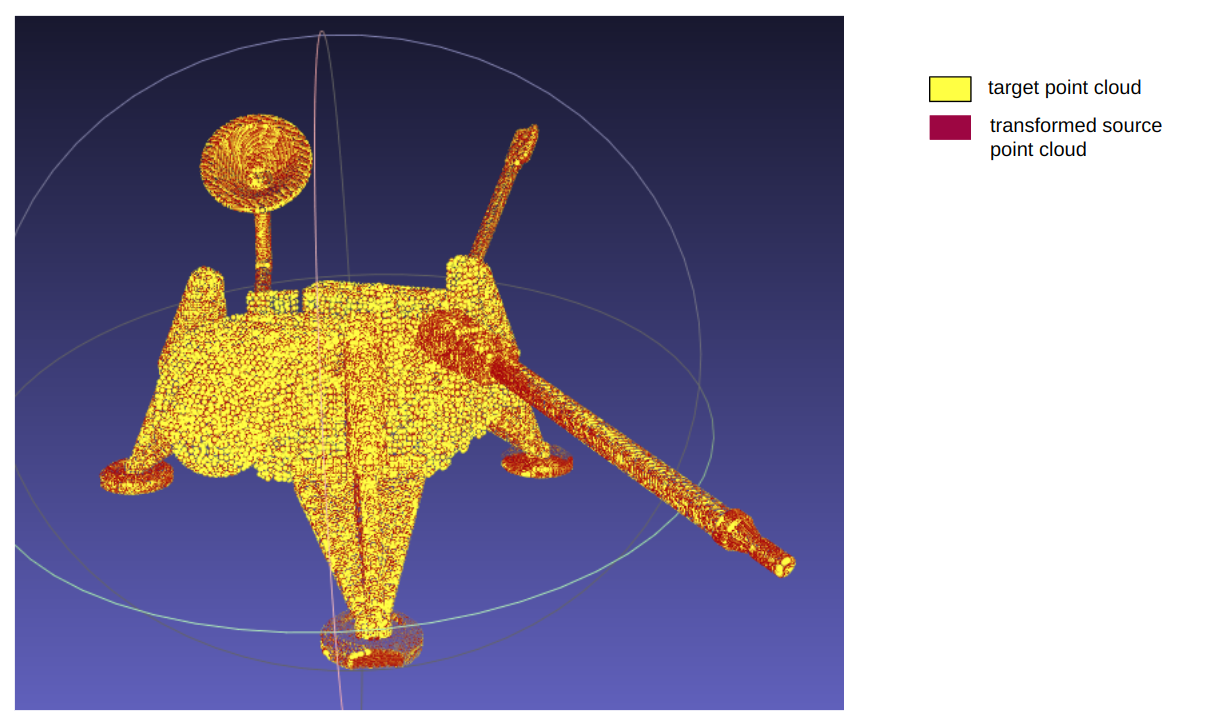}
    \caption{Transformed source and target point clouds for Experiment 1 in Section \ref{ex:experiment_1} with NASA Viking Lander.}
    \label{fig:lander-source-and-target-after-reg-expt1}
\end{figure}

\begin{figure}
    \centering
    \includegraphics[width=0.8\textwidth]{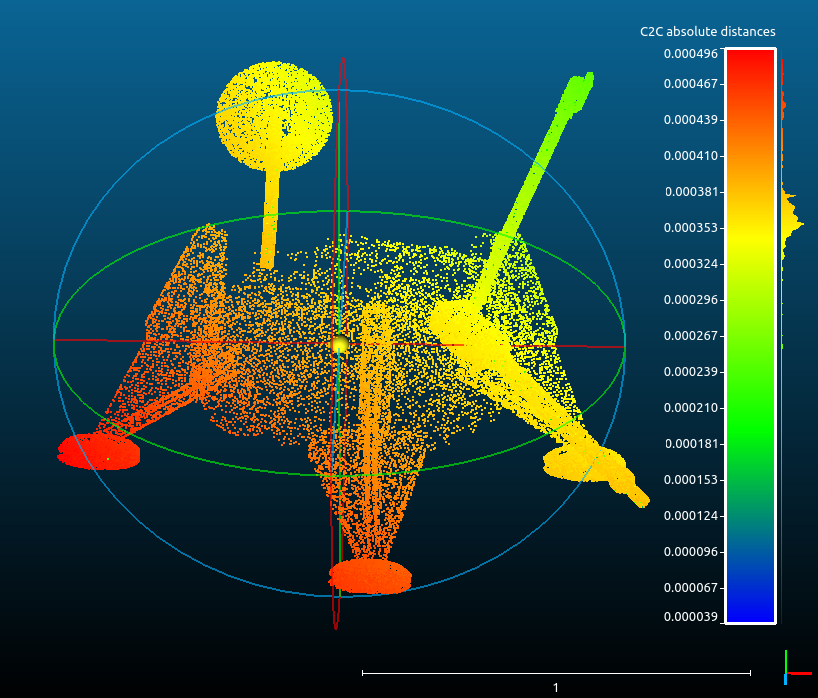}
    \caption{Visualization of alignment of transformed source and target point clouds for Experiment 1 in Section \ref{ex:experiment_1} with NASA Viking Lander. C2C distance is a cloud-to-cloud approximate nearest neighbor distance computed by CloudCompare \citep{girardeau2016cloudcompare}. }
    \label{fig:lander-alignment-after-reg-expt1}
\end{figure}

\begin{figure}
    \centering
    \includegraphics[width=0.7\textwidth]{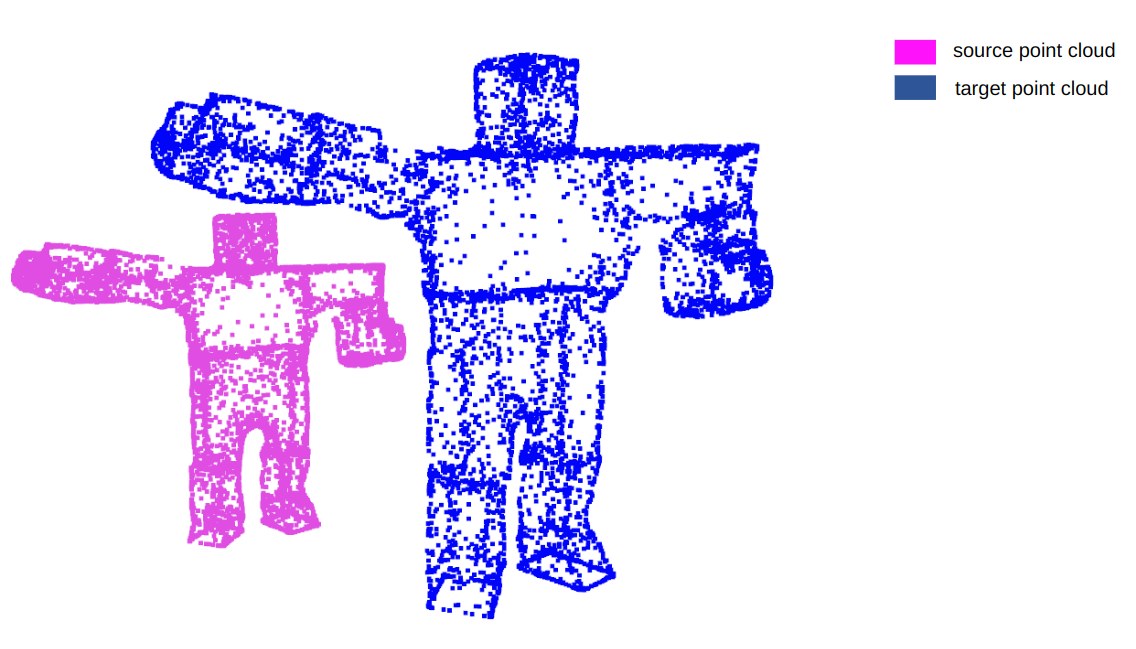}
    \caption{Source and target point clouds for Experiment 1 in Section \ref{ex:experiment_1} with Cubebot.}
    \label{fig:cubebot-source-and-target-expt1}
\end{figure}

\begin{figure}
    \centering
    \includegraphics[width=0.7\textwidth]{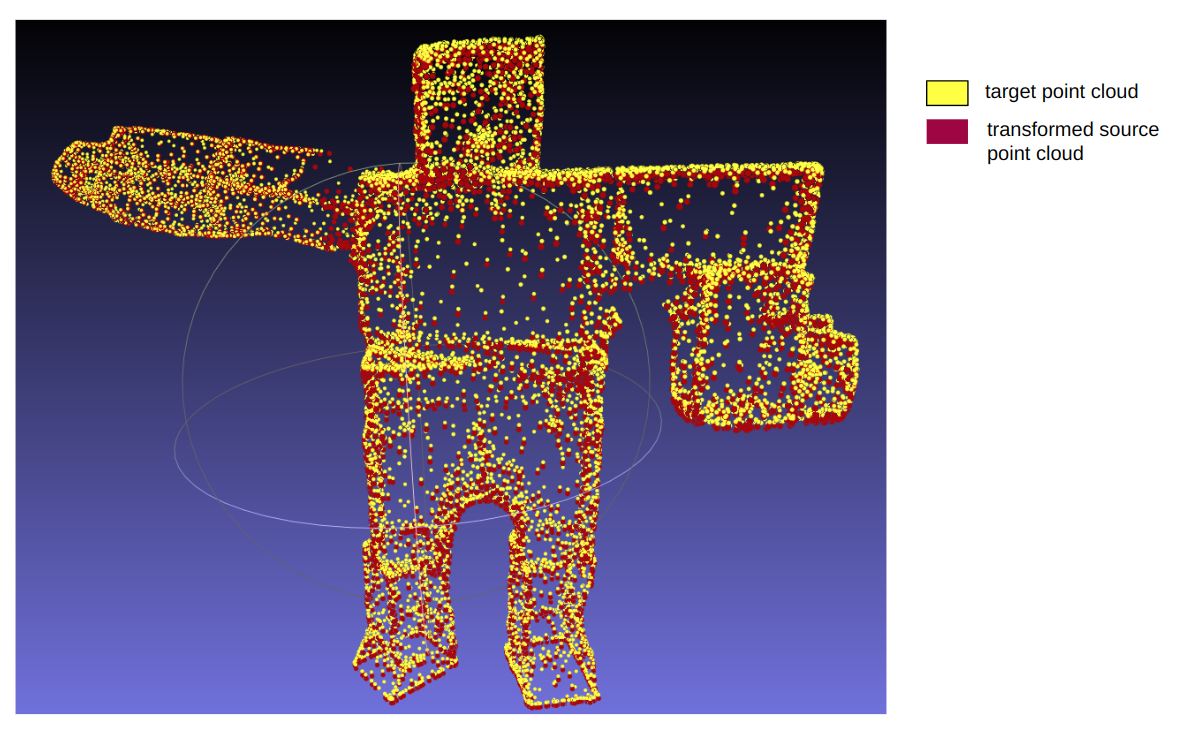}
    \caption{Transformed source and target point clouds for Experiment 1 in Section \ref{ex:experiment_1} with Cubebot.}
    \label{fig:cubebot-source-and-target-after-reg-expt1}
\end{figure}

\begin{figure}
    \centering
    \includegraphics[width=0.7\textwidth]{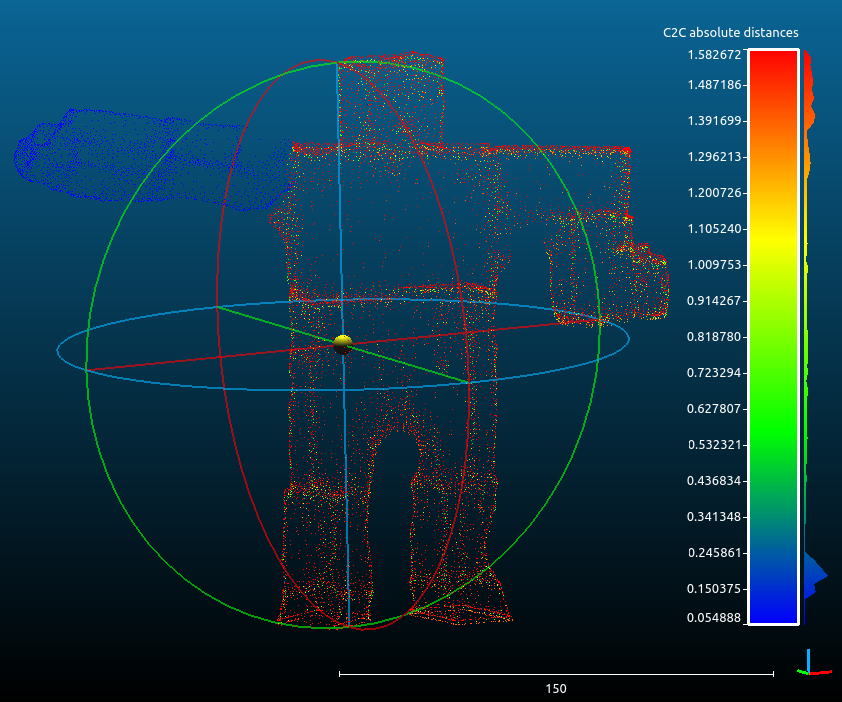}
    \caption{Visualization of alignment of transformed source and target point clouds for Experiment 1 in Section \ref{ex:experiment_1} with Cubebot. C2C distance is a cloud-to-cloud approximate nearest neighbour distance computed by CloudCompare \citep{girardeau2016cloudcompare}. }
    \label{fig:cubebot-alignment-after-reg-expt1}
\end{figure}

Figures \ref{fig:lander-source-and-target-expt1} and \ref{fig:cubebot-source-and-target-expt1} show the source and target point clouds for NASA Viking Lander and Cubebot respectively. The respective transformed source and target point clouds after registration are in Figures \ref{fig:lander-source-and-target-after-reg-expt1} and \ref{fig:cubebot-source-and-target-after-reg-expt1}. Figures \ref{fig:lander-alignment-after-reg-expt1} and \ref{fig:cubebot-alignment-after-reg-expt1} show the degree of alignment through a map of the nearest neighbor distance between the transformed-source and target point clouds. We observe that our workflow is able to accurately register the source and target point clouds for the degenerate case of no deformations and low noise.

\subsection{Experiment 2: Non-rigid deformation in one part, low noise}
\label{ex:experiment_2}

In the second experiment, we introduce a non-rigid deformation in the target point cloud. In the case of NASA Viking Lander, the ``dish'' along with the ``dish-pole'' is deformed. In the case of Cubebot, the ``left-hand'' is deformed. For alignment, we must carry out a whole-body rigid transformation followed by part-wise tuning. In this task, the target scan is clean. 

\subsubsection{Discussion}

\begin{figure}
    \centering
    \includegraphics[width=0.8\textwidth]{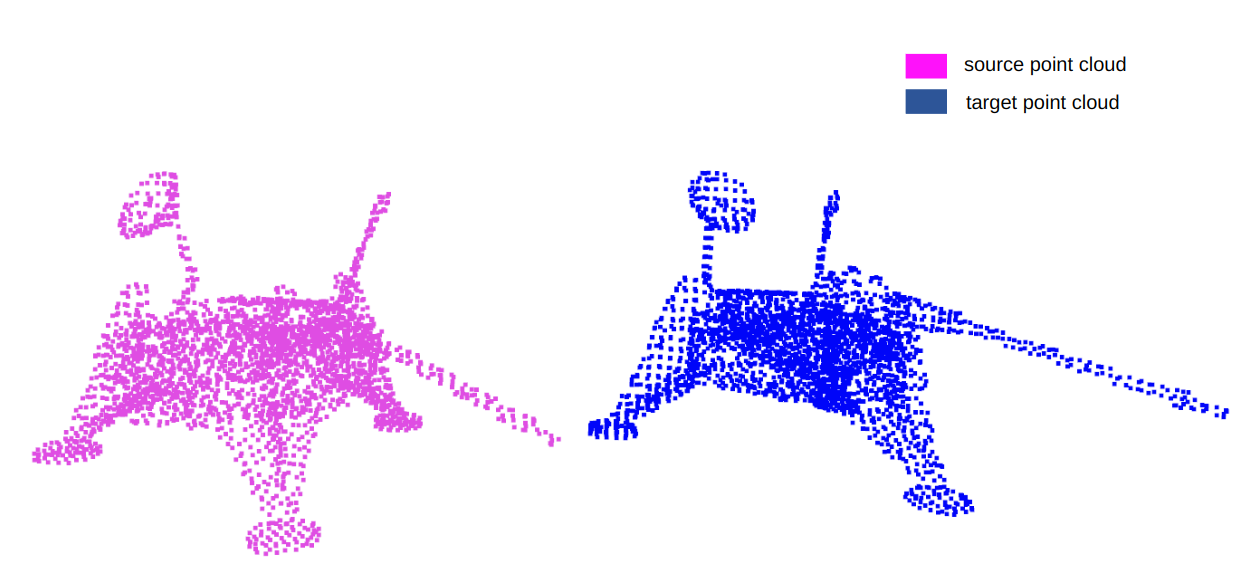}
    \caption{Source and target point clouds for Experiment 2 in Section \ref{ex:experiment_2} with NASA Viking Lander.}
    \label{fig:lander-source-and-target-expt2}
\end{figure}

\begin{figure}
    \centering
    \includegraphics[width=0.8\textwidth]{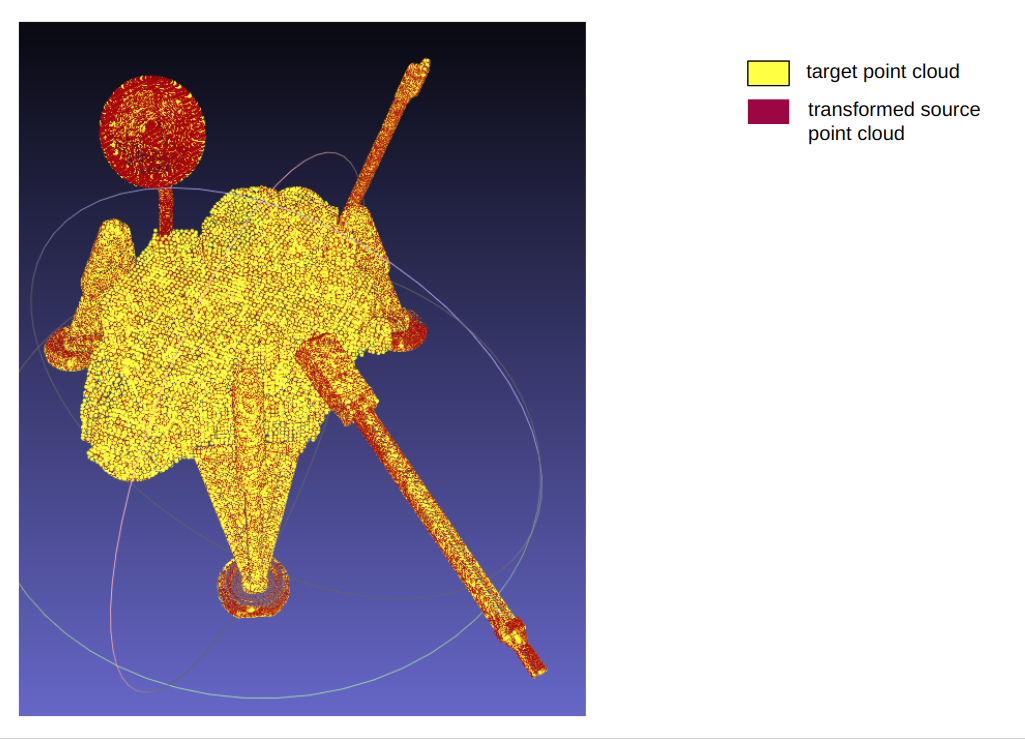}
    \caption{Transformed source and target point clouds for Experiment 2 in Section \ref{ex:experiment_2} with NASA Viking Lander.}
    \label{fig:lander-source-and-target-after-reg-expt2}
\end{figure}

\begin{figure}
    \centering
    \includegraphics[width=0.8\textwidth]{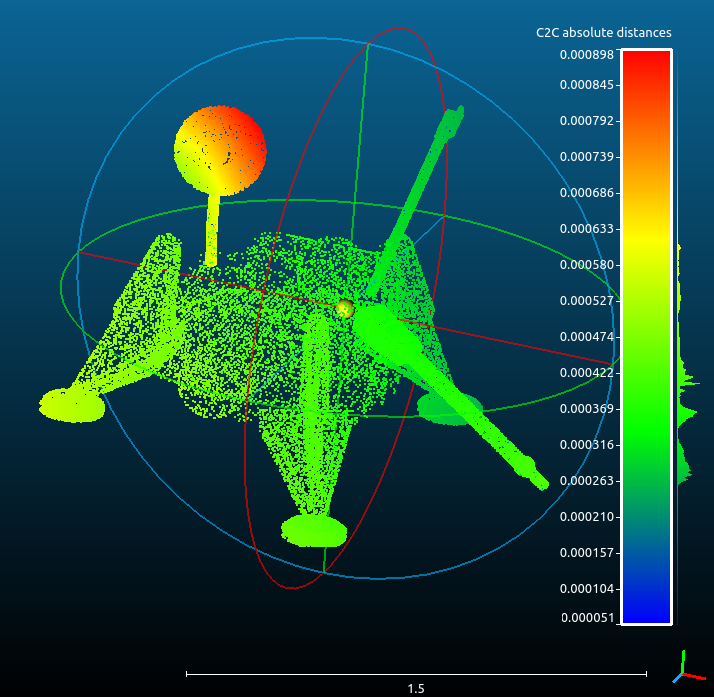}
    \caption{Visualization of alignment of transformed source and target point clouds for Experiment 2 in Section \ref{ex:experiment_2} with NASA Viking Lander. C2C distance is a cloud-to-cloud approximate nearest neighbor distance computed by CloudCompare \citep{girardeau2016cloudcompare}. }
    \label{fig:lander-alignment-after-reg-expt2}
\end{figure}

\begin{figure}
    \centering
    \includegraphics[width=0.7\textwidth]{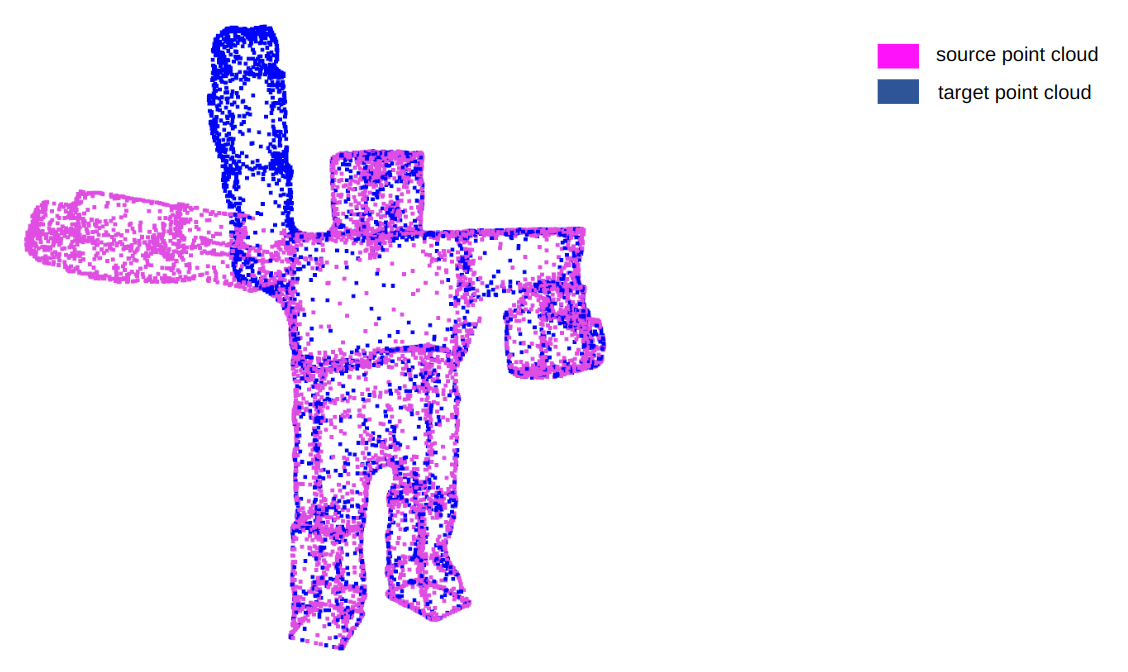}
    \caption{Source and target point clouds for Experiment 2 in Section \ref{ex:experiment_2} with Cubebot.}
    \label{fig:cubebot-source-and-target-expt2}
\end{figure}

\begin{figure}
    \centering
    \includegraphics[width=0.7\textwidth]{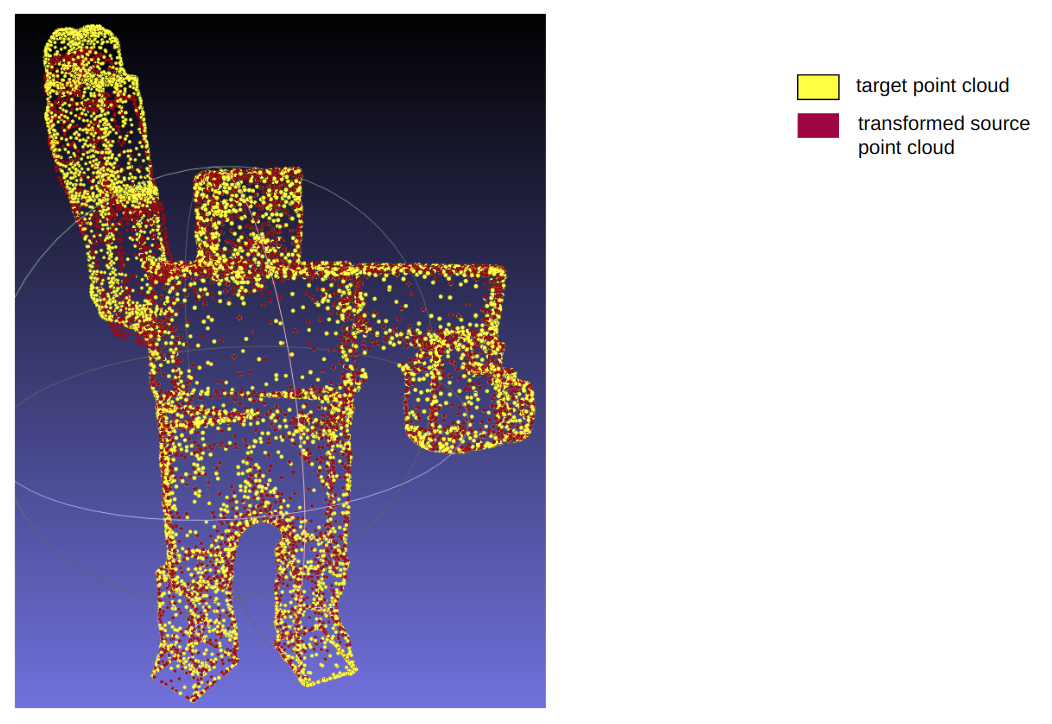}
    \caption{Transformed source and target point clouds for Experiment 2 in Section \ref{ex:experiment_2} with Cubebot.}
    \label{fig:cubebot-source-and-target-after-reg-expt2}
\end{figure}

\begin{figure}
    \centering
    \includegraphics[width=0.7\textwidth]{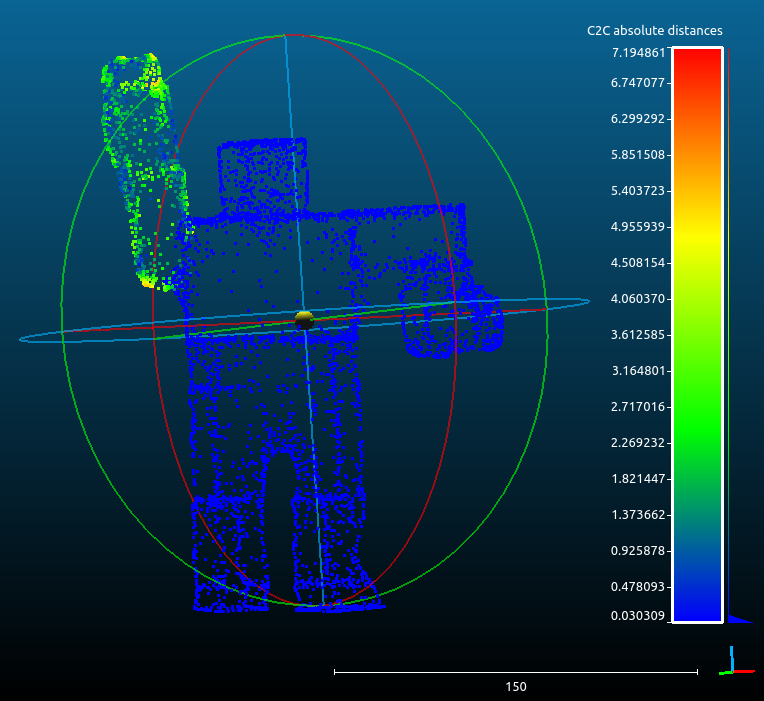}
    \caption{Visualization of alignment of transformed source and target point clouds for Experiment 2 in Section \ref{ex:experiment_2} with Cubebot. C2C distance is a cloud-to-cloud approximate nearest neighbor distance computed by CloudCompare \citep{girardeau2016cloudcompare}. }
    \label{fig:cubebot-alignment-after-reg-expt2}
\end{figure}

Figures \ref{fig:lander-source-and-target-expt2} and \ref{fig:cubebot-source-and-target-expt2} show the source and target point clouds for NASA Viking Lander and Cubebot respectively. The respective transformed source and target point clouds after registration are in Figures \ref{fig:lander-source-and-target-after-reg-expt2} and \ref{fig:cubebot-source-and-target-after-reg-expt2}. Figures \ref{fig:lander-alignment-after-reg-expt2} and \ref{fig:cubebot-alignment-after-reg-expt2} show the degree of alignment through a map of the nearest neighbour distance between the transformed source and target point clouds. We observe that our workflow is able to accurately register the source and target point clouds in the case of the Lander. We can observe a small misalignment around the deformed hand of the Cubebot as can be anticipated in the case of sparse point clouds.

\subsection{Experiment 3: Non-rigid deformation in multiple parts, low noise}
\label{ex:experiment_3}

In this experiment, we consider a slightly more difficult and a more generic version of Experiment 2 (Section \ref{ex:experiment_2}) where multiple parts of the object undergo non-rigid deformations. In case of the NASA Viking Lander, the ``dish'', ``dish-pole'', ``sensor'', ``sensor-base'' and ``sensor-pole'' are deformed. In case of the Cubebot, the ``left-hand'' and the ``left-leg'' are deformed. 

\subsubsection{Discussion}

\begin{figure}
    \centering
    \includegraphics[width=0.8\textwidth]{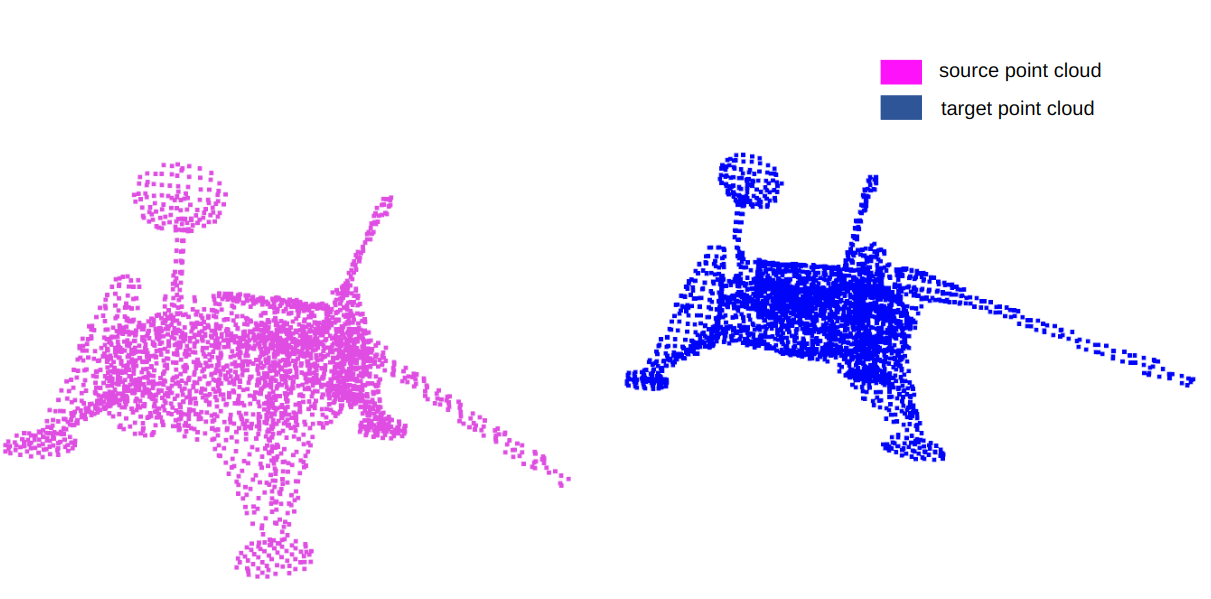}
    \caption{Source and target point clouds for Experiment 3 in Section \ref{ex:experiment_3} with NASA Viking Lander.}
    \label{fig:lander-source-and-target-expt3}
\end{figure}

\begin{figure}
    \centering
    \includegraphics[width=0.8\textwidth]{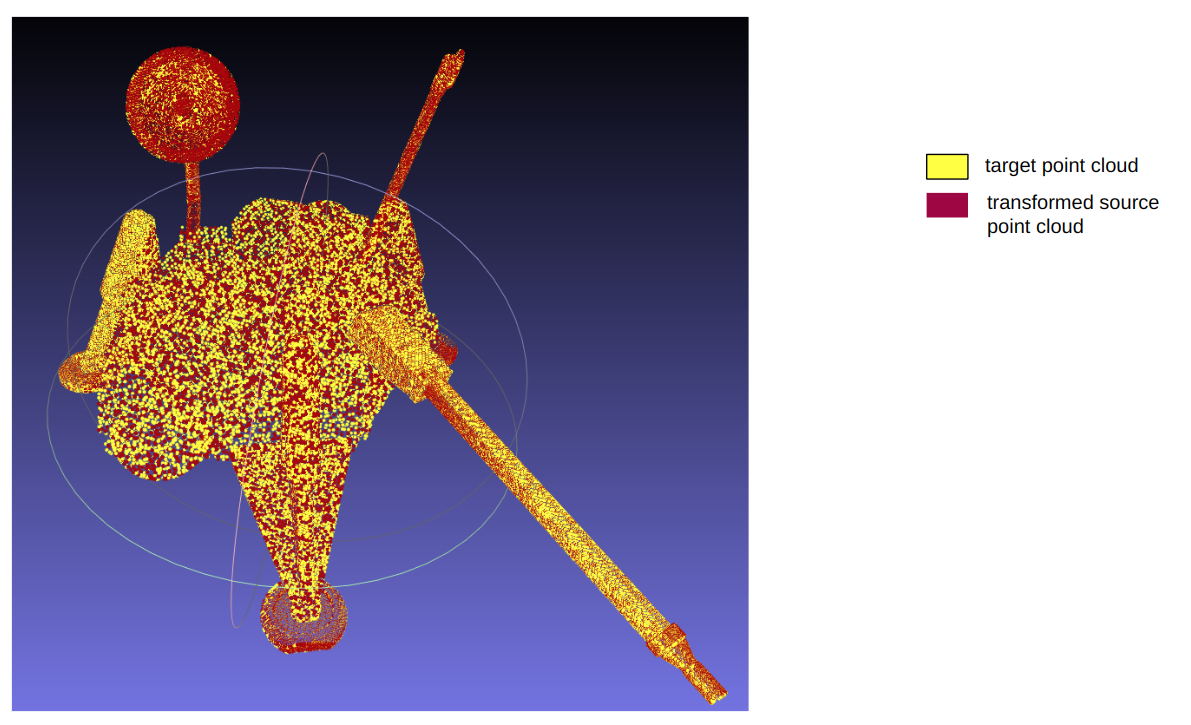}
    \caption{Transformed source and target point clouds for Experiment 3 in Section \ref{ex:experiment_3} with NASA Viking Lander.}
    \label{fig:lander-source-and-target-after-reg-expt3}
\end{figure}

\begin{figure}
    \centering
    \includegraphics[width=0.8\textwidth]{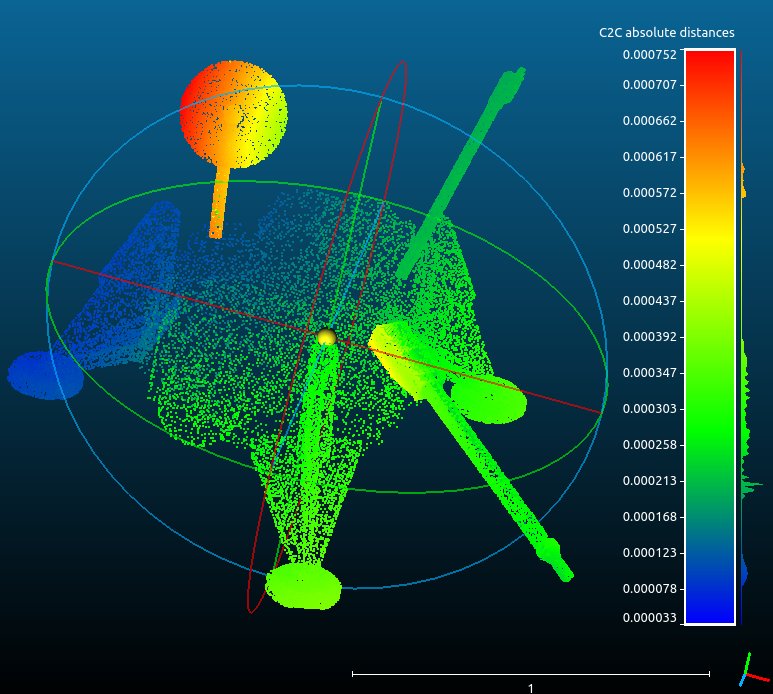}
    \caption{Visualization of alignment of transformed source and target point clouds for Experiment 3 in Section \ref{ex:experiment_3} with NASA Viking Lander. C2C distance is a cloud-to-cloud approximate nearest neighbor distance computed by CloudCompare \citep{girardeau2016cloudcompare}. }
    \label{fig:lander-alignment-after-reg-expt3}
\end{figure}

\begin{figure}
    \centering
    \includegraphics[width=0.7\textwidth]{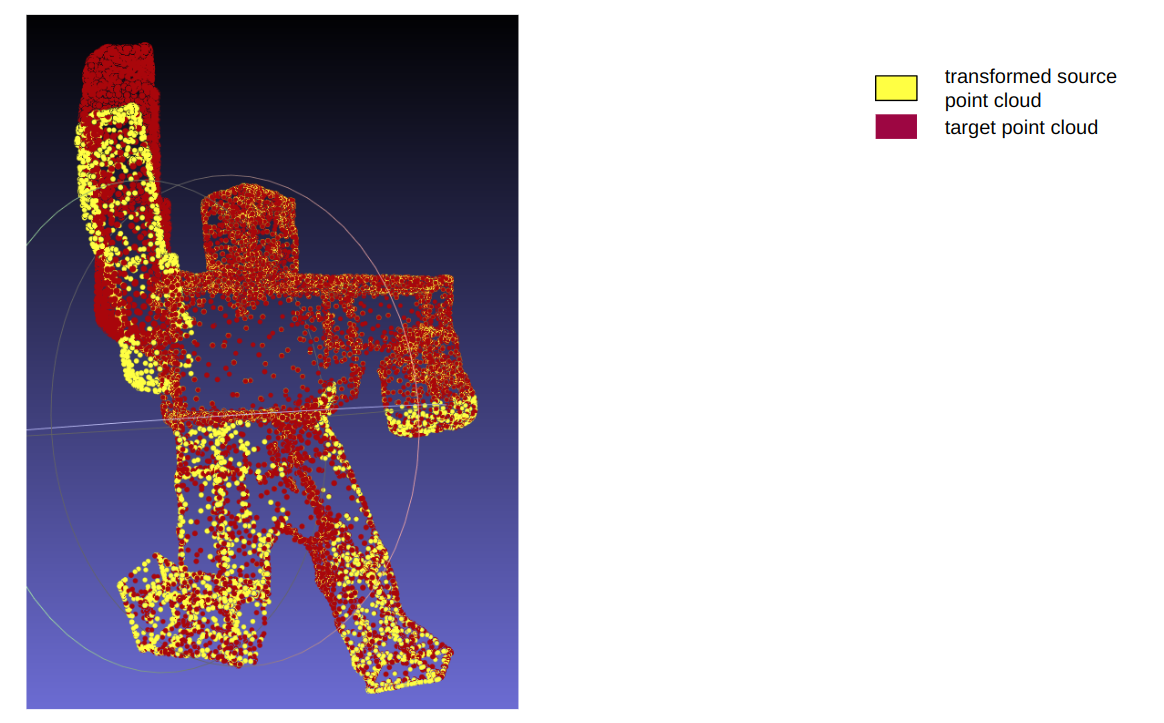}
    \caption{Transformed source and target point clouds for Experiment 3 in Section \ref{ex:experiment_3} with Cubebot.}
    \label{fig:cubebot-source-and-target-after-reg-expt3}
\end{figure}

\begin{figure}
    \centering
    \includegraphics[width=0.7\textwidth]{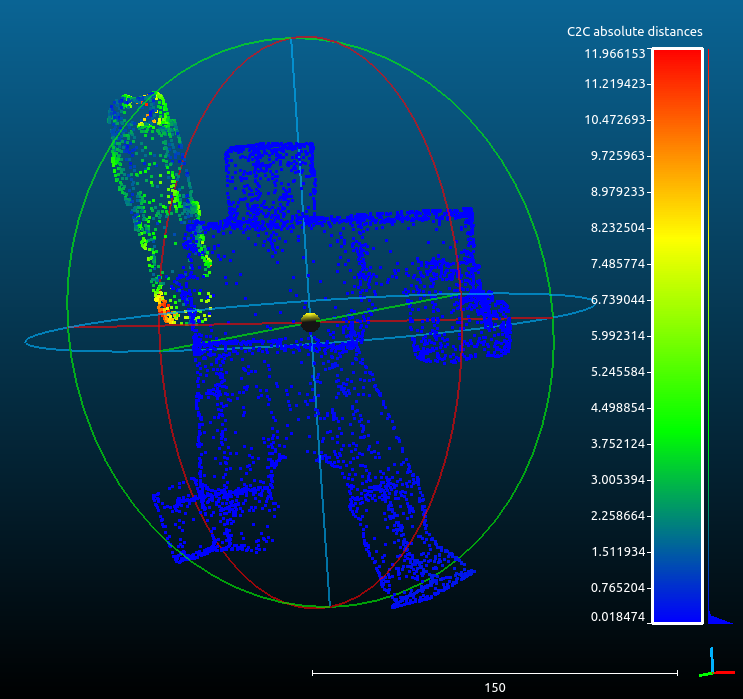}
    \caption{Visualization of alignment of transformed source and target point clouds for Experiment 3 in Section \ref{ex:experiment_3} with Cubebot. C2C distance is a cloud-to-cloud approximate nearest neighbour distance computed by CloudCompare \citep{girardeau2016cloudcompare}. }
    \label{fig:cubebot-alignment-after-reg-expt3}
\end{figure}

Figures \ref{fig:lander-source-and-target-expt3} and \ref{fig:cubebot-source-and-target-after-reg-expt3} show the source and target point clouds for NASA Viking Lander and Cubebot respectively. The respective transformed source and target point clouds after registration are in Figures \ref{fig:lander-source-and-target-after-reg-expt3} and \ref{fig:cubebot-source-and-target-after-reg-expt3}. Figures \ref{fig:lander-alignment-after-reg-expt3} and \ref{fig:cubebot-alignment-after-reg-expt3} show the degree of alignment through a map of the nearest neighbor distance between the transformed-source and target point clouds. We observe that our workflow is able to accurately register the source and target point clouds for the Lander. As in Experiment 2, we can observe a bit of misalignment around in the deformed hand of the Cubebot as can be anticipated in the case of sparse point clouds.

\begin{figure}
    \centering
    \includegraphics[width=0.7\textwidth]{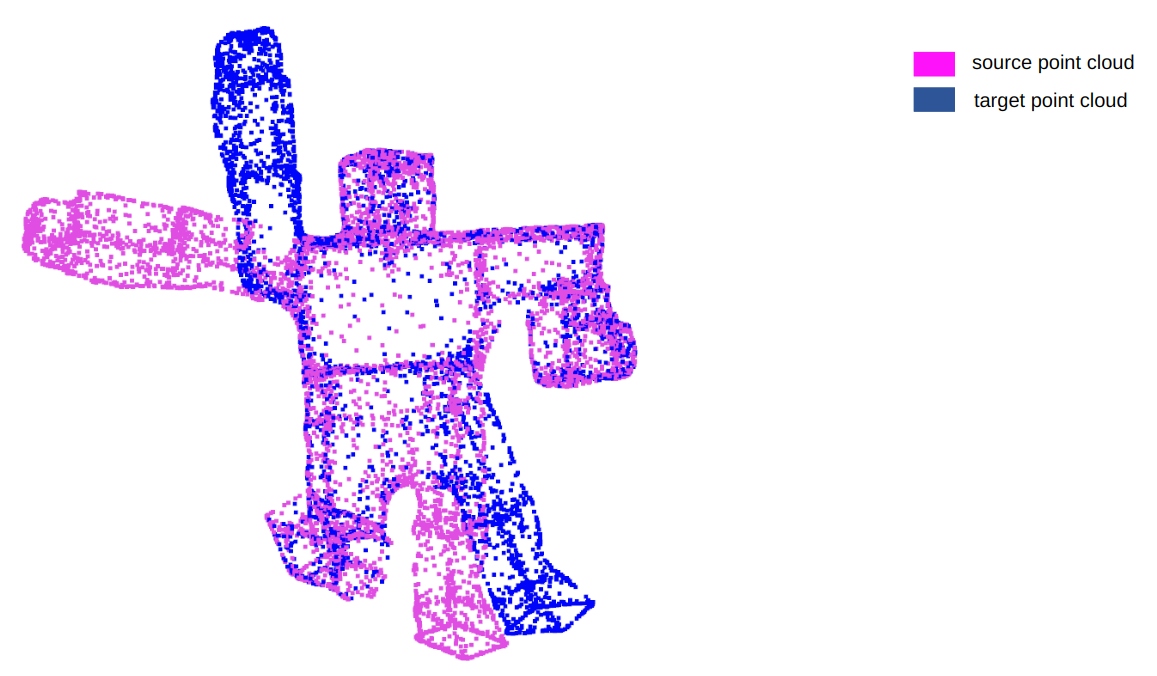}
    \caption{Source and target point clouds for Experiment 3 in Section \ref{ex:experiment_3} with Cubebot.}
    \label{fig:cubebot-source-and-target-expt3}
\end{figure}

\subsection{Experiment 4: Non-rigid deformation in multiple parts, Noisy target point cloud}
\label{ex:experiment_4}

This is the most difficult setting where we not only have multiple deformations but also a noisy scan of the target point cloud with holes. This closely mimics real-world use cases in a mixed reality where it is common to have holes and missing parts in a 3D scan. Like Experiment 3, in the case of the NASA Viking Lander, the ``dish'', ``dish-pole'', ``sensor'', ``sensor-base'' and ``sensor-pole'' are deformed. In the case of the Cubebot, the ``left-hand'' and the ``left-leg'' are deformed. 

\subsubsection{Discussion}

\begin{figure}
    \centering
    \includegraphics[width=0.8\textwidth]{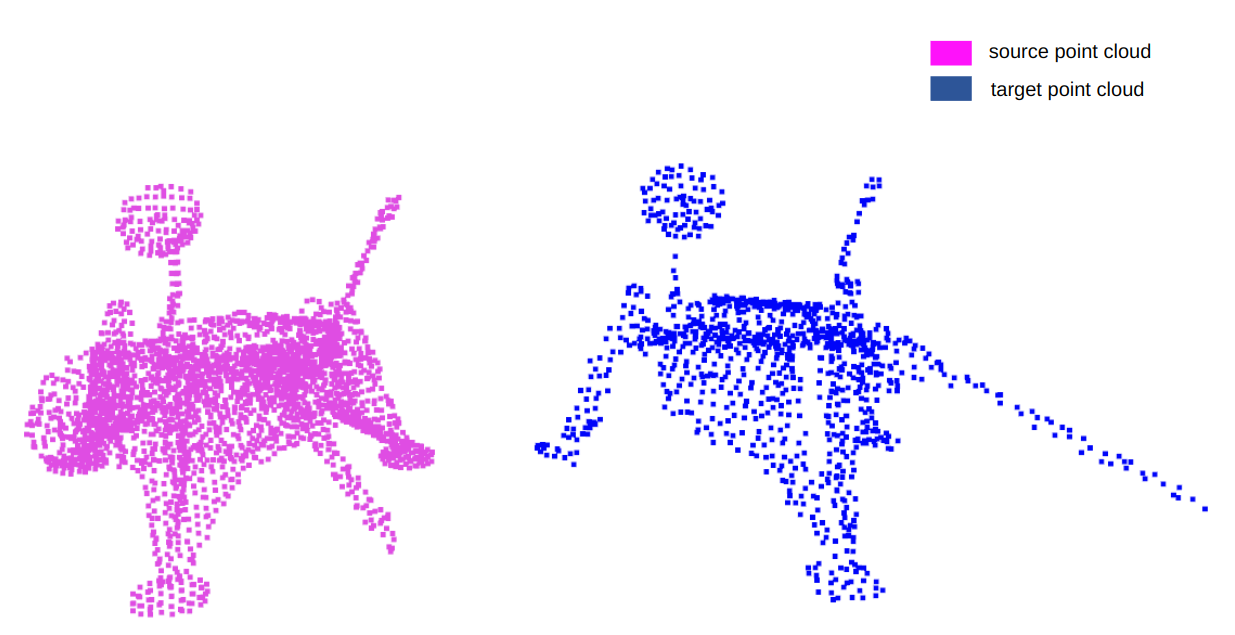}
    \caption{Source and target point clouds for Experiment 4 in Section \ref{ex:experiment_4} with NASA Viking Lander.}
    \label{fig:lander-source-and-target-expt4}
\end{figure}

\begin{figure}
    \centering
    \includegraphics[width=0.8\textwidth]{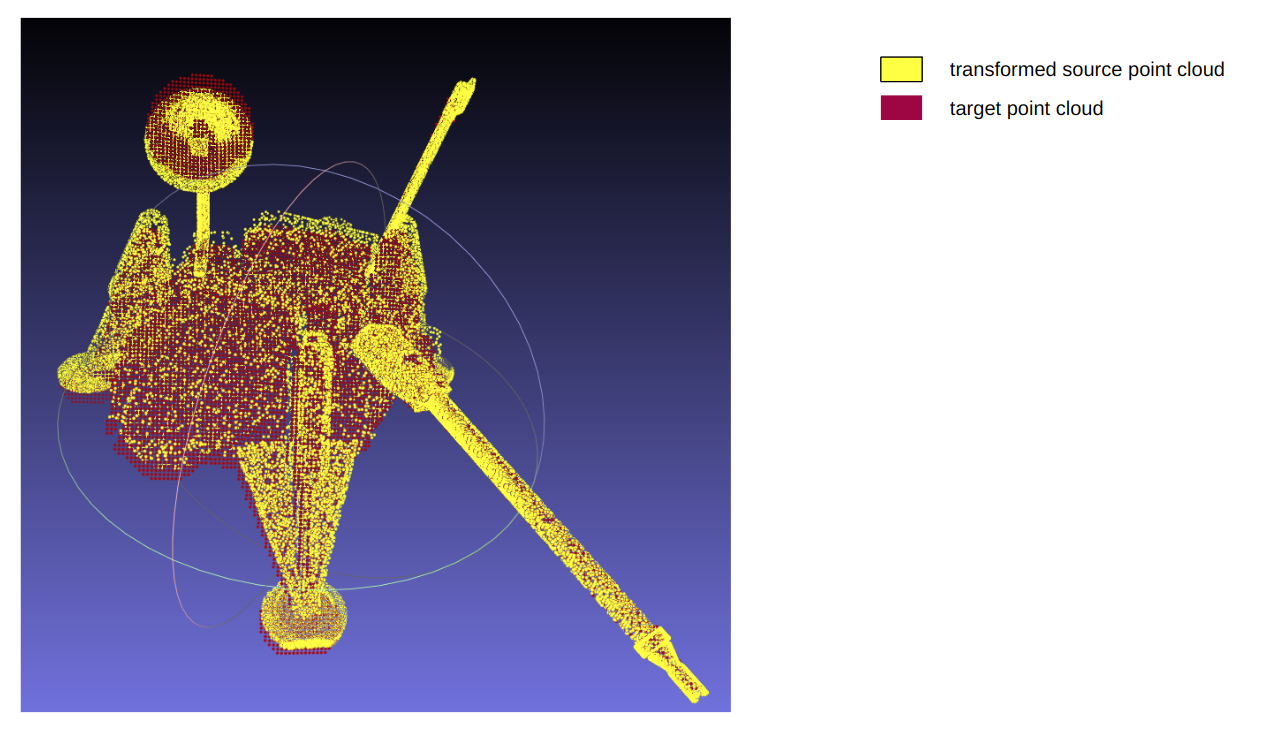}
    \caption{Transformed source and target point clouds for Experiment 4 in Section \ref{ex:experiment_4} with NASA Viking Lander.}
    \label{fig:lander-source-and-target-after-reg-expt4}
\end{figure}

\begin{figure}
    \centering
    \includegraphics[width=0.8\textwidth]{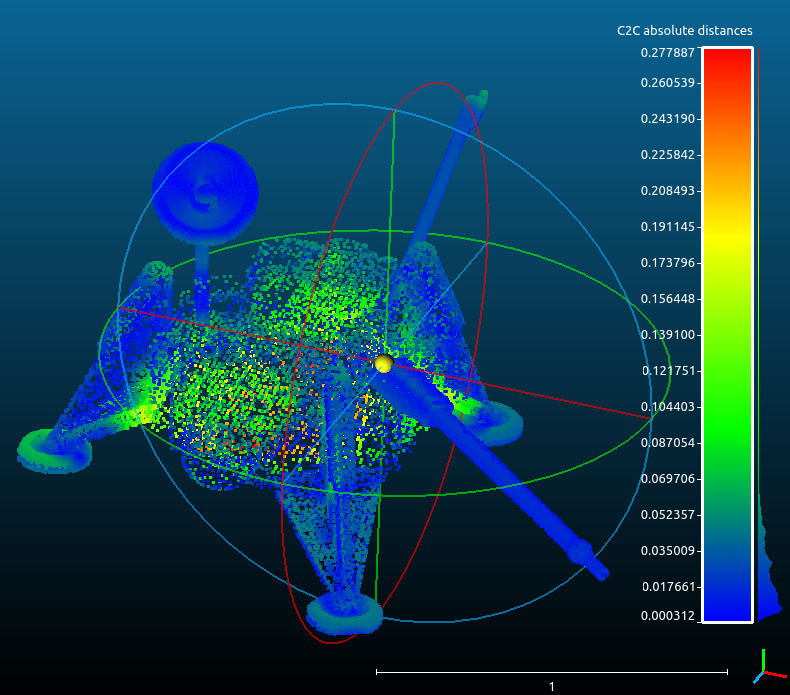}
    \caption{Visualization of alignment of transformed source and target point clouds for Experiment 4 in Section \ref{ex:experiment_4} with NASA Viking Lander. C2C distance is a cloud-to-cloud approximate nearest neighbour distance computed by CloudCompare \citep{girardeau2016cloudcompare}. }
    \label{fig:lander-alignment-after-reg-expt4}
\end{figure}

\begin{figure}
    \centering
    \includegraphics[width=0.7\textwidth]{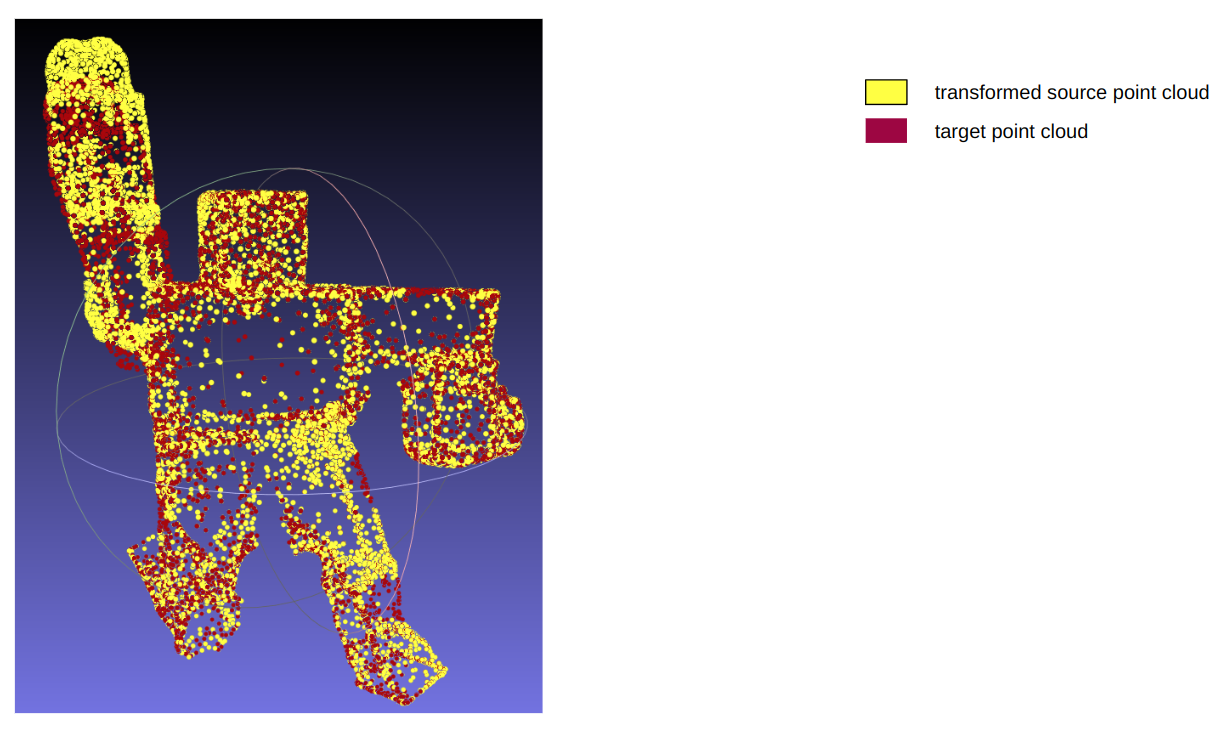}
    \caption{Transformed source and target point clouds for Experiment 4 in Section \ref{ex:experiment_4} with Cubebot.}
    \label{fig:cubebot-source-and-target-after-reg-expt4}
\end{figure}

\begin{figure}
    \centering
    \includegraphics[width=0.7\textwidth]{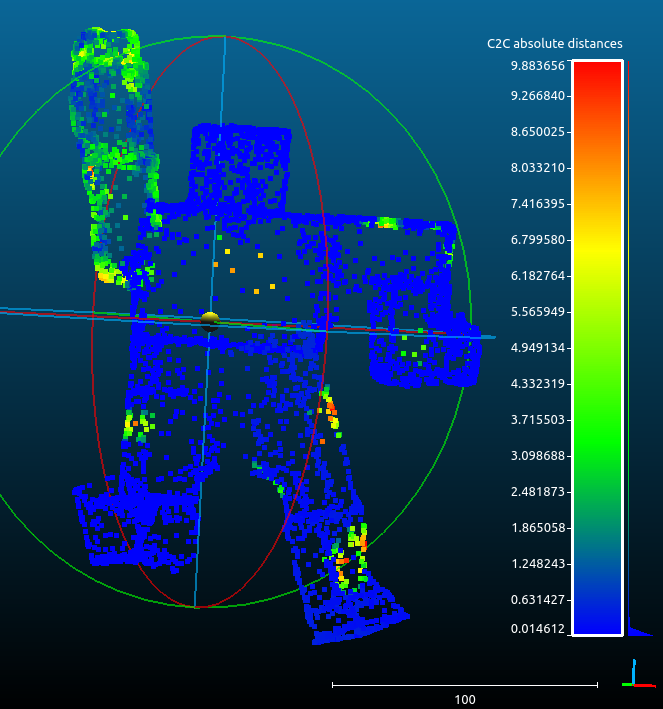}
    \caption{Visualization of alignment of transformed source and target point clouds for Experiment 4 in Section \ref{ex:experiment_4} with Cubebot. C2C distance is a cloud-to-cloud approximate nearest neighbour distance computed by CloudCompare \citep{girardeau2016cloudcompare}. }
    \label{fig:cubebot-alignment-after-reg-expt4}
\end{figure}

Figures \ref{fig:lander-source-and-target-expt4} and \ref{fig:cubebot-source-and-target-after-reg-expt4} shows the source and target point clouds for NASA Viking Lander and Cubebot respectively. The respective transformed source and target point clouds after registration are in Figures \ref{fig:lander-source-and-target-after-reg-expt4} and \ref{fig:cubebot-source-and-target-after-reg-expt4}. Figures \ref{fig:lander-alignment-after-reg-expt4} and \ref{fig:cubebot-alignment-after-reg-expt4} show the degree of alignment through a map of the nearest neighbour distance between the transformed source and target point clouds. We observe that in both cases -- Lander and Cubebot -- alignment is successful. This demonstrates that our proposed algorithm is robust to noise, holes and missing areas in the target point cloud, and hence is a viable approach for real-world mixed reality applications.

\begin{figure}
    \centering
    \includegraphics[width=0.7\textwidth]{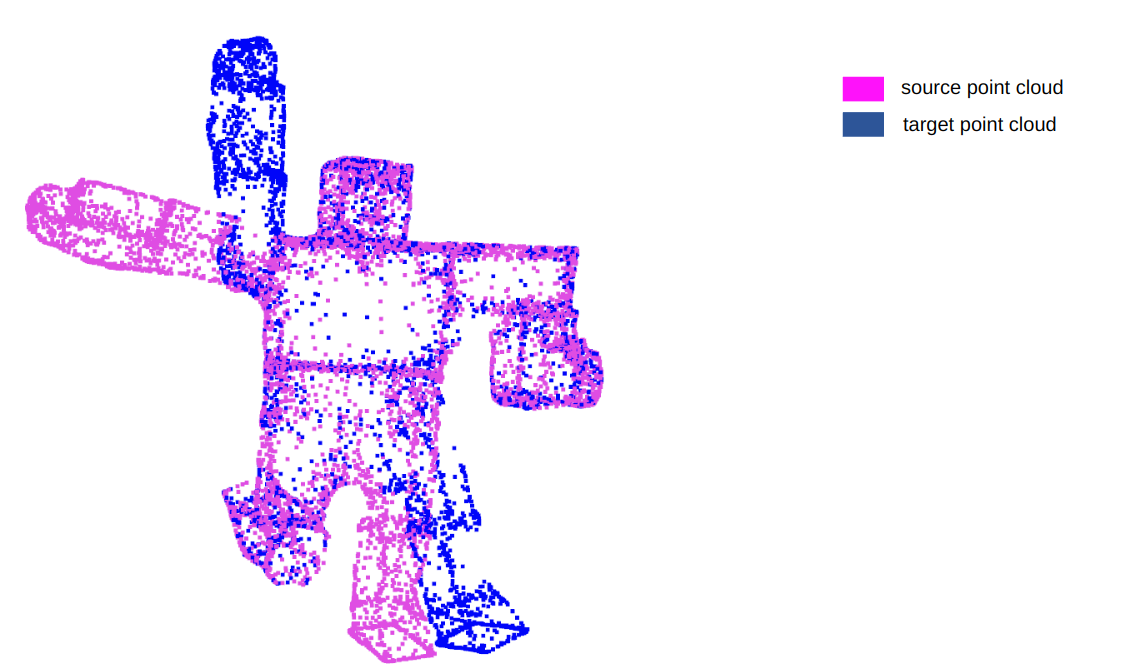}
    \caption{Source and target point clouds for Experiment 4 in Section \ref{ex:experiment_4} with Cubebot.}
    \label{fig:cubebot-source-and-target-expt4}
\end{figure}

\section{Hyperparameter values}
We present the hyperparameters of our study in Table \ref{tab:hyperparams}.

\begin{table}[!h]
\caption{Table of hyperparameters.}
\label{tab:hyperparams}
\centering
\begin{tabular}{|c|cc|}
\hline
\multicolumn{1}{|c|}{\multirow{2}{*}{Hyperparameter}} & \multicolumn{2}{c|}{Values}                       \\ \cline{2-3} 
\multicolumn{1}{|c|}{}                                & \multicolumn{1}{c|}{NASA Viking Lander} & Cubebot \\ \hline
$f_{retention}^{subsample}$                           & \multicolumn{1}{c|}{$0.5$}              & $0.4$   \\ \hline
$d_{max}^{corr}$                                      & \multicolumn{1}{c|}{$20$}               & $20$    \\ \hline
$n_{min}^{corr}$                                      & \multicolumn{1}{c|}{$5$}                & $5$     \\ \hline
\end{tabular}
\end{table}


\section{Conclusion}
In this chapter, we presented the results of empirical evaluation of the approach proposed in Chapter \ref{ch:proposed_method}. We presented $4$ experiments with increasing levels of difficulty on two real world objects that posed two major kinds of difficulty. We demonstrated the robustness of the proposed approach and discussed the failure modes.

%% file: Chapters/Chapter7.tex
\chapter{Conclusion}
\label{ch:conclusion}
\addcontentsline{toc}{chapter}{Conclusion}
\lhead{\emph{Conclusion}}

\section{Introduction}
In this section, we conclude the thesis with a recapitulation of the studies presented in the previous chapters. We also discuss the limitations of our approach and motivate future directions of research.

\section{Summary of achievements}
The aim of this thesis was to study point cloud registration in the context of deformable objects and Mixed Reality (MR) applications. We focussed on the task of alignment of CAD models with real world 3D scans as this is the one of the crucial use cases in enterprise MR at the present date. Since real time operation is crucial for a seamless MR experience, we aimed to identify crucial rate-limiting steps and optimize them for our use cases. 

Here is a summary of our achievements towards the goals of this thesis:
\begin{itemize}
    \item We present a thorough literature survey of the state-of-the-art point cloud registration algorithms that are capable of dealing with non-rigid deformations between the source and target point clouds.
    \item We implemented a data capture and rendering pipeline on Microsoft Hololens 2, a leading MR platform.
    \item We implemented one of the state-of-the-art non-rigid point cloud matching algorithms - Lepard \citep{li2022lepard} - and evaluated it on real world scans from Hololens 2.
    \item We identified the rate limiting step as the non-rigid registration / warping step that came after the inference of feature-point correspondences from the Lepard pipeline. 
    \item We narrowed our scope to objects that are composed of rigid parts that can move relative to one another about hinges and joints. These comprise the most common subjects in non-rigid registration applications in enterprise mixed reality use cases.
    \item We proposed a fast non-rigid registration workflow that leverages the graph structure of the source CAD model, correspondences predicted by Lepard, RANSAC and ICP and demonstrated high efficacy on our real world test scans even in the presence of noise, holes and missing parts.
    
\end{itemize}

\section{Contribution to knowledge}

In this thesis, we made the following findings:

\begin{itemize}
    \item Improvement in the speed of non-rigid registration can be achieved by leveraging prior knowledge about the structure of the source cloud.
    
    \item We proposed a 3-step registration workflow that reduced the sensitivity to noise and produced robust registration results.

\end{itemize}

\section{Recommendations for future work}

The method presented in this thesis leverages the structure of the object. However this might not always be available. An interesting research direction would be to segment the parts of a 3D object that move as a whole by tracking the movements of the object. This will enable us to use our proposed method for the more generic non-rigid point cloud registration use case.

%% file: Appendices/Appendix-A.tex
\chapter{Research Proposal}
\setboolean{@twoside}{false}
\includepdf[pages=-, offset=75 -75]{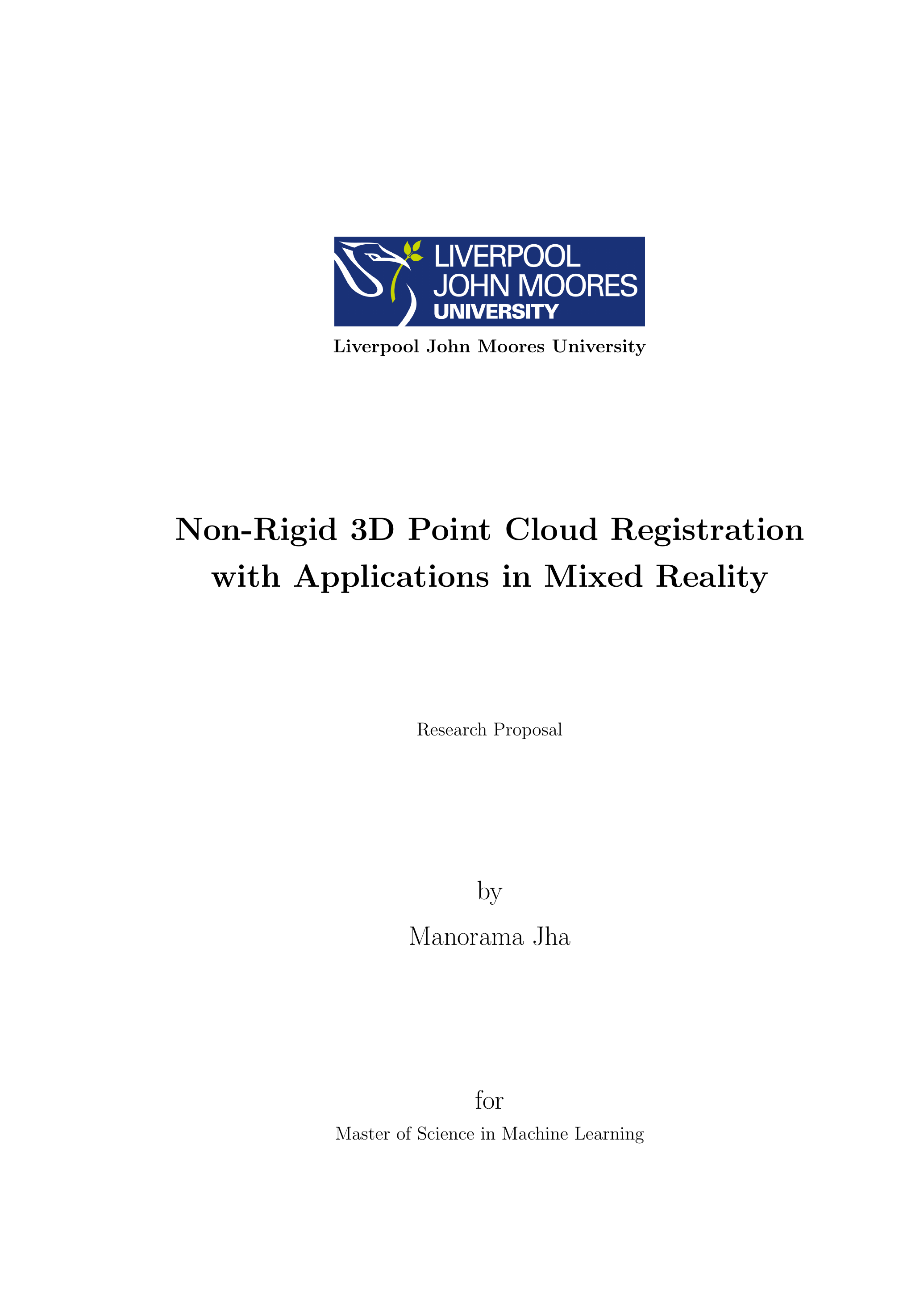}
\clearpage

\lhead{\emph{Research Proposal}}
\section{Summary of the Proposed Research}

In this project we aim to study the problem of non-rigid partial point cloud registration for Mixed Reality applications. Mixed Reality (MR) is transforming the industry by providing unique opportunities in engineering design, remote operations and collaboration. We specifically intend to implement a state-of-the-art point cloud registration algorithm - Lepard \citep{li2022lepard} - on Microsoft HoloLens 2 Mixed Reality Platform \citep{ungureanu2020hololens} and investigate potential causes of drift and latency bottlenecks in real-world MR use cases. The anticipated outcomes of this research project are a detailed account of failure modes of Lepard and related algorithms and a robust and fast inference pipeline that is robust to outliers and noise in real-life 3D scans and capable of supporting real-time Mixed Reality applications.

\section{Research Question}
In this thesis we will address the following research question:

\begin{center}
\begin{minipage}{0.7\textwidth}
“Can we make state-of-the-art non-rigid point cloud registration algorithms robust and low-latency for Mixed Reality use cases in real-world environments?”
\end{minipage}
\end{center}

\section{Aim and Objectives}
In this thesis, we aim to study the problem of 3D Point Cloud Registration for non-rigid objects. Apart from wide-spread relevance in computer vision, robotics and medical science this problem statement is central in many Mixed Reality applications. Although significant advancements have been made in 3D representation learning and computer vision in recent years, several engineering challenges stand in the way of deploying these algorithms at scale. The objectives of this thesis are as follows:

\begin{enumerate}
    \item Prepare a thorough literature review of prominent approaches to 3D Point Cloud Registration with emphasis on non-rigid objects and relevance in Mixed Reality.
    \item Implement the current state-of-the-art algorithm - Lepard \citep{li2022lepard} - on industry leading Mixed Reality platform - Microsoft HoloLens 2 \citep{ungureanu2020hololens}.
    \item Evaluate real-time 3D point cloud registration for deformable and non-deformable objects in uncontrolled real-world settings in the presence and absence of distraction and with varying levels of noise in the 3D scans.
    \item Identify and mitigate any causes of drift or misalignment and potential bottlenecks slowing down the inference pipeline.
\end{enumerate}

\section{Background}
In this section we will briefly introduce some important concepts that form the foundation of this thesis.

\subsection{Mixed Reality}
\subsubsection{Introduction}
Mixed Reality (MR) is a form of eXtended Reality (XR) that allows the user to render and interact with virtual 3D holograms in a real-world environment \citep{speicher2019mixed,rokhsaritalemi2020review}. In an interior design use case, MR would allow the user to render the hologram of a couch in an empty room and move it around the room with hand gestures \citep{huang2018augmented} to find out where in the room it would look best \citep{kent2021mixed,jain2019current}. One of the primary use cases of this technology is to overlay or register a virtual reference model of an object on top of a real instance of the same object for comparison. For example, in manufacturing, a virtual reference model of a precision engineering part can be overlaid on a damaged instance for inspection, repair \citep{abate2013mixed,borsci2015empirical} and training \citep{mueller2003marvel,wang2004mixed}. In remote assistance, a circuit diagram can be overlaid on a real switchboard and remote instructions can be issued for debugging the connections \citep{ladwig2018literature}. In surgery, pre-operative scans (like MRI, CT Scan, etc.) can be overlaid on the patient’s organ in real-time for precise guidance to the point of surgery \citep{chen2017recent}. 

\subsubsection{MR Headsets}
\begin{figure}[!h]
    \centering
    \includegraphics[width=\textwidth]{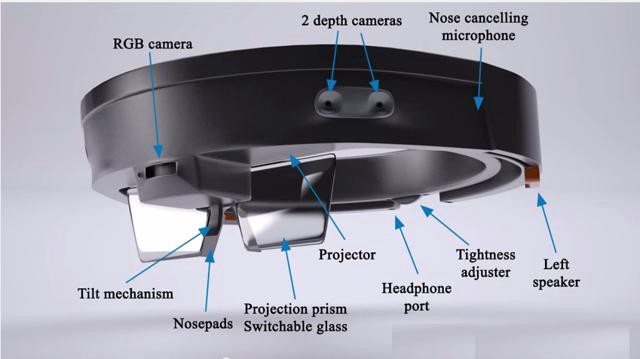}
    \caption{Microsoft HoloLens 2, the Mixed Reality platform used in our study.}
    \label{fig:hololens-headset-appendix}
\end{figure}

The most important hardware component of a Mixed Reality system is the Head Mounted Display (HMD) or Headset. Some popular examples are Microsoft HoloLens \cite{ungureanu2020hololens} and Magic Leap. Mixed Reality headsets have a transparent visor through which the user can see the world and different kinds of optics and display elements that project the virtual objects in the field of view of the user. An elaborate set of RGB cameras and laser scanners is used for spatial mapping, localization and tracking. MR headsets also have an Inertial Measurement Unit (IMU) consisting of an accelerometer, a magnetometer and a gyroscope that tracks the movement of the user’s head through space in six degrees of freedom. Figure \ref{fig:hololens-headset-appendix} shows the components of Microsoft HoloLens 2, the MR headset that we plan to use in this thesis. MR Headsets provide multi-modal user interfaces that combine head gestures, voice commands and eye gaze.

\subsubsection{Challenges}
MR headsets must be power efficient, lightweight and ergonomic. Spatial mapping using multiple modalities, scene reconstruction, gesture recognition, voice recognition, and high-polygon rendering are some primary tasks that are at the core of most Mixed Reality experiences. These 3D Computer Vision and Computer Graphics tasks are compute-intensive and the limited on-device resources are not adequate for achieving these at a latency low enough for supporting an immersive MR experience. To mitigate this issue, most MR platforms employ a hybrid architecture where the compute-intensive tasks like those involving deep neural networks and rendering are off-loaded to high-performance and specialised hardware (e.g. GPU, FPGA) in the cloud and the on-board hardware is used only for encoding-decoding of data, communication with the cloud and interactions with the user (UX).

\subsection{Point Cloud Registration}
\subsubsection{Introduction}
Point cloud registration, also called Alignment (in MR jargon), is the task of finding a spatial transformation (such as rotation, transformation and scaling) which when applied to a source point cloud results in point-wise superposition with a destination point cloud of the same object/scene. These algorithms fall into two categories - rigid and non-rigid - depending on whether deformations in the object/scene between the two scans can be addressed.

\subsubsection{Approaches to Solution}
The general approach to point-cloud registration involves two steps: a) feature extraction, and b) feature matching and registration.

\noindent\textbf{Feature extraction}\\
Each point in a point cloud must be assigned a feature vector that represents its 3D position and context including local geometry, colour and texture \citep{weinmann2017geometric}. The first step is to define a neighbourhood for each point. Some common examples are spherical \citep{lee2002perceptual,linsen2001local} or cylindrical \citep{filin2005neighborhood,niemeyer2014contextual} neighbourhoods parameterized by radius \citep{lee2002perceptual,filin2005neighborhood} or the number of nearest neighbours by Euclidean distance \citep{linsen2001local,niemeyer2014contextual}. These parameters give us the means to control the scale at which local 3D structures must be encoded. The value of the scale parameter is usually chosen using prior knowledge \citep{weinmann2015contextual} or learned from data \citep{weinmann2015semantic, mitra2003estimating,lalonde2005scale,demantke2011dimensionality} and multi-scale approaches are also popular \citep{niemeyer2014contextual,brodu20123d,schmidt2014contextual}. Once a neighbourhood of a point is defined, feature extraction encodes the local 3D geometry to attach semantics or context to the point. Certain shape primitives can be obtained by computing the eigenvalues of a 3D structure tensor constructed using the spatial coordinates of neighbouring points \citep{jutzi2009nearest,west2004context,pauly2003multi}. Other features that are extracted are angular characteristics \citep{munoz2009contextual}, height \citep{mallet2011relevance}, moments \citep{hackel2016fast}, surface properties, slope, vertical profiles and 2D projections \citep{guo2015classification}, shape distributions \citep{osada2002shape,blomley2016classification}, point-feature histograms \citep{rusu2009fast}. 

Modern approaches use Deep Neural Network \citep{goodfellow2016deep} based representation learning. Projection Networks project the 3D point cloud onto 2D image planes from multiple view-points and use 2D Convolutional Neural Networks (2D-CNN) \citep{goodfellow2016deep} to process them \citep{su2015multi,boulch2017unstructured,lawin2017deep}. Voxel-based methods project the point cloud onto a 3D grid \citep{maturana2015voxnet,roynard2018classification,ben20183dmfv}. Sparse data structures like octree and hash-maps are used for better efficiency and larger context sizes \citep{riegler2017octnet,graham20183d}. These grids are further processed using 3D Convolutional Neural Networks (3D-CNN) \citep{goodfellow2016deep}. The main drawbacks of these approaches arise from the loss of details in the original point cloud structure during projection onto grids \citep{thomas2019kpconv}. Graph Convolutional Networks address this problem by retaining the original position of each point and combining features on local surface patches \citep{verma2018feastnet,wang2019dynamic}. However, their representation is invariant to the deformations of those patches in Euclidean space which is not helpful for estimating non-rigid transformations between point clouds \citep{thomas2019kpconv}. Pointwise Networks like PointNet \citep{qi2017pointnet} apply a shared neural network to each point followed by global max-pooling. This approach set new benchmarks in point-cloud classification and different variants using Multi-Layer Perceptron (MLP) [\citep{qi2017pointnet++}, \citep{liu2019point2sequence,li2018so} and CNN \citep{hua2018pointwise,xu2018spidercnn,groh2018flex,atzmon2018point}.

Kernel Point Convolution (KPConv) \citep{thomas2019kpconv} is one of the recent breakthroughs in point cloud representation learning and is particularly suitable for deformable point clouds. It learns a kernel function to compute pointwise filters and increases representation power using deformable kernels. Instead of a grid-shaped kernel (as is the case with regular 2D and 3D CNNs), the kernel points in KPConv are spread freely in space. Each kernel point accumulates the features of the point-cloud points within a spherical neighbourhood around itself with weights that decay as the points get farther. In deformable KPConv, each kernel point also has a learnable offset that allows it to learn to adapt the shape of a kernel to different inputs. We will use KPConv to learn representations of our point clouds in the experiments for this thesis.

\noindent\textbf{Feature matching and registration}\\
After the points have been represented using feature vectors, the next task is to find corresponding points between the given pair of point clouds. Traditional methods use different variants of the RANdom-SAmple Consensus (RANSAC) algorithm for finding matching points \citep{holz2015registration,schnabel2007efficient}. RANSAC is an iterative algorithm and works by classifying all possible correspondences into inliers and outliers. RANSAC is simple yet powerful and a bulk of the literature on point cloud registration uses RANSAC with both handcrafted and learned features \citep{ao2021spinnet,bai2020d3feat,choy2019fully,deng2018ppfnet,deng2018ppf,el2021unsupervisedr,yu2021cofinet,zeng20173dmatch}. For rigid point clouds, another popular algorithm is Iterative Closest Point (ICP) \citep{arun1987least,besl1992method,zhang1994iterative}. ICP assumes that the two point clouds are spatially close. In each iteration, for each point in the source point cloud, the closest point in the destination cloud is assigned as the matching point. Then the rigid transformation (rotation and translation) is computed by minimising mean-square loss. These steps are repeated till convergence. Direct Registration approaches combine feature extraction, feature matching and registration within a single architecture through end-to-end pose optimization \citep{besl1992method,zhou2016fast,aoki2019pointnetlk,choy2020deep}. 

Non-rigid correspondence has the added challenge of accounting for deformations. Different approaches have been proposed including projective correspondence \citep{newcombe2015dynamicfusion}, Siamese Network \citep{schmidt2016self} and Scene Flow estimation \citep{li2021neural,liu2019flownet3d,puy2020flot}. Non-rigid correspondence is also studied in Geometry Processing where the input data is in the form of manifold surfaces. Methods like isometric deformation \citep{huang2008non}, latent code optimization \citep{groueix20183d} and functional maps \citep{ovsjanikov2012functional} have been studied in this context.

\subsubsection{Challenges}
\noindent\textbf{Lack of ordering or structure}\\
A point cloud is a collection of points in 3D space with no ordering or structure (like a grid). The absence of a grid structure makes it difficult to apply traditional deep learning methods like CNNs because they leverage the grid structure of their inputs (images/videos) for learning representations. As described in the previous section, there have been attempts \citep{lawin2017deep,maturana2015voxnet,roynard2018classification,ben20183dmfv}, to coerce/quantize/project point clouds to grids in order to harness the power of traditional CNNs while trading off the loss of the inherent structural details of the point cloud. New forms of convolution like KPConv \citep{thomas2019kpconv} also attempt to address this challenge with flexible non-grid kernels.

\begin{figure}[!h]
    \centering
    \includegraphics[width=0.7\textwidth]{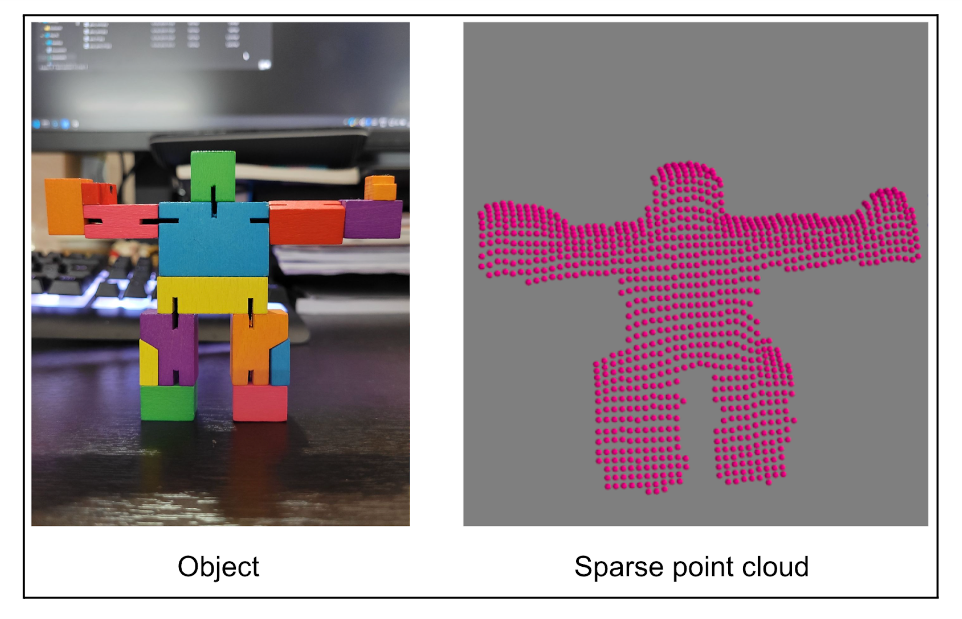}
    \caption{Example of sparse point cloud scan}
    \label{fig:sparse-pc-appendix}
\end{figure}

\noindent\textbf{Sparsity}\\
Most real-world 3D scans tend to be sparse \ref{fig:sparse-pc-appendix}. As a result of this, a large amount of computing gets wasted processing zero entries. For efficient processing of sparse point clouds using Deep Neural Networks, the recently proposed Minkowski Engine \citep{choy20194d} has proved to be a game changer. It represents point clouds as a position-indexed array and performs computation only for invalid regions. It can also leverage specialist hardware like GPU for greater throughput.

\begin{figure}[!h]
    \centering
    \includegraphics[width=0.5\textwidth]{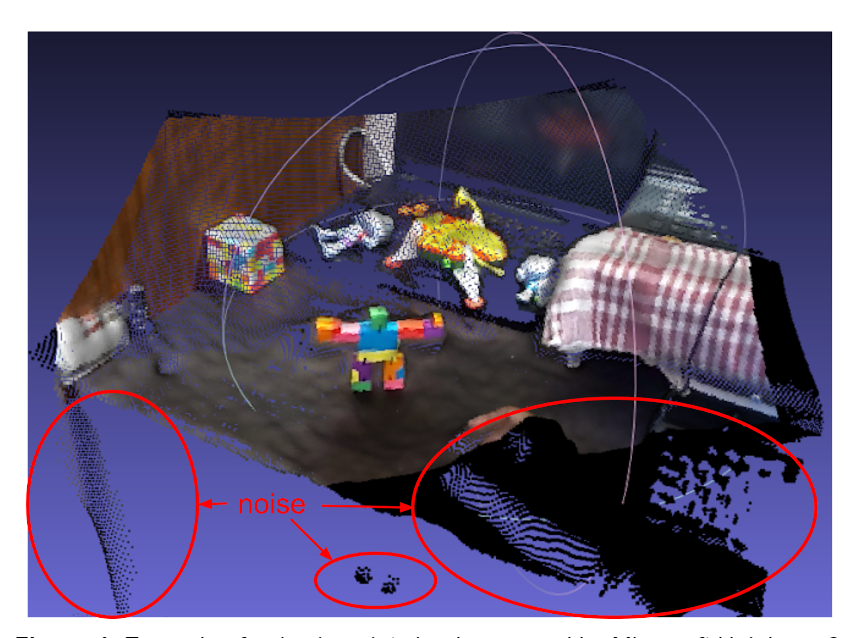}
    \caption{Example of noise in point clouds scanned by Microsoft HoloLens 2.}
    \label{fig:noisy-pc-appendix}
\end{figure}

\noindent\textbf{Noise and Outliers}\\
Real-world 3D scans can be dirty with a large amount of noise and outliers \ref{fig:noisy-pc-appendix}. These are points that do not belong to the intended surface. One of the sources of outliers is a specular reflection from shiny surfaces like metal. In rigid registration, the RANSAC algorithm \citep{holz2015registration,schnabel2007efficient} attempts to remove the outliers that do not conform with the model with maximum consensus. Learning approaches to outlier detection and removal include \citep{bai2021pointdsc,pais20203dregnet}.

\begin{figure}[!h]
    \centering
    \includegraphics[width=\textwidth]{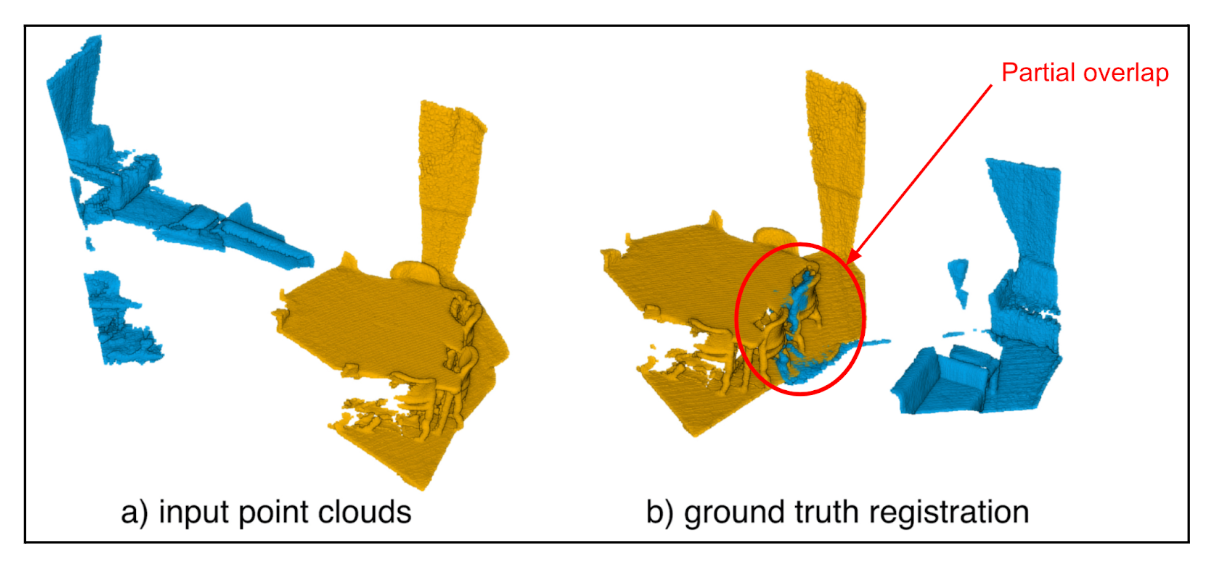}
    \caption{Example of partial overlap between the source and target point clouds}
    \label{fig:partial-overlap-appendix}
\end{figure}

\noindent\textbf{Partial Overlap}\\
In a real-world use case of point cloud registration, the source and target point clouds may only have a partial match \ref{fig:partial-overlap-appendix} albeit the fact that they are of the same object. This might be due to occlusion, object motion, or viewpoint change. A variety of methods have been proposed to perform the registration of point clouds that have a partial match \citep{thomas2019kpconv,xu2021omnet, attaiki2021dpfm, rodola2017partial}.

\section{Problem Statement}
\subsection{Definition}
We use the same notation as the authors of Lepard in our paper. Let $\mathbf{S} \in \mathbb{R}^{n\times 3}$ be the source point cloud (with $n$ points) and $\mathbf{T} \in \mathbb{R}^{m\times 3}$ be the target point cloud (with $m$ points). Our goal is to calculate a warp function $\mathcal{W}: \mathbb{R}\rightarrow \mathbb{R}$ that maps $\mathbf{S}$ to $\mathbf{T}$. For an arbitrary non-rigid deformation, $\mathcal{W}$ generalizes to a dense per-point warp field. For the rigid case, $\mathcal{W}$ can be parameterized as a $SE(3)$ transformation which is a single rotation and a single transformation applied to the whole body of $\mathbf{S}$. 

\subsection{Robust real-world performance}
The system must be robust to real-world test conditions that contain noise and occlusion. For evaluation, we will collect data in cluttered environments using Microsoft Azure Kinect and Microsoft HoloLens 2.

\section{Methodology}
In this section, we describe our proposed research methodology.

\subsection{Baseline Algorithm}
We choose Lepard \citep{li2022lepard} as our baseline algorithm as it is one of the current state-of-the-art methods for partial non-rigid point cloud registration. The algorithm uses powerful KPConv \citep{thomas2019kpconv} for learning deformable point cloud representations. In order to add positional information to the translation invariant KPConv representation, the authors use Rotary Positional Encoding \citep{su2021roformer}. For point cloud matching it uses RANSAC algorithm \citep{holz2015registration,schnabel2007efficient} for rigid objects and deformable ICP for non-rigid objects. A transformer is used for position-aware feature matching and Rigid Fitting with Soft Prucrustes \citep{arun1987least} is used to obtain the final Rotation and Translation values for the transformation/warp function.

\subsection{Training Objectives}
The following are the loss functions used to train Lepard \citep{li2022lepard}:

\begin{enumerate}
    \item \textbf{Matching Loss:} The Focal Loss over the confidence matrix returned by the matching layer is minimised.
    \item \textbf{Warping Loss:} This is the L1 Loss between the target point cloud and the source point cloud warped by the Rotation and Translation matrices returned by the Procrustes layer.
\end{enumerate}

\subsection{Evaluation Metrics}
The following are the metrics used for benchmarking point cloud registration algorithms:

\begin{enumerate}
    \item \textbf{Inlier ratio (IR).} Inlier Ratio is the fraction of correct matches in the predicted correspondences set.
    \item \textbf{Non-rigid Feature Matching Recall (NFMR).} NFMR measures the fraction of ground-truth matches that were successfully recovered by the correspondence prediction algorithm.
\end{enumerate}

\subsection{Datasets}
We will consider training and fine-tuning our models on the following datasets.
3DMatch \citep{zeng20173dmatch} and 3DLoMatch \citep{huang2021predator} : These are datasets for rigid-body partial point cloud registration consisting of indoor scans.
4DMatch and 4DLoMatch \citep{li2022lepard}: These are datasets for non-rigid partial point cloud registration derived from DeforminObjects4d \citep{li20214dcomplete}, a dataset of animated characters.

\subsection{Real World Evaluation}
\begin{figure}[!h]
    \centering
    \includegraphics[width=0.8\textwidth]{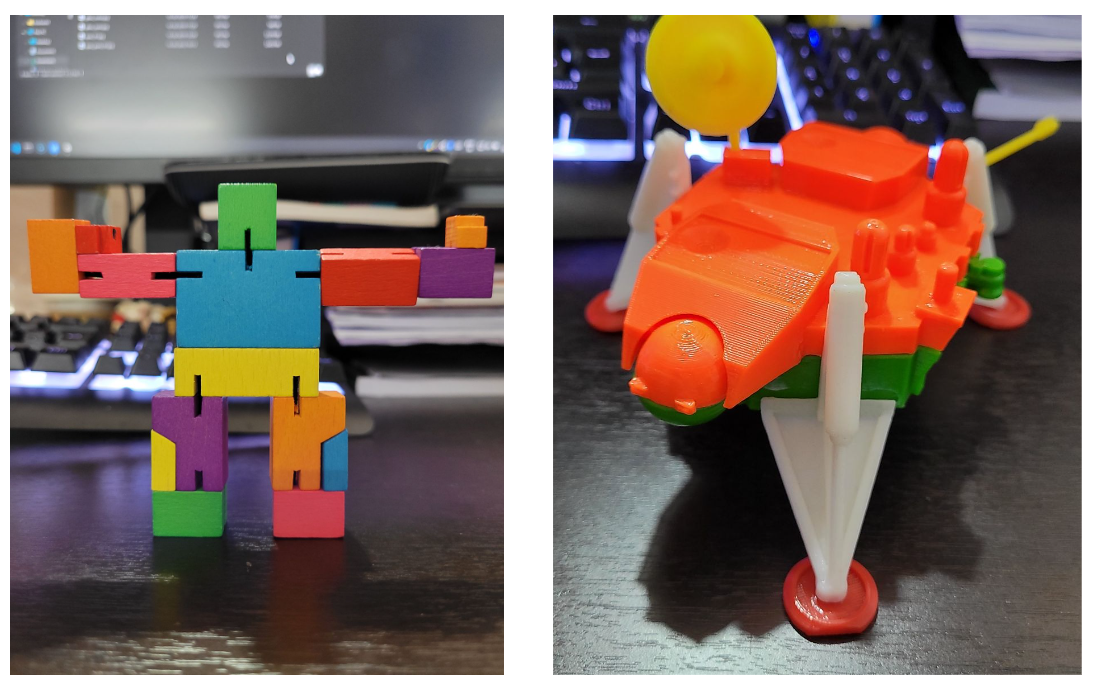}
    \caption{Real-world objects used for our study. (Left) Cubebot robot model. (Right) Viking NASA Lander model.}
    \label{fig:Lander-ROBO-appendix}
\end{figure}

\begin{figure}[!h]
    \centering
    \includegraphics[width=0.6\textwidth]{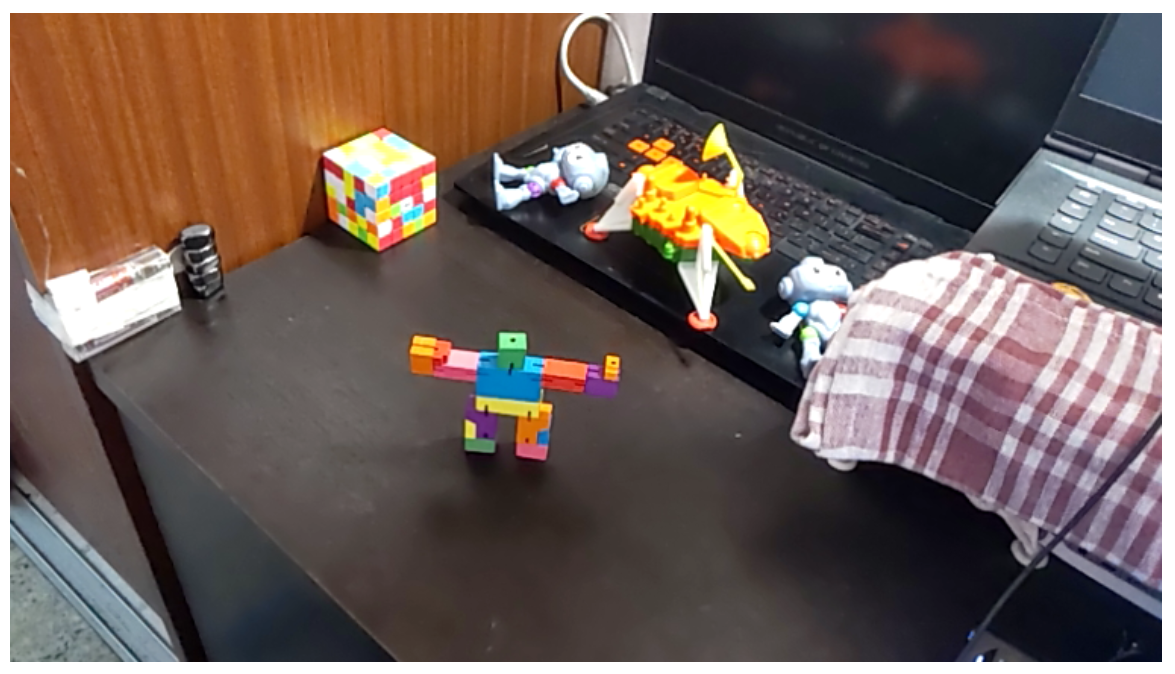}
    \caption{Real-world environment (cluttered) used in our tests.}
    \label{fig:real-world-cluttered-appendix}
\end{figure}

4DMatch and 4DLoMatch \citep{li2022lepard} are composed of animated objects. However, for real world use cases in Mixed Reality, Robotics, etc. we must evaluate the performance of non-rigid registration on noisy scans of real deformable objects. Figure \ref{fig:Lander-ROBO-appendix} shows a few test objects that we plan to use for evaluation in this project. Figure \ref{fig:real-world-cluttered-appendix} shows our test environment that is cluttered and represents a typical real-world usage environment of Mixed Reality devices.

\section{Expected Outcome}
Point cloud registration for non-rigid partial point clouds is a highly challenging computational problem. Although we focus on Mixed Reality applications in this thesis, the same approaches have profound significance in Indoor and Enterprise Robotics, Autonomous Driving, Unmanned Aerial Vehicles and Computer Graphics. Mixed Reality systems with non-rigid point cloud registration capabilities can be used for remote maintenance and repair of deformable precision engineering components like cables, hoses and valves \citep{abate2013mixed} and in medical science for rendering 3D diagnostic scans like CT and MRI over deformable and dynamic organs during surgery \citep{kersten2013state} - among others.

In this thesis we will implement a state-of-the-art point cloud registration algorithm - Lepard \citep{li2022lepard} - on Microsoft HoloLens 2 Mixed Reality Platform \citep{ungureanu2020hololens} and study the performance of the integrated system on a wide range of objects in real-world cluttered environments. We will study the failure modes and potential bottlenecks in the inference pipeline affecting latency and investigate the root causes. This study will enable us to develop an improved version of Lepard that is more robust to outliers and noise and has a fast inference pipeline for supporting real-time Mixed Reality applications.

\section{Required Resources}
The required resources for this thesis can be grouped into two categories: hardware and software.

\subsection{Hardware}
Training and inference using Deep Neural Networks with the Lepard architecture \citep{li2022lepard} in a reasonable amount of time will require access to Graphics Processing Units or GPUs. I have access to two high-performance laptop computers with one GPU each. The following are the details:

\begin{itemize}
    \item 2020 Lenovo Legion with RTX-2080 GPU
    \item 2016 Asus Republic Of Gamers with GTX-1080 GPU
\end{itemize}

I also have access to a blade server with Intel Xeon processors if any additional requirements arise during the project.

Since this thesis focuses on non-rigid alignment for Mixed Reality applications, I will need access to a MR headset. I have access to a 2022 Microsoft HoloLens 2 Enterprise Edition headset. For scanning 3D point clouds, I will use my HoloLens 2 and a Microsoft Azure Kinect camera that I also have access to.

\subsection{Software}
For implementing Lepard \citep{li2022lepard}, I will use the GitHub repository: \url{https://github.com/rabbityl/lepard} as starter code. Running point cloud registration on HoloLens 2 will require access to the HoloLens 2 Research Mode \citep{ungureanu2020hololens} that I also have multiple years of experience in. 

For evaluation of the quality of alignment, I will use CloudCompare \citep{girardeau2016cloudcompare} and HoloLens 2 Spectator View that I also have access to.

\section{Plan of Work}

Table \ref{table:Gantt-Chart} is a Gantt Chart with a time plan for the project considering November 25th, 2022 as the deadline for thesis submission.

%
\begin{table}[]
\centering
\tiny
\caption{Gantt Chart showing the plan of work for this thesis.}
\label{table:Gantt-Chart}
\begin{tabular}{|p{6cm}|l|l|l|l|}
\hline
\multicolumn{1}{|c|}{\textbf{Task Name}} &
  \multicolumn{1}{c|}{\textbf{August}} &
  \multicolumn{1}{c|}{\textbf{September}} &
  \multicolumn{1}{c|}{\textbf{October}} &
  \multicolumn{1}{c|}{\textbf{November}} \\ \hline
Implementation of Lepard on HoloLens 2.       & \cellcolor[HTML]{34FF34} &                          &                          &                          \\ \hline
Evaluation of rigid and non-rigid point cloud registration on custom data (HoloLens 2) &
   &
  \cellcolor[HTML]{34FF34} &
  \cellcolor[HTML]{34FF34} &
   \\ \hline
Investigation of failure modes and mitigation &                          &                          & \cellcolor[HTML]{34FF34} & \cellcolor[HTML]{34FF34} \\ \hline
Thesis writing                                &                          & \cellcolor[HTML]{34FF34} & \cellcolor[HTML]{34FF34} & \cellcolor[HTML]{34FF34} \\ \hline
\end{tabular}
\end{table}

\subsection{Contingency Plan}
The baseline algorithm that we have chosen - Lepard \citep{li2022lepard} - is state-of-the-art. However, implementing it might take longer than expected. Since our training setup is different from the authors’, our results might not be in line with the ones that have been reported by the authors in the paper and this is a common problem with a lot of modern deep neural network based works. In that case, our contingency plan is to use a functional map based state-of-the-art baseline called SyNoRiM \citep{huang2022multiway} that we already have working on our HoloLens 2 setup.

%% file: Thesis.bbl
\begin{thebibliography}{}

\bibitem[\protect\citeauthoryear{Abate, Narducci \& Ricciardi}{Abate
  et~al.}{2013}]{abate2013mixed}
Abate A.~F.,  Narducci F.,    Ricciardi S.,  2013, in International Conference
  on Virtual, Augmented and Mixed Reality Mixed reality environment for mission
  critical systems servicing and repair.
pp 201--210

\bibitem[\protect\citeauthoryear{Ao, Hu, Yang, Markham \& Guo}{Ao
  et~al.}{2021}]{ao2021spinnet}
Ao S.,  Hu Q.,  Yang B.,  Markham A.,    Guo Y.,  2021, in Proceedings of the
  IEEE/CVF Conference on Computer Vision and Pattern Recognition Spinnet:
  Learning a general surface descriptor for 3d point cloud registration.
pp 11753--11762

\bibitem[\protect\citeauthoryear{Aoki, Goforth, Srivatsan \& Lucey}{Aoki
  et~al.}{2019}]{aoki2019pointnetlk}
Aoki Y.,  Goforth H.,  Srivatsan R.~A.,    Lucey S.,  2019, in Proceedings of
  the IEEE/CVF conference on computer vision and pattern recognition
  Pointnetlk: Robust \& efficient point cloud registration using pointnet.
pp 7163--7172

\bibitem[\protect\citeauthoryear{Arun, Huang \& Blostein}{Arun
  et~al.}{1987}]{arun1987least}
Arun K.~S.,  Huang T.~S.,    Blostein S.~D.,  1987, IEEE Transactions on
  pattern analysis and machine intelligence, pp 698--700

\bibitem[\protect\citeauthoryear{Attaiki, Pai \& Ovsjanikov}{Attaiki
  et~al.}{2021}]{attaiki2021dpfm}
Attaiki S.,  Pai G.,    Ovsjanikov M.,  2021, in 2021 International Conference
  on 3D Vision (3DV) Dpfm: Deep partial functional maps.
pp 175--185

\bibitem[\protect\citeauthoryear{Atzmon, Maron \& Lipman}{Atzmon
  et~al.}{2018}]{atzmon2018point}
Atzmon M.,  Maron H.,    Lipman Y.,  2018, arXiv preprint arXiv:1803.10091

\bibitem[\protect\citeauthoryear{Bai, Luo, Zhou, Chen, Li, Hu, Fu \& Tai}{Bai
  et~al.}{2021}]{bai2021pointdsc}
Bai X.,  Luo Z.,  Zhou L.,  Chen H.,  Li L.,  Hu Z.,  Fu H.,    Tai C.-L.,
  2021, in Proceedings of the IEEE/CVF Conference on Computer Vision and
  Pattern Recognition Pointdsc: Robust point cloud registration using deep
  spatial consistency.
pp 15859--15869

\bibitem[\protect\citeauthoryear{Bai, Luo, Zhou, Fu, Quan \& Tai}{Bai
  et~al.}{2020}]{bai2020d3feat}
Bai X.,  Luo Z.,  Zhou L.,  Fu H.,  Quan L.,    Tai C.-L.,  2020, in
  Proceedings of the IEEE/CVF conference on computer vision and pattern
  recognition D3feat: Joint learning of dense detection and description of 3d
  local features.
pp 6359--6367

\bibitem[\protect\citeauthoryear{Ben-Shabat, Lindenbaum \& Fischer}{Ben-Shabat
  et~al.}{2018}]{ben20183dmfv}
Ben-Shabat Y.,  Lindenbaum M.,    Fischer A.,  2018, IEEE Robotics and
  Automation Letters, 3, 3145

\bibitem[\protect\citeauthoryear{Besl \& McKay}{Besl \&
  McKay}{1992}]{besl1992method}
Besl P.~J.,  McKay N.~D.,  1992, in Sensor fusion IV: control paradigms and
  data structures Vol.~1611, Method for registration of 3-d shapes.
pp 586--606

\bibitem[\protect\citeauthoryear{Blomley, Jutzi \& Weinmann}{Blomley
  et~al.}{2016}]{blomley2016classification}
Blomley R.,  Jutzi B.,    Weinmann M.,  2016, ISPRS Annals of Photogrammetry,
  Remote Sensing \& Spatial Information Sciences, 3

\bibitem[\protect\citeauthoryear{Borsci, Lawson \& Broome}{Borsci
  et~al.}{2015}]{borsci2015empirical}
Borsci S.,  Lawson G.,    Broome S.,  2015, Computers in Industry, 67, 17

\bibitem[\protect\citeauthoryear{Boulch, Le~Saux \& Audebert}{Boulch
  et~al.}{2017}]{boulch2017unstructured}
Boulch A.,  Le~Saux B.,    Audebert N.,  2017, 3dor@ eurographics, 3, 1

\bibitem[\protect\citeauthoryear{Brodu \& Lague}{Brodu \&
  Lague}{2012}]{brodu20123d}
Brodu N.,  Lague D.,  2012, ISPRS journal of photogrammetry and remote sensing,
  68, 121

\bibitem[\protect\citeauthoryear{Chen, Day, Tang \& John}{Chen
  et~al.}{2017}]{chen2017recent}
Chen L.,  Day T.~W.,  Tang W.,    John N.~W.,  2017, in 2017 IEEE international
  symposium on mixed and augmented reality (ISMAR) Recent developments and
  future challenges in medical mixed reality.
pp 123--135

\bibitem[\protect\citeauthoryear{Choy, Dong \& Koltun}{Choy
  et~al.}{2020}]{choy2020deep}
Choy C.,  Dong W.,    Koltun V.,  2020, in Proceedings of the IEEE/CVF
  conference on computer vision and pattern recognition Deep global
  registration.
pp 2514--2523

\bibitem[\protect\citeauthoryear{Choy, Gwak \& Savarese}{Choy
  et~al.}{2019}]{choy20194d}
Choy C.,  Gwak J.,    Savarese S.,  2019, in Proceedings of the IEEE/CVF
  Conference on Computer Vision and Pattern Recognition 4d spatio-temporal
  convnets: Minkowski convolutional neural networks.
pp 3075--3084

\bibitem[\protect\citeauthoryear{Choy, Park \& Koltun}{Choy
  et~al.}{2019}]{choy2019fully}
Choy C.,  Park J.,    Koltun V.,  2019, in Proceedings of the IEEE/CVF
  International Conference on Computer Vision Fully convolutional geometric
  features.
pp 8958--8966

\bibitem[\protect\citeauthoryear{Cignoni, Ranzuglia, Callieri, Corsini,
  Ganovelli, Pietroni \& Tarini}{Cignoni et~al.}{2011}]{cignoni2011meshlab}
Cignoni P.,  Ranzuglia G.,  Callieri M.,  Corsini M.,  Ganovelli F.,  Pietroni
  N.,    Tarini M.,  2011

\bibitem[\protect\citeauthoryear{Demantk{\'e}, Mallet, David \&
  Vallet}{Demantk{\'e} et~al.}{2011}]{demantke2011dimensionality}
Demantk{\'e} J.,  Mallet C.,  David N.,    Vallet B.,  2011, in Laserscanning
  Dimensionality based scale selection in 3d lidar point clouds

\bibitem[\protect\citeauthoryear{Deng, Birdal \& Ilic}{Deng
  et~al.}{2018a}]{deng2018ppf}
Deng H.,  Birdal T.,    Ilic S.,  2018a, in Proceedings of the European
  Conference on Computer Vision (ECCV) Ppf-foldnet: Unsupervised learning of
  rotation invariant 3d local descriptors.
pp 602--618

\bibitem[\protect\citeauthoryear{Deng, Birdal \& Ilic}{Deng
  et~al.}{2018b}]{deng2018ppfnet}
Deng H.,  Birdal T.,    Ilic S.,  2018b, in Proceedings of the IEEE conference
  on computer vision and pattern recognition Ppfnet: Global context aware local
  features for robust 3d point matching.
pp 195--205

\bibitem[\protect\citeauthoryear{Durrant-Whyte \& Bailey}{Durrant-Whyte \&
  Bailey}{2006}]{durrant2006simultaneous}
Durrant-Whyte H.,  Bailey T.,  2006, IEEE robotics \& automation magazine, 13,
  99

\bibitem[\protect\citeauthoryear{El~Banani, Gao \& Johnson}{El~Banani
  et~al.}{2021}]{el2021unsupervisedr}
El~Banani M.,  Gao L.,    Johnson J.,  2021, in Proceedings of the IEEE/CVF
  Conference on Computer Vision and Pattern Recognition Unsupervisedr\&r:
  Unsupervised point cloud registration via differentiable rendering.
pp 7129--7139

\bibitem[\protect\citeauthoryear{Ester, Kriegel, Sander, Xu et~al.,}{Ester
  et~al.}{1996}]{ester1996density}
Ester M.,  Kriegel H.-P.,  Sander J.,  Xu X.,    et~al., 1996, in kdd Vol.~96,
  A density-based algorithm for discovering clusters in large spatial databases
  with noise..
pp 226--231

\bibitem[\protect\citeauthoryear{Filin \& Pfeifer}{Filin \&
  Pfeifer}{2005}]{filin2005neighborhood}
Filin S.,  Pfeifer N.,  2005, Photogrammetric Engineering \& Remote Sensing,
  71, 743

\bibitem[\protect\citeauthoryear{Girardeau-Montaut}{Girardeau-Montaut}{2016}]{girardeau2016cloudcompare}
Girardeau-Montaut D.,  2016, France: EDF R\&D Telecom ParisTech, 11

\bibitem[\protect\citeauthoryear{Glasmachers}{Glasmachers}{2017}]{glasmachers2017limits}
Glasmachers T.,  2017, in Asian conference on machine learning Limits of
  end-to-end learning.
pp 17--32

\bibitem[\protect\citeauthoryear{Goodfellow, Bengio \& Courville}{Goodfellow
  et~al.}{2016}]{goodfellow2016deep}
Goodfellow I.,  Bengio Y.,    Courville A.,  2016, Deep learning.
MIT press

\bibitem[\protect\citeauthoryear{Graham, Engelcke \& Van Der~Maaten}{Graham
  et~al.}{2018}]{graham20183d}
Graham B.,  Engelcke M.,    Van Der~Maaten L.,  2018, in Proceedings of the
  IEEE conference on computer vision and pattern recognition 3d semantic
  segmentation with submanifold sparse convolutional networks.
pp 9224--9232

\bibitem[\protect\citeauthoryear{Groh, Wieschollek \& Lensch}{Groh
  et~al.}{2018}]{groh2018flex}
Groh F.,  Wieschollek P.,    Lensch H.,  2018, in Asian Conference on Computer
  Vision Flex-convolution.
pp 105--122

\bibitem[\protect\citeauthoryear{Groueix, Fisher, Kim, Russell \&
  Aubry}{Groueix et~al.}{2018}]{groueix20183d}
Groueix T.,  Fisher M.,  Kim V.~G.,  Russell B.~C.,    Aubry M.,  2018, in
  Proceedings of the European Conference on Computer Vision (ECCV) 3d-coded: 3d
  correspondences by deep deformation.
pp 230--246

\bibitem[\protect\citeauthoryear{Guo, Huang, Zhang \& Sohn}{Guo
  et~al.}{2015}]{guo2015classification}
Guo B.,  Huang X.,  Zhang F.,    Sohn G.,  2015, ISPRS Journal of
  Photogrammetry and Remote Sensing, 100, 71

\bibitem[\protect\citeauthoryear{Hackel, Wegner \& Schindler}{Hackel
  et~al.}{2016}]{hackel2016fast}
Hackel T.,  Wegner J.~D.,    Schindler K.,  2016, ISPRS annals of the
  photogrammetry, remote sensing and spatial information sciences, 3, 177

\bibitem[\protect\citeauthoryear{Holz, Ichim, Tombari, Rusu \& Behnke}{Holz
  et~al.}{2015}]{holz2015registration}
Holz D.,  Ichim A.~E.,  Tombari F.,  Rusu R.~B.,    Behnke S.,  2015, IEEE
  Robotics \& Automation Magazine, 22, 110

\bibitem[\protect\citeauthoryear{Hua, Tran \& Yeung}{Hua
  et~al.}{2018}]{hua2018pointwise}
Hua B.-S.,  Tran M.-K.,    Yeung S.-K.,  2018, in Proceedings of the IEEE
  conference on computer vision and pattern recognition Pointwise convolutional
  neural networks.
pp 984--993

\bibitem[\protect\citeauthoryear{Huang, Birdal, Gojcic, Guibas \& Hu}{Huang
  et~al.}{2022}]{huang2022multiway}
Huang J.,  Birdal T.,  Gojcic Z.,  Guibas L.~J.,    Hu S.-M.,  2022, IEEE
  Transactions on Pattern Analysis and Machine Intelligence

\bibitem[\protect\citeauthoryear{Huang, Adams, Wicke \& Guibas}{Huang
  et~al.}{2008}]{huang2008non}
Huang Q.-X.,  Adams B.,  Wicke M.,    Guibas L.~J.,  2008, in Computer Graphics
  Forum Vol.~27, Non-rigid registration under isometric deformations.
pp 1449--1457

\bibitem[\protect\citeauthoryear{Huang, Gojcic, Usvyatsov, Wieser \&
  Schindler}{Huang et~al.}{2021}]{huang2021predator}
Huang S.,  Gojcic Z.,  Usvyatsov M.,  Wieser A.,    Schindler K.,  2021, in
  Proceedings of the IEEE/CVF Conference on computer vision and pattern
  recognition Predator: Registration of 3d point clouds with low overlap.
pp 4267--4276

\bibitem[\protect\citeauthoryear{Huang, Alem, Tecchia \& Duh}{Huang
  et~al.}{2018}]{huang2018augmented}
Huang W.,  Alem L.,  Tecchia F.,    Duh H. B.-L.,  2018, Journal on Multimodal
  User Interfaces, 12, 77

\bibitem[\protect\citeauthoryear{Hunter}{Hunter}{2007}]{hunter2007matplotlib}
Hunter J.~D.,  2007, Computing in science \& engineering, 9, 90

\bibitem[\protect\citeauthoryear{Jain \& Werth}{Jain \&
  Werth}{2019}]{jain2019current}
Jain S.,  Werth D.,  2019, in International Conference on Human-Computer
  Interaction Current state of mixed reality technology for digital retail: a
  literature review.
pp 22--37

\bibitem[\protect\citeauthoryear{Jerald, Giokaris, Woodall, Hartbolt, Chandak
  \& Kuntz}{Jerald et~al.}{2014}]{jerald2014developing}
Jerald J.,  Giokaris P.,  Woodall D.,  Hartbolt A.,  Chandak A.,    Kuntz S.,
  2014, in 2014 IEEE Virtual Reality (VR) Developing virtual reality
  applications with unity.
pp~1--3

\bibitem[\protect\citeauthoryear{Jutzi \& Gross}{Jutzi \&
  Gross}{2009}]{jutzi2009nearest}
Jutzi B.,  Gross H.,  2009, The International Archives of the Photogrammetry,
  Remote Sensing and Spatial Information Sciences, 38, 4

\bibitem[\protect\citeauthoryear{Kent, Snider, Gopsill \& Hicks}{Kent
  et~al.}{2021}]{kent2021mixed}
Kent L.,  Snider C.,  Gopsill J.,    Hicks B.,  2021, Design Studies, 77,
  101046

\bibitem[\protect\citeauthoryear{Kersten-Oertel, Jannin \&
  Collins}{Kersten-Oertel et~al.}{2013}]{kersten2013state}
Kersten-Oertel M.,  Jannin P.,    Collins D.~L.,  2013, Computerized Medical
  Imaging and Graphics, 37, 98

\bibitem[\protect\citeauthoryear{Kirkley \& Kirkley}{Kirkley \&
  Kirkley}{2005}]{kirkley2005creating}
Kirkley S.~E.,  Kirkley J.~R.,  2005, TechTrends, 49, 42

\bibitem[\protect\citeauthoryear{Ladwig \& Geiger}{Ladwig \&
  Geiger}{2018}]{ladwig2018literature}
Ladwig P.,  Geiger C.,  2018, in International Conference on Remote Engineering
  and Virtual Instrumentation A literature review on collaboration in mixed
  reality.
pp 591--600

\bibitem[\protect\citeauthoryear{Lalonde, Unnikrishnan, Vandapel \&
  Hebert}{Lalonde et~al.}{2005}]{lalonde2005scale}
Lalonde J.-F.,  Unnikrishnan R.,  Vandapel N.,    Hebert M.,  2005, in Fifth
  International Conference on 3-D Digital Imaging and Modeling (3DIM'05) Scale
  selection for classification of point-sampled 3d surfaces.
pp 285--292

\bibitem[\protect\citeauthoryear{Lawin, Danelljan, Tosteberg, Bhat, Khan \&
  Felsberg}{Lawin et~al.}{2017}]{lawin2017deep}
Lawin F.~J.,  Danelljan M.,  Tosteberg P.,  Bhat G.,  Khan F.~S.,    Felsberg
  M.,  2017, in International Conference on Computer Analysis of Images and
  Patterns Deep projective 3d semantic segmentation.
pp 95--107

\bibitem[\protect\citeauthoryear{Lee \& Schenk}{Lee \&
  Schenk}{2002}]{lee2002perceptual}
Lee I.,  Schenk T.,  2002, International Archives of Photogrammetry Remote
  Sensing and Spatial Information Sciences, 34, 193

\bibitem[\protect\citeauthoryear{Li, Sumner \& Pauly}{Li
  et~al.}{2008}]{li2008global}
Li H.,  Sumner R.~W.,    Pauly M.,  2008, in Computer graphics forum Vol.~27,
  Global correspondence optimization for non-rigid registration of depth scans.
pp 1421--1430

\bibitem[\protect\citeauthoryear{Li, Chen \& Lee}{Li et~al.}{2018}]{li2018so}
Li J.,  Chen B.~M.,    Lee G.~H.,  2018, in Proceedings of the IEEE conference
  on computer vision and pattern recognition So-net: Self-organizing network
  for point cloud analysis.
pp 9397--9406

\bibitem[\protect\citeauthoryear{Li, Kaesemodel~Pontes \& Lucey}{Li
  et~al.}{2021}]{li2021neural}
Li X.,  Kaesemodel~Pontes J.,    Lucey S.,  2021, Advances in Neural
  Information Processing Systems, 34, 7838

\bibitem[\protect\citeauthoryear{Li \& Harada}{Li \&
  Harada}{2022}]{li2022lepard}
Li Y.,  Harada T.,  2022, in Proceedings of the IEEE/CVF Conference on Computer
  Vision and Pattern Recognition Lepard: Learning partial point cloud matching
  in rigid and deformable scenes.
pp 5554--5564

\bibitem[\protect\citeauthoryear{Li, Takehara, Taketomi, Zheng \&
  Nie{\ss}ner}{Li et~al.}{2021}]{li20214dcomplete}
Li Y.,  Takehara H.,  Taketomi T.,  Zheng B.,    Nie{\ss}ner M.,  2021, in
  Proceedings of the IEEE/CVF International Conference on Computer Vision
  4dcomplete: Non-rigid motion estimation beyond the observable surface.
pp 12706--12716

\bibitem[\protect\citeauthoryear{Linsen \& Prautzsch}{Linsen \&
  Prautzsch}{2001}]{linsen2001local}
Linsen L.,  Prautzsch H.,  2001, in Eurographics (Short Presentations) Local
  versus global triangulations.

\bibitem[\protect\citeauthoryear{Liu, Han, Liu \& Zwicker}{Liu
  et~al.}{2019}]{liu2019point2sequence}
Liu X.,  Han Z.,  Liu Y.-S.,    Zwicker M.,  2019, in Proceedings of the AAAI
  Conference on Artificial Intelligence Vol.~33, Point2sequence: Learning the
  shape representation of 3d point clouds with an attention-based sequence to
  sequence network.
pp 8778--8785

\bibitem[\protect\citeauthoryear{Liu, Qi \& Guibas}{Liu
  et~al.}{2019}]{liu2019flownet3d}
Liu X.,  Qi C.~R.,    Guibas L.~J.,  2019, in Proceedings of the IEEE/CVF
  conference on computer vision and pattern recognition Flownet3d: Learning
  scene flow in 3d point clouds.
pp 529--537

\bibitem[\protect\citeauthoryear{Mallet, Bretar, Roux, Soergel \&
  Heipke}{Mallet et~al.}{2011}]{mallet2011relevance}
Mallet C.,  Bretar F.,  Roux M.,  Soergel U.,    Heipke C.,  2011, ISPRS
  journal of photogrammetry and remote sensing, 66, S71

\bibitem[\protect\citeauthoryear{Maturana \& Scherer}{Maturana \&
  Scherer}{2015}]{maturana2015voxnet}
Maturana D.,  Scherer S.,  2015, in 2015 IEEE/RSJ international conference on
  intelligent robots and systems (IROS) Voxnet: A 3d convolutional neural
  network for real-time object recognition.
pp 922--928

\bibitem[\protect\citeauthoryear{Mitra \& Nguyen}{Mitra \&
  Nguyen}{2003}]{mitra2003estimating}
Mitra N.~J.,  Nguyen A.,  2003, in Proceedings of the nineteenth annual
  symposium on Computational geometry Estimating surface normals in noisy point
  cloud data.
pp 322--328

\bibitem[\protect\citeauthoryear{Mueller \& Ferreira}{Mueller \&
  Ferreira}{2003}]{mueller2003marvel}
Mueller D.,  Ferreira J.~M.,  2003, in Proceedings of the Technology Enhanced
  Learning International Conference (TEL 03) Marvel: A mixed-reality learning
  environment for vocational training in mechatronics

\bibitem[\protect\citeauthoryear{Munoz, Bagnell, Vandapel \& Hebert}{Munoz
  et~al.}{2009}]{munoz2009contextual}
Munoz D.,  Bagnell J.~A.,  Vandapel N.,    Hebert M.,  2009, in 2009 IEEE
  Conference on Computer Vision and Pattern Recognition Contextual
  classification with functional max-margin markov networks.
pp 975--982

\bibitem[\protect\citeauthoryear{Newcombe, Fox \& Seitz}{Newcombe
  et~al.}{2015}]{newcombe2015dynamicfusion}
Newcombe R.~A.,  Fox D.,    Seitz S.~M.,  2015, in Proceedings of the IEEE
  conference on computer vision and pattern recognition Dynamicfusion:
  Reconstruction and tracking of non-rigid scenes in real-time.
pp 343--352

\bibitem[\protect\citeauthoryear{Niemeyer, Rottensteiner \& Soergel}{Niemeyer
  et~al.}{2014}]{niemeyer2014contextual}
Niemeyer J.,  Rottensteiner F.,    Soergel U.,  2014, ISPRS journal of
  photogrammetry and remote sensing, 87, 152

\bibitem[\protect\citeauthoryear{Osada, Funkhouser, Chazelle \& Dobkin}{Osada
  et~al.}{2002}]{osada2002shape}
Osada R.,  Funkhouser T.,  Chazelle B.,    Dobkin D.,  2002, ACM Transactions
  on Graphics (TOG), 21, 807

\bibitem[\protect\citeauthoryear{Ovsjanikov, Ben-Chen, Solomon, Butscher \&
  Guibas}{Ovsjanikov et~al.}{2012}]{ovsjanikov2012functional}
Ovsjanikov M.,  Ben-Chen M.,  Solomon J.,  Butscher A.,    Guibas L.,  2012,
  ACM Transactions on Graphics (ToG), 31, 1

\bibitem[\protect\citeauthoryear{Pais, Ramalingam, Govindu, Nascimento,
  Chellappa \& Miraldo}{Pais et~al.}{2020}]{pais20203dregnet}
Pais G.~D.,  Ramalingam S.,  Govindu V.~M.,  Nascimento J.~C.,  Chellappa R.,
   Miraldo P.,  2020, in Proceedings of the IEEE/CVF conference on computer
  vision and pattern recognition 3dregnet: A deep neural network for 3d point
  registration.
pp 7193--7203

\bibitem[\protect\citeauthoryear{Pauly, Keiser \& Gross}{Pauly
  et~al.}{2003}]{pauly2003multi}
Pauly M.,  Keiser R.,    Gross M.,  2003, in Computer graphics forum Vol.~22,
  Multi-scale feature extraction on point-sampled surfaces.
pp 281--289

\bibitem[\protect\citeauthoryear{Puy, Boulch \& Marlet}{Puy
  et~al.}{2020}]{puy2020flot}
Puy G.,  Boulch A.,    Marlet R.,  2020, in European conference on computer
  vision Flot: Scene flow on point clouds guided by optimal transport.
pp 527--544

\bibitem[\protect\citeauthoryear{Qi, Su, Mo \& Guibas}{Qi
  et~al.}{2017}]{qi2017pointnet}
Qi C.~R.,  Su H.,  Mo K.,    Guibas L.~J.,  2017, in Proceedings of the IEEE
  conference on computer vision and pattern recognition Pointnet: Deep learning
  on point sets for 3d classification and segmentation.
pp 652--660

\bibitem[\protect\citeauthoryear{Qi, Yi, Su \& Guibas}{Qi
  et~al.}{2017}]{qi2017pointnet++}
Qi C.~R.,  Yi L.,  Su H.,    Guibas L.~J.,  2017, Advances in neural
  information processing systems, 30

\bibitem[\protect\citeauthoryear{Ramachandran \& Varoquaux}{Ramachandran \&
  Varoquaux}{2011}]{ramachandran2011mayavi}
Ramachandran P.,  Varoquaux G.,  2011, Computing in Science \& Engineering, 13,
  40

\bibitem[\protect\citeauthoryear{Rawat \& Wang}{Rawat \&
  Wang}{2017}]{rawat2017deep}
Rawat W.,  Wang Z.,  2017, Neural computation, 29, 2352

\bibitem[\protect\citeauthoryear{Riegler, Osman~Ulusoy \& Geiger}{Riegler
  et~al.}{2017}]{riegler2017octnet}
Riegler G.,  Osman~Ulusoy A.,    Geiger A.,  2017, in Proceedings of the IEEE
  conference on computer vision and pattern recognition Octnet: Learning deep
  3d representations at high resolutions.
pp 3577--3586

\bibitem[\protect\citeauthoryear{Rodol{\`a}, Cosmo, Bronstein, Torsello \&
  Cremers}{Rodol{\`a} et~al.}{2017}]{rodola2017partial}
Rodol{\`a} E.,  Cosmo L.,  Bronstein M.,  Torsello A.,    Cremers D., , 2017,
  Partial functional correspondence In Computer Graphics Forum

\bibitem[\protect\citeauthoryear{Rokhsaritalemi, Sadeghi-Niaraki \&
  Choi}{Rokhsaritalemi et~al.}{2020}]{rokhsaritalemi2020review}
Rokhsaritalemi S.,  Sadeghi-Niaraki A.,    Choi S.-M.,  2020, Applied Sciences,
  10, 636

\bibitem[\protect\citeauthoryear{Roynard, Deschaud \& Goulette}{Roynard
  et~al.}{2018}]{roynard2018classification}
Roynard X.,  Deschaud J.-E.,    Goulette F.,  2018, arXiv preprint
  arXiv:1804.03583

\bibitem[\protect\citeauthoryear{Rusu, Blodow \& Beetz}{Rusu
  et~al.}{2009}]{rusu2009fast}
Rusu R.~B.,  Blodow N.,    Beetz M.,  2009, in 2009 IEEE international
  conference on robotics and automation Fast point feature histograms (fpfh)
  for 3d registration.
pp 3212--3217

\bibitem[\protect\citeauthoryear{Schmidt, Niemeyer, Rottensteiner \&
  Soergel}{Schmidt et~al.}{2014}]{schmidt2014contextual}
Schmidt A.,  Niemeyer J.,  Rottensteiner F.,    Soergel U.,  2014, IEEE
  Geoscience and Remote Sensing Letters, 11, 1614

\bibitem[\protect\citeauthoryear{Schmidt, Newcombe \& Fox}{Schmidt
  et~al.}{2016}]{schmidt2016self}
Schmidt T.,  Newcombe R.,    Fox D.,  2016, IEEE Robotics and Automation
  Letters, 2, 420

\bibitem[\protect\citeauthoryear{Schnabel, Wahl \& Klein}{Schnabel
  et~al.}{2007}]{schnabel2007efficient}
Schnabel R.,  Wahl R.,    Klein R.,  2007, in Computer graphics forum Vol.~26,
  Efficient ransac for point-cloud shape detection.
pp 214--226

\bibitem[\protect\citeauthoryear{Singh, Mittal \& Bhatia}{Singh
  et~al.}{2019}]{singh20193d}
Singh R.~D.,  Mittal A.,    Bhatia R.~K.,  2019, Multimedia Tools and
  Applications, 78, 15951

\bibitem[\protect\citeauthoryear{Speicher, Hall \& Nebeling}{Speicher
  et~al.}{2019}]{speicher2019mixed}
Speicher M.,  Hall B.~D.,    Nebeling M.,  2019, in Proceedings of the 2019 CHI
  conference on human factors in computing systems What is mixed reality?.
pp 1--15

\bibitem[\protect\citeauthoryear{Su, Maji, Kalogerakis \& Learned-Miller}{Su
  et~al.}{2015}]{su2015multi}
Su H.,  Maji S.,  Kalogerakis E.,    Learned-Miller E.,  2015, in Proceedings
  of the IEEE international conference on computer vision Multi-view
  convolutional neural networks for 3d shape recognition.
pp 945--953

\bibitem[\protect\citeauthoryear{Su, Lu, Pan, Wen \& Liu}{Su
  et~al.}{2021}]{su2021roformer}
Su J.,  Lu Y.,  Pan S.,  Wen B.,    Liu Y.,  2021, arXiv preprint
  arXiv:2104.09864

\bibitem[\protect\citeauthoryear{Thomas, Qi, Deschaud, Marcotegui, Goulette \&
  Guibas}{Thomas et~al.}{2019}]{thomas2019kpconv}
Thomas H.,  Qi C.~R.,  Deschaud J.-E.,  Marcotegui B.,  Goulette F.,    Guibas
  L.~J.,  2019, in Proceedings of the IEEE/CVF international conference on
  computer vision Kpconv: Flexible and deformable convolution for point clouds.
pp 6411--6420

\bibitem[\protect\citeauthoryear{Ungureanu, Bogo, Galliani, Sama, Duan,
  Meekhof, St{\"u}hmer, Cashman, Tekin, Sch{\"o}nberger et~al.,}{Ungureanu
  et~al.}{2020}]{ungureanu2020hololens}
Ungureanu D.,  Bogo F.,  Galliani S.,  Sama P.,  Duan X.,  Meekhof C.,
  St{\"u}hmer J.,  Cashman T.~J.,  Tekin B.,  Sch{\"o}nberger J.~L.,    et~al.,
  2020, arXiv preprint arXiv:2008.11239

\bibitem[\protect\citeauthoryear{Vaswani, Shazeer, Parmar, Uszkoreit, Jones,
  Gomez, Kaiser \& Polosukhin}{Vaswani et~al.}{2017}]{vaswani2017attention}
Vaswani A.,  Shazeer N.,  Parmar N.,  Uszkoreit J.,  Jones L.,  Gomez A.~N.,
  Kaiser {\L}.,    Polosukhin I.,  2017, Advances in neural information
  processing systems, 30

\bibitem[\protect\citeauthoryear{Verma, Boyer \& Verbeek}{Verma
  et~al.}{2018}]{verma2018feastnet}
Verma N.,  Boyer E.,    Verbeek J.,  2018, in Proceedings of the IEEE
  conference on computer vision and pattern recognition Feastnet:
  Feature-steered graph convolutions for 3d shape analysis.
pp 2598--2606

\bibitem[\protect\citeauthoryear{Wang, Dunston \& Skiniewski}{Wang
  et~al.}{2004}]{wang2004mixed}
Wang X.,  Dunston P.~S.,    Skiniewski M.,  2004, in Proceedings of the 21st
  International Symposium on Automation and Robotics in Construction (ISARC
  2004) Mixed reality technology applications in construction equipment
  operator training.
pp 21--25

\bibitem[\protect\citeauthoryear{Wang, Sun, Liu, Sarma, Bronstein \&
  Solomon}{Wang et~al.}{2019}]{wang2019dynamic}
Wang Y.,  Sun Y.,  Liu Z.,  Sarma S.~E.,  Bronstein M.~M.,    Solomon J.~M.,
  2019, Acm Transactions On Graphics (tog), 38, 1

\bibitem[\protect\citeauthoryear{Weinmann, Jutzi, Hinz \& Mallet}{Weinmann
  et~al.}{2015}]{weinmann2015semantic}
Weinmann M.,  Jutzi B.,  Hinz S.,    Mallet C.,  2015, ISPRS Journal of
  Photogrammetry and Remote Sensing, 105, 286

\bibitem[\protect\citeauthoryear{Weinmann, Jutzi, Mallet \& Weinmann}{Weinmann
  et~al.}{2017}]{weinmann2017geometric}
Weinmann M.,  Jutzi B.,  Mallet C.,    Weinmann M.,  2017, ISPRS Annals of
  Photogrammetry, Remote Sensing \& Spatial Information Sciences, 4

\bibitem[\protect\citeauthoryear{Weinmann, Schmidt, Mallet, Hinz, Rottensteiner
  \& Jutzi}{Weinmann et~al.}{2015}]{weinmann2015contextual}
Weinmann M.,  Schmidt A.,  Mallet C.,  Hinz S.,  Rottensteiner F.,    Jutzi B.,
   2015, ISPRS Annals of the Photogrammetry, Remote Sensing and Spatial
  Information Sciences II-3 (2015), Nr. W4, 2, 271

\bibitem[\protect\citeauthoryear{West, Webb, Lersch, Pothier, Triscari \&
  Iverson}{West et~al.}{2004}]{west2004context}
West K.~F.,  Webb B.~N.,  Lersch J.~R.,  Pothier S.,  Triscari J.~M.,
  Iverson A.~E.,  2004, in Automatic Target Recognition XIV Vol.~5426,
  Context-driven automated target detection in 3d data.
pp 133--143

\bibitem[\protect\citeauthoryear{Xu, Liu, Wang, Liu \& Zeng}{Xu
  et~al.}{2021}]{xu2021omnet}
Xu H.,  Liu S.,  Wang G.,  Liu G.,    Zeng B.,  2021, in Proceedings of the
  IEEE/CVF International Conference on Computer Vision Omnet: Learning
  overlapping mask for partial-to-partial point cloud registration.
pp 3132--3141

\bibitem[\protect\citeauthoryear{Xu, Fan, Xu, Zeng \& Qiao}{Xu
  et~al.}{2018}]{xu2018spidercnn}
Xu Y.,  Fan T.,  Xu M.,  Zeng L.,    Qiao Y.,  2018, in Proceedings of the
  European Conference on Computer Vision (ECCV) Spidercnn: Deep learning on
  point sets with parameterized convolutional filters.
pp 87--102

\bibitem[\protect\citeauthoryear{Yu, Li, Saleh, Busam \& Ilic}{Yu
  et~al.}{2021}]{yu2021cofinet}
Yu H.,  Li F.,  Saleh M.,  Busam B.,    Ilic S.,  2021, Advances in Neural
  Information Processing Systems, 34, 23872

\bibitem[\protect\citeauthoryear{Zeng, Song, Nie{\ss}ner, Fisher, Xiao \&
  Funkhouser}{Zeng et~al.}{2017}]{zeng20173dmatch}
Zeng A.,  Song S.,  Nie{\ss}ner M.,  Fisher M.,  Xiao J.,    Funkhouser T.,
  2017, in Proceedings of the IEEE conference on computer vision and pattern
  recognition 3dmatch: Learning local geometric descriptors from rgb-d
  reconstructions.
pp 1802--1811

\bibitem[\protect\citeauthoryear{Zhang}{Zhang}{1994}]{zhang1994iterative}
Zhang Z.,  1994, International journal of computer vision, 13, 119

\bibitem[\protect\citeauthoryear{Zhou, Park \& Koltun}{Zhou
  et~al.}{2016}]{zhou2016fast}
Zhou Q.-Y.,  Park J.,    Koltun V.,  2016, in European conference on computer
  vision Fast global registration.
pp 766--782

\bibitem[\protect\citeauthoryear{Zhou, Park \& Koltun}{Zhou
  et~al.}{2018}]{zhou2018open3d}
Zhou Q.-Y.,  Park J.,    Koltun V.,  2018, arXiv preprint arXiv:1801.09847

\bibitem[\protect\citeauthoryear{Zollh{\"o}fer, Nie{\ss}ner, Izadi, Rehmann,
  Zach, Fisher, Wu, Fitzgibbon, Loop, Theobalt et~al.,}{Zollh{\"o}fer
  et~al.}{2014}]{zollhofer2014real}
Zollh{\"o}fer M.,  Nie{\ss}ner M.,  Izadi S.,  Rehmann C.,  Zach C.,  Fisher
  M.,  Wu C.,  Fitzgibbon A.,  Loop C.,  Theobalt C.,    et~al., 2014, ACM
  Transactions on Graphics (ToG), 33, 1

\end{thebibliography}
